# Testing and Confidence Intervals for High Dimensional Proportional Hazards Model

Ethan X. Fang   Yang Ning   Han Liu[*]

December 1st, 2014

## Abstract

This paper proposes a decorrelation-based approach to test hypotheses and construct confidence intervals for the low dimensional component of high dimensional proportional hazards models. Motivated by the geometric projection principle, we propose new decorrelated score, Wald and partial likelihood ratio statistics. Without assuming model selection consistency, we prove the asymptotic normality of these test statistics, establish their semiparametric optimality. We also develop new procedures for constructing pointwise confidence intervals for the baseline hazard function and baseline survival function. Thorough numerical results are provided to back up our theory.

## 1  Introduction

The proportional hazards model (Cox, 1972) is one of the most important tools for analyzing time to event data, and finds wide applications in epidemiology, medicine, economics, and sociology (Kalbfleisch and Prentice, 2011). This model is semiparametric by treating the baseline hazard function as an infinite dimensional nuisance parameter. To infer the finite dimensional parameter of interest, Cox (1972, 1975) proposes the partial likelihood approach which is invariant to the baseline hazard function. In low dimensional settings, Tsiatis (1981); Andersen and Gill (1982) have established the consistency and asymptotic normality of the maximum partial likelihood estimator.

In high dimensional settings when the number of covariates $d$ is larger than the sample size $n$, the partial maximum likelihood estimation is an ill-posed problem. To solve this problem, we resort to the penalized estimators (Tibshirani, 1997; Fan and Li, 2002; Gui and Li, 2005). Under the condition $d = o(n^{1/4})$, Cai et al. (2005) establish the oracle properties for the maximum penalized partial likelihood estimator using the SCAD penalty. Other types of estimation procedures and their theoretical properties are studied by Zhang and Lu (2007); Wang et al. (2009); Antoniadis et al. (2010); Zhao and Li (2012). In particular, under the ultra-high dimensional regime that $d = o(\exp(n/s))$, Bradic et al. (2011); Huang et al. (2013); Kong and Nan (2014) establish the oracle properties and statistical error bounds of maximum penalized partial likelihood estimator, where $s$ denotes the number of nonzero elements in the parametric component of the Cox model.

---

[*]Department of Operations Research and Financial Engineering, Princeton University, Princeton, NJ 08544, USA; e-mail: {xingyuan,yangning,hanliu}@princeton.edu

Though significant progress has been made towards developing the estimation theory. Little work exists on the inferential aspects (e.g., testing hypothesis or constructing confidence intervals) of high dimensional proportional hazard models. A notable exception is Bradic et al. (2011), who establish the limiting distribution of the oracle estimator. However, such a result hinges on model selection consistency, which is not always possible in applications. To the best of our knowledge, uncertainty assessment for low dimensional parameters of high dimensional proportional hazards model remains an open problem. This paper aims to close this gap by developing valid inferential procedures and theory for high dimensional proportional hazards models. In particular, we test hypotheses and construct confidence intervals for a scalar component of a $d$ dimensional parameter vector[1]. Compared with existing work, our method does not require any types of irrepresentable condition or the minimal signal strength condition, thus is more practical in applications.

More specifically, we develop a unified inferential framework by extending the classical score, Wald and partial likelihood ratio tests to high dimensional hazards models. The key ingredient of our construction of these tests is a novel high dimensional decorrelation device of the score function. Theoretically, we establish the asymptotic distributions of these test statistics under the null. Using the same idea, we construct optimal confidence intervals for the parameters of interest. In addition, we consider the problems on inferring the baseline hazard and survival functions and separately establish their asymptotic normalities.

The rest of this paper is organized as follows. In Section 2, we provide some background on the proportional hazards model. In Section 3, we propose the methods for conducting hypothesis testing and constructing confidence intervals for low dimensional components of regression parameters. In Section 4, we provide theoretical analysis of the proposed methods. The inference on the baseline hazard function is studied in Section 5. In Section 6, we investigate the empirical performance of these methods. Section 7 contains the summary and discussions. More technical details and an extension to the multivariate failure time data are presented in the Appendix.

## 2 Background

We start with an introduction of notation. Let $\mathbf{a} = (a_1, ..., a_d)^T \in \mathbb{R}^d$ be a $d$ dimensional vector and $\mathbf{A} = [a_{jk}] \in \mathbb{R}^{d \times d}$ be a $d$ by $d$ matrix. Let $\text{supp}(\mathbf{a}) = \{j : a_j \neq 0\}$. For $0 < q < \infty$, we define $\ell_0$, $\ell_q$ and $\ell_\infty$ vector norms as $\|\mathbf{a}\|_0 = \text{card}(\text{supp}(\mathbf{a}))$, $\|\mathbf{a}\|_q = (\sum_{j=1}^d \|\mathbf{a}_j\|^q)^{1/q}$ and $\|\mathbf{a}\|_\infty = \max_{1 \leq j \leq d} |a_j|$. We define the matrix $\ell_\infty$-norm as the elementwise sup-norm that $\|\mathbf{A}\|_\infty = \max_{1 \leq j,k \leq d} |a_{jk}|$. Let $\mathbf{I}_d$ be the identity matrix in $\mathbb{R}^{d \times d}$. For a sequence of random variables $\{X_n\}_{n=1}^\infty$ and a random variable $Y$, we denote $X_n$ weakly converges to $Y$ by $X_n \xrightarrow{d} Y$. We denote $[n] = \{1, \ldots, n\}$.

---

[1] It is straightforward to extend the setting from univariate scalar to multivariate parameter vector.

## 2.1 Cox's Proportional Hazards Model

We briefly review the Cox's proportional hazards model. Let $Q$ be the time to event; $R$ be the censoring time, and $\boldsymbol{X}(t) = (X_1(t), ..., X_d(t))^T$ be the $d$ dimensional time dependent covariates at time $t$. We consider the non-informative censoring setting that $Q$ and $R$ are conditionally independent given $\boldsymbol{X}(t)$. Let $W = \min\{Q, R\}$ and $\Delta = \mathbf{1}\{Q \leq R\}$ denote the observed survival time and censoring indicator. Let $\tau$ be the end of study time. We observe $n$ independent copies of $\{(\boldsymbol{X}(t), W, \Delta) : 0 \leq t \leq \tau\}$,

$$\Big\{(\boldsymbol{X}_i(t), W_i, \Delta_i) : 0 \leq t \leq \tau\Big\}_{i \in [n]}.$$

We denote $\lambda\{t|\boldsymbol{X}(t)\}$ as the conditional hazard rate function at time $t$ given the covariates $\boldsymbol{X}(t)$. Under the proportional hazards model, we assume that

$$\lambda\{t|\boldsymbol{X}(t)\} = \lambda_0(t)\exp\{\boldsymbol{X}^T(t)\boldsymbol{\beta}^*\},$$

where $\lambda_0(t)$ is an unknown baseline hazard rate function, and $\boldsymbol{\beta}^* \in \mathbb{R}^d$ is an unknown parameter.

## 2.2 Penalized Estimation

Following Andersen and Gill (1982), we introduce some counting process notation. For each $i$, let $N_i(t) := \mathbf{1}\{W_i \leq t, \Delta_i = 1\}$ be the counting process, and $Y_i(t) := \mathbf{1}\{W_i \geq t\}$ be the at risk process for subject $i$. Assume that the process $Y_i(t)$ is left continuous with its right-hand limits satisfying $\mathbb{P}(Y_i(t) = 1, 0 \leq t \leq \tau) > C_\tau$ for some positive constant $C_\tau$. The negative log-partial likelihood is

$$\mathcal{L}(\boldsymbol{\beta}) = -\frac{1}{n}\Big(\sum_{i=1}^n \int_0^\tau \boldsymbol{X}_i^T(u)\boldsymbol{\beta}dN_i(u) - \int_0^\tau \log\Big[\sum_{i=1}^n Y_i(u)\exp\{\boldsymbol{X}_i^T(u)\boldsymbol{\beta}\}\Big]d\overline{N}(u)\Big),$$

where $\overline{N}(t) = \sum_{i=1}^n N_i(t)$.

When the dimension $d$ is fixed and smaller than the sample size $n$, $\boldsymbol{\beta}^*$ can be estimated by the maximum partial likelihood estimator (Andersen and Gill, 1982). However, in high dimensional settings where $n < d$, the maximum partial likelihood estimator is not well defined. To solve this problem, Fan and Li (2002) impose the sparsity assumption and propose the penalized estimator

$$\widehat{\boldsymbol{\beta}} := \operatorname*{argmin}_{\boldsymbol{\beta} \in \mathbb{R}^d}\big\{\mathcal{L}(\boldsymbol{\beta}) + \mathcal{P}_\lambda(\boldsymbol{\beta})\big\}, \tag{2.1}$$

where $\mathcal{P}_\lambda(\cdot)$ is a sparsity-inducing penalty function and $\lambda$ is a tuning parameter. Bradic et al. (2011) and Huang et al. (2013) establish the rates of convergence and oracle properties of the maximum penalized partial likelihood estimators $\widehat{\boldsymbol{\beta}}$ using SCAD and Lasso penalties. For notational simplicity, we focus on the Lasso penalized estimator in this paper and indicate that similar properties hold for the SCAD penalty. Existing works generally impose the following assumptions.

**Assumption 2.1.** The difference of the covariates is uniformly bounded:

$$\sup_{0 \leq t \leq \tau} \max_{i,i' \leq n} \max_{1 \leq j \leq d} |X_{ij}(t) - X_{i'j}(t)| \leq C_X,$$

for some constant $C_X > 0$.

**Assumption 2.2.** For any set $\mathcal{S} \subset \{1, ..., d\}$ where $|\mathcal{S}| \asymp s$ and any vector $\mathbf{v}$ belonging to the cone $\mathcal{C}(\xi, \mathcal{S}) = \{\mathbf{v} \in \mathbb{R}^d : \|\mathbf{v}_{\mathcal{S}^C}\|_1 \leq \xi \|\mathbf{v}_{\mathcal{S}}\|_1\}$, it holds that

$$\kappa\big(\xi, \mathcal{S}; \nabla^2 \mathcal{L}(\boldsymbol{\beta}^*)\big) = \inf_{\mathbf{0} \neq \mathbf{v} \in \mathcal{C}(\xi,\mathcal{S})} \frac{s^{1/2}\{\mathbf{v}^T \nabla^2 \mathcal{L}(\boldsymbol{\beta}^*)\mathbf{v}\}^{1/2}}{\|\mathbf{v}_{\mathcal{S}}\|_1} \geq \lambda_{\min} > 0.$$

Note that the bounded covariate condition in Assumption 2.1, which is imposed by both Bradic et al. (2011) and Huang et al. (2013), holds in most real applications. Assumption 2.2 is known as the compatibility factor condition which is also used by Huang et al. (2013). This assumption essentially bounds the minimal eigenvalue of the Hessian matrix $\nabla^2 \mathcal{L}(\boldsymbol{\beta}^*)$ from below for those directions within the cone $\mathcal{C}(\xi, \mathcal{S})$. In particular, the validity of this assumption has been verified in Theorem 4.1 of Huang et al. (2013). Under these assumptions, Huang et al. (2013) derive the rate of convergence of the Lasso estimator $\widehat{\boldsymbol{\beta}}$ under the $\ell_1$-norm. More specifically, they prove that under Assumptions 2.1 and 2.2, if $\|\boldsymbol{\beta}^*\|_0 = s$ and $\lambda \asymp \sqrt{n^{-1} \log d}$, it holds that

$$\|\widehat{\boldsymbol{\beta}} - \boldsymbol{\beta}^*\|_1 = \mathcal{O}_\mathbb{P}(s\lambda), \tag{2.2}$$

which establishes the estimation consistency in the high dimensional regime.

**Additional Notations** For a vector $\mathbf{u}$, we denote $\mathbf{u}^{\otimes 0} = 1$, $\mathbf{u}^{\otimes 1} = \mathbf{u}$ and $\mathbf{u}^{\otimes 2} = \mathbf{u}\mathbf{u}^T$. Denote

$$\begin{aligned} S^{(r)}(t, \boldsymbol{\beta}) &= \frac{1}{n} \sum_{i=1}^n \boldsymbol{X}_i^{\otimes r}(t) Y_i(t) \exp\{\boldsymbol{X}_i^T(t)\boldsymbol{\beta}\} \text{ for } r = 0, 1, 2, \quad \overline{\boldsymbol{Z}}(t, \boldsymbol{\beta}) = \frac{S^{(1)}(t, \boldsymbol{\beta})}{S^{(0)}(t, \boldsymbol{\beta})}, \\ \mathbf{V}_n(t, \boldsymbol{\beta}) &= \sum_{i=1}^n \frac{Y_i(t) \exp\{\boldsymbol{X}_i(t)^T \boldsymbol{\beta}\}}{n S^{(0)}(t, \boldsymbol{\beta})} \{\boldsymbol{X}_i(t) - \overline{\boldsymbol{Z}}(t, \boldsymbol{\beta})\}^{\otimes 2} = \frac{S^{(2)}(t, \boldsymbol{\beta})}{S^{(0)}(t, \boldsymbol{\beta})} - \overline{\boldsymbol{Z}}(t, \boldsymbol{\beta})^{\otimes 2}. \end{aligned} \tag{2.3}$$

The gradient of $\mathcal{L}(\boldsymbol{\beta})$ is

$$\nabla \mathcal{L}(\boldsymbol{\beta}) = \frac{\partial \mathcal{L}(\boldsymbol{\beta})}{\partial \boldsymbol{\beta}} = -\frac{1}{n} \sum_{i=1}^n \int_0^\tau \{\boldsymbol{X}_i(u) - \overline{\boldsymbol{Z}}(u, \boldsymbol{\beta})\} dN_i(u), \tag{2.4}$$

and the Hessian matrix of $\mathcal{L}(\boldsymbol{\beta})$ is

$$\nabla^2 \mathcal{L}(\boldsymbol{\beta}) = \frac{1}{n} \int_0^\tau \mathbf{V}_n(u, \boldsymbol{\beta}) d\overline{N}(u) = \frac{1}{n} \int_0^\tau \Big\{\frac{S^{(2)}(u, \boldsymbol{\beta})}{S^{(0)}(u, \boldsymbol{\beta})} - \overline{\boldsymbol{Z}}(u, \boldsymbol{\beta})^{\otimes 2}\Big\} d\overline{N}(u). \tag{2.5}$$

We denote the population versions of above defined quantities by

$$\mathbf{s}^{(r)}(t, \boldsymbol{\beta}) = \mathbb{E}\big[Y(t) \boldsymbol{X}(t)^{\otimes r} \exp\{\boldsymbol{X}(t)^T \boldsymbol{\beta}\}\big] \text{ for } r = 0, 1, 2; \quad \mathbf{e}(t, \boldsymbol{\beta}) = \mathbf{s}^{(1)}(t, \boldsymbol{\beta}) / \mathbf{s}^{(0)}(t, \boldsymbol{\beta}), \tag{2.6}$$

and

$$\mathbf{H}(\boldsymbol{\beta}) = \mathbb{E}\Big[\int_0^\tau \Big\{\frac{\mathbf{s}^{(2)}(t, \boldsymbol{\beta})}{\mathbf{s}^{(0)}(t, \boldsymbol{\beta})} - \mathbf{e}(t, \boldsymbol{\beta})^{\otimes 2}\Big\} dN(t)\Big], \text{ and } \mathbf{H}^* = \mathbf{H}(\boldsymbol{\beta}^*), \tag{2.7}$$

where $\mathbf{H}^*$ is the Fisher information matrix based on the partial likelihood.

# 3 Testing Hyptheses and Constructing Confidence Intervals

While estimation consistency has been established in high dimensions, it remains challenging to develop inferential procedures (e.g., confidence intervals and testing) for high dimensional proportional hazards model. In this section, we propose three novel hypothesis testing procedures. The proposed tests can be viewed as high dimensional counterparts of the conventional score, Wald, and partial likelihood ratio tests.

Hereafter, for notational simplicity, we partition the vector $\boldsymbol{\beta}$ as $\boldsymbol{\beta} = (\alpha, \boldsymbol{\theta}^T)^T$, where $\alpha = \beta_1 \in \mathbb{R}$ is the parameter of interest; $\boldsymbol{\theta} = (\beta_2, ..., \beta_d)^T \in \mathbb{R}^{d-1}$ is the vector of nuisance parameters, and we denote $\mathcal{L}(\boldsymbol{\beta})$ by $\mathcal{L}(\alpha, \boldsymbol{\theta})$. Let $\nabla^2_{\alpha\alpha}\mathcal{L}(\boldsymbol{\beta})$, $\nabla^2_{\alpha\boldsymbol{\theta}}\mathcal{L}(\boldsymbol{\beta})$ and $\nabla^2_{\boldsymbol{\theta}\boldsymbol{\theta}}\mathcal{L}(\boldsymbol{\beta})$ be the corresponding partitions of $\nabla^2\mathcal{L}(\boldsymbol{\beta})$. Let $\mathbf{H}^*_{\alpha\alpha}$, $\mathbf{H}^*_{\alpha\boldsymbol{\theta}}$ and $\mathbf{H}^*_{\boldsymbol{\theta}\boldsymbol{\theta}}$ be the corresponding partitions of $\mathbf{H}^*$, where $\mathbf{H}^*$ is defined in (2.7). For instances, $\mathbf{H}^*_{\boldsymbol{\theta}\alpha} = \mathbf{H}^*_{2:d,1} \in \mathbb{R}^{d-1}$ and $\nabla^2_{\boldsymbol{\theta}\boldsymbol{\theta}}\mathcal{L}(\boldsymbol{\beta}) = \nabla^2_{2:d,2:d}\mathcal{L}(\boldsymbol{\beta}) \in \mathbb{R}^{(d-1)\times(d-1)}$. Throughout this paper, without loss of generality, we test the hypothesis $H_0$: $\alpha^* = 0$ versus $H_1$: $\alpha^* \neq 0$. Note that the extension to tests for a multi-dimensional vector $\boldsymbol{\alpha}^* \in \mathbb{R}^{d_0}$, where $d_0$ is fixed, is straightforward.

## 3.1 Decorrelated Score Test

In the classical low dimensional setting, we can exploit the profile partial score function

$$S(\alpha) = \nabla_\alpha \mathcal{L}(\alpha, \boldsymbol{\theta})|_{\boldsymbol{\theta} = \widehat{\boldsymbol{\theta}}(\alpha)}$$

to conduct test, where $\widehat{\boldsymbol{\theta}}(\alpha) = \operatorname{argmin}_{\boldsymbol{\theta}} \mathcal{L}(\alpha, \boldsymbol{\theta})$ is the maximum partial likelihood estimator for $\boldsymbol{\theta}$ with a fixed $\alpha$. Under the null hypothesis that $\alpha^* = 0$, when $d$ is fixed while $n$ goes to infinity, it holds that $\sqrt{n}S(0) \xrightarrow{d} N(0, \mathbf{H}^*_{\alpha\alpha})$. If $n(\mathbf{H}^*_{\alpha\alpha})^{-1}S^2(0)$ is larger than the $(1-\eta)$th quantile of a chi-squared distribution with one degree of freedom, we reject the null hypothesis. Classical asymptotic theory shows that this procedure controls type I error with significance level $\eta$.

However, in high dimensions, the profile partial score function $S(\alpha)$ with $\widehat{\boldsymbol{\theta}}(\alpha)$ replaced by a penalized estimator, say the corresponding components of $\widehat{\boldsymbol{\beta}}$ in (2.1), does not yield a tractable limiting distribution due to the existence of a large number of nuisance parameters. To address this problem, we construct a new type of score function for $\alpha$ that is asymptotically normal even in high dimensions. The key component of our procedure is a high dimensional decorrelation device, aiming to handle the impact of the high dimensional nuisance vector.

More specifically, we propose a decorrelated score test for $H_0$: $\alpha^* = 0$. We first estimate $\boldsymbol{\theta}^*$ by $\widehat{\boldsymbol{\theta}}$ using the $\ell_1$ penalized estimator $\widehat{\boldsymbol{\beta}}$ in (2.1). Next, we calculate a linear combination of the partial score function $\nabla_{\boldsymbol{\theta}}\mathcal{L}(0, \widehat{\boldsymbol{\theta}})$ to best approximate $\nabla_\alpha \mathcal{L}(0, \widehat{\boldsymbol{\theta}})$. The population version of the vector of coefficients in the best linear combination can be calculated as

$$\begin{aligned}
\mathbf{w}^* &= \underset{\mathbf{w}}{\operatorname{argmin}}\, \mathbb{E}\big\{\nabla_\alpha \mathcal{L}(0, \boldsymbol{\theta}^*) - \mathbf{w}^T \nabla_{\boldsymbol{\theta}}\mathcal{L}(0, \boldsymbol{\theta}^*)\big\}^2 \\
&= \mathbb{E}\{\nabla_{\boldsymbol{\theta}}\mathcal{L}(0, \boldsymbol{\theta}^*)\nabla_{\boldsymbol{\theta}}\mathcal{L}(0, \boldsymbol{\theta}^*)^T\}^{-1}\mathbb{E}\{\nabla_{\boldsymbol{\theta}}\mathcal{L}(0, \boldsymbol{\theta}^*)\nabla_\alpha\mathcal{L}(0, \boldsymbol{\theta}^*)\} = \mathbf{H}^{*-1}_{\boldsymbol{\theta}\boldsymbol{\theta}}\mathbf{H}^*_{\boldsymbol{\theta}\alpha},
\end{aligned} \quad (3.1)$$

where the last equality is by the second Bartlett identity (Tsiatis, 1981). In fact, $\mathbf{w}^{*T}\nabla_{\boldsymbol{\theta}}\mathcal{L}(0, \boldsymbol{\theta}^*)$ can be interpreted as the projection of $\nabla_\alpha \mathcal{L}(0, \boldsymbol{\theta}^*)$ onto the linear span of the partial score function $\nabla_{\boldsymbol{\theta}}\mathcal{L}(0, \boldsymbol{\theta}^*)$. In high dimensions, one cannot directly estimate $\mathbf{w}^*$ by the corresponding sample version since the problem is ill-posed. Motivated by the definition of $\mathbf{w}^*$ in (3.1), we estimate it by the Dantzig selector,

$$\widehat{\mathbf{w}} = \underset{\mathbf{w} \in \mathbb{R}^{d-1}}{\operatorname{argmin}} \|\mathbf{w}\|_1, \text{ subject to } \|\nabla^2_{\alpha\boldsymbol{\theta}}\mathcal{L}(\widehat{\boldsymbol{\beta}}) - \mathbf{w}^T \nabla^2_{\boldsymbol{\theta}\boldsymbol{\theta}}\mathcal{L}(\widehat{\boldsymbol{\beta}})\|_\infty \leq \lambda', \quad (3.2)$$

where $\lambda'$ is a tuning parameter. Since $\mathbf{w}^*$ is of high dimension $d-1$, we impose the sparsity condition on $\mathbf{w}^*$. Given $\widehat{\boldsymbol{\theta}}$ and $\widehat{\mathbf{w}}$, we propose a decorrelated score function for $\alpha$ as

$$\widehat{U}(\alpha, \widehat{\boldsymbol{\theta}}) = \nabla_\alpha \mathcal{L}(\alpha, \widehat{\boldsymbol{\theta}}) - \widehat{\mathbf{w}}^T \nabla_{\boldsymbol{\theta}} \mathcal{L}(\alpha, \widehat{\boldsymbol{\theta}}). \quad (3.3)$$

Geometrically, the decorrelated score function is approximately orthogonal to any component of the nuisance score function $\nabla_{\boldsymbol{\theta}} \mathcal{L}(0, \boldsymbol{\theta}^*)$. This orthogonality property, which does not hold for the original score function $\nabla_\alpha \mathcal{L}(\alpha, \widehat{\boldsymbol{\theta}})$, reduces the variability caused by the nuisance parameters. A geometric illustration of the decorrelation-based methods is provided in Figure 1, which also incorporates the illustration of the decorrelated Wald and partial likelihood ratio tests to be introduced in the following subsections. Technically, the uncertainty of estimating $\boldsymbol{\theta}$ in the partial score function $\nabla_\alpha \mathcal{L}(\alpha, \widehat{\boldsymbol{\theta}})$ can be reduced by subtracting the decorrelation term $\widehat{\mathbf{w}}^T \nabla_{\boldsymbol{\theta}} \mathcal{L}(\alpha, \widehat{\boldsymbol{\theta}})$. As will be shown in the next section, this is a key step to establish the result that the decorrelated score function $\widehat{U}(0, \widehat{\boldsymbol{\theta}})$ weakly converges to $N(0, H_{\alpha|\boldsymbol{\theta}})$ under the null, where $H_{\alpha|\boldsymbol{\theta}} = \mathbf{H}^*_{\alpha\alpha} - \mathbf{H}^*_{\alpha\boldsymbol{\theta}} \mathbf{H}^{*-1}_{\boldsymbol{\theta}\boldsymbol{\theta}} \mathbf{H}^*_{\boldsymbol{\theta}\alpha}$. This further explains why the decorrelated score function $\widehat{U}(\alpha, \widehat{\boldsymbol{\theta}})$ rather than the original score function $\nabla_\alpha \mathcal{L}(\alpha, \widehat{\boldsymbol{\theta}})$ should be used as the inferential function in high dimensions. On the other hand, in the low dimensional setting, it can be shown that the decorrelated score function $\widehat{U}(\alpha, \widehat{\boldsymbol{\theta}})$ is asymptotically equivalent to the profile partial score function $S(\alpha)$.

To test if $\alpha^* = 0$, we need to standardize $\widehat{U}(0, \widehat{\boldsymbol{\theta}})$ in order to construct the test statistic. We estimate $H_{\alpha|\boldsymbol{\theta}}$ by

$$\widehat{H}_{\alpha|\boldsymbol{\theta}} = \nabla^2_{\alpha\alpha}\mathcal{L}(\widehat{\alpha}, \widehat{\boldsymbol{\theta}}) - \widehat{\mathbf{w}}^T \nabla^2_{\boldsymbol{\theta}\alpha}\mathcal{L}(\widehat{\alpha}, \widehat{\boldsymbol{\theta}}). \quad (3.4)$$

Hence, we define the decorrelated score test statistic as

$$\widehat{S}_n = n \widehat{H}^{-1}_{\alpha|\boldsymbol{\theta}} \widehat{U}^2(0, \widehat{\boldsymbol{\theta}}), \text{ where } \widehat{U}(0, \widehat{\boldsymbol{\theta}}) \text{ and } \widehat{H}_{\alpha|\boldsymbol{\theta}} \text{ are defined in (3.3) and (3.4).} \quad (3.5)$$

In the next section, we show that under the null, $\widehat{S}_n$ converges weakly to a chi-squared distribution with one degree of freedom. Given a significance level $\eta \in (0,1)$, the score test $\psi_S(\eta)$ is

$$\psi_S(\eta) = \begin{cases} 0 & \text{if } \widehat{S}_n \leq \chi^2_1(1-\eta) \\ 1 & \text{otherwise} \end{cases}, \quad (3.6)$$

where $\chi^2_1(1-\eta)$ denotes the $(1-\eta)$th quantile of a chi-squared random variable with one degree of freedom, and the null hypothesis $\alpha^* = 0$ is rejected if and only if $\psi_S(\eta) = 1$.

## 3.2 Confidence Intervals and Decorrelated Wald Test

The decorrelated score test does not provide a confidence interval for $\alpha^*$ with a desired coverage probability. In low dimensions, by examining the limiting distribution of the maximum partial likelihood estimator, we can get a confidence interval for $\alpha^*$ (Andersen and Gill, 1982), which is equivalent to the classical Wald test. This subsection extends the classical Wald test for the proportional hazards model to high dimensional settings to construct confidence intervals for the parameters of interest.

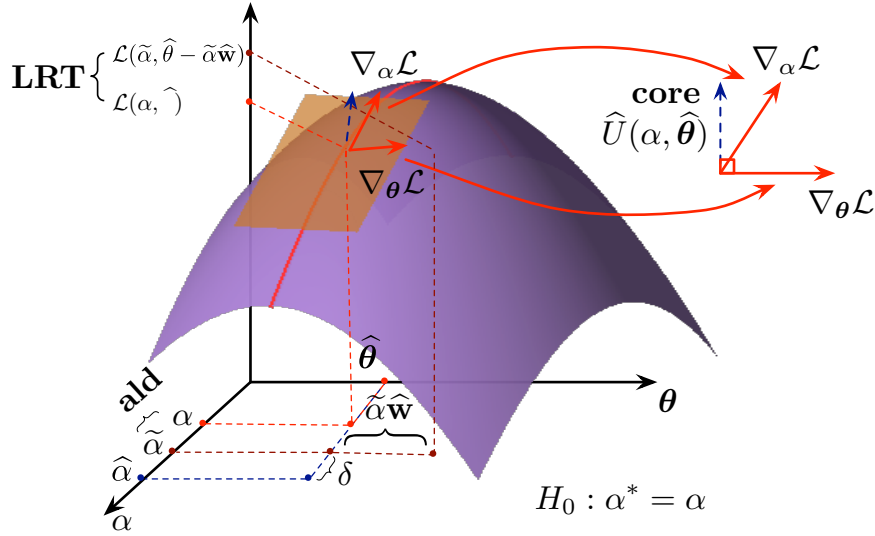

Figure 1: Geometric illustration of the decorrelated score, Wald and partial likelihood ratio tests. The purple surface corresponds to the log-partial likelihood function. The orange plane is the tangent plane of the surface at point $(\alpha, \widehat{\boldsymbol{\theta}})$. The two red arrows in the orange plane represent $\nabla_\alpha \mathcal{L}$ and $\nabla_{\boldsymbol{\theta}} \mathcal{L}$. The correlated score function in blue is the projection of $\nabla_\alpha \mathcal{L}$ onto the space orthogonal to $\nabla_{\boldsymbol{\theta}} \mathcal{L}$. Given Lasso estimator $\widehat{\alpha}$, the decorrelated Wald estimator is $\widetilde{\alpha} = \widehat{\alpha} - \delta$, where $\delta = \{\partial \widehat{U}(\widehat{\alpha}, \widehat{\boldsymbol{\theta}})/\partial \alpha\}^{-1} \widehat{U}(\widehat{\alpha}, \widehat{\boldsymbol{\theta}})$. The decorrelated partial likelihood ratio test compares the log-partial likelihood function values at $(\alpha, \widehat{\boldsymbol{\theta}})$ and $(\widetilde{\alpha}, \widehat{\boldsymbol{\theta}} - \widetilde{\alpha}\widehat{\mathbf{w}})$.

The key idea of performing Wald test is to derive a regular estimator for $\alpha^*$. Our procedure is based on the deccorelated score function $\widehat{U}(\alpha, \widehat{\boldsymbol{\theta}})$ in (3.3). Since $\widehat{U}(\alpha, \widehat{\boldsymbol{\theta}})$ serves as an approximately unbiased estimating equation for $\alpha$, the root of the equation $\widehat{U}(\alpha, \widehat{\boldsymbol{\theta}}) = 0$ with respect to $\alpha$ defines an estimator for $\alpha^*$. However, searching for the root may be computationally intensive, especially when $\alpha$ is multi-dimensional. To reduce the computational cost, we exploit a closed-form estimator $\widetilde{\alpha}$ obtained by linearizing $\widehat{U}(\alpha, \widehat{\boldsymbol{\theta}}) = 0$ at the initial estimator $\widehat{\alpha}$. More specifically, let $\widehat{\boldsymbol{\beta}} = (\widehat{\alpha}, \widehat{\boldsymbol{\theta}}^T)^T$ be the $\ell_1$ penalized estimator in (2.1), we adopt the following one-step estimator,

$$\widetilde{\alpha} = \widehat{\alpha} - \left\{\frac{\partial \widehat{U}(\widehat{\alpha}, \widehat{\boldsymbol{\theta}})}{\partial \alpha}\right\}^{-1} \widehat{U}(\widehat{\alpha}, \widehat{\boldsymbol{\theta}}), \text{ where } \widehat{U}(\widehat{\alpha}, \widehat{\boldsymbol{\theta}}) = \nabla_\alpha \mathcal{L}(\widehat{\alpha}, \widehat{\boldsymbol{\theta}}) - \widehat{\mathbf{w}}^T \nabla_{\boldsymbol{\theta}} \mathcal{L}(\widehat{\alpha}, \widehat{\boldsymbol{\theta}}). \qquad (3.7)$$

In the next section, we prove that $\sqrt{n}(\widetilde{\alpha} - \alpha^*)$ converges weakly to $N(0, H_{\alpha|\boldsymbol{\theta}}^{-1})$. Hence, let $Z_{1-\eta/2}$ be the $(1-\eta/2)$-th quantile of $N(0, 1)$. We show that

$$\left[\widetilde{\alpha} - n^{-1/2} Z_{1-\eta/2} \widehat{H}_{\alpha|\boldsymbol{\theta}}^{-1/2}, \widetilde{\alpha} + n^{-1/2} Z_{1-\eta/2} \widehat{H}_{\alpha|\boldsymbol{\theta}}^{-1/2}\right]$$

is a $100(1-\eta)\%$ confidence interval for $\alpha^*$.

From the perspective of hypothesis testing, the decorrelated Wald test statistic for $H_0$: $\alpha^* = 0$ versus $H_1$: $\alpha^* \neq 0$ is

$$\widehat{W}_n = n\widehat{H}_{\alpha|\boldsymbol{\theta}}\widetilde{\alpha}^2, \quad \text{where } \widetilde{\alpha} \text{ and } \widehat{H}_{\alpha|\boldsymbol{\theta}} \text{ are defined in (3.7) and (3.4), respectively.} \tag{3.8}$$

Consequently, the decorrelated Wald test at significance level $\eta$ is

$$\psi_W(\eta) = \begin{cases} 0 & \text{if } \widehat{W}_n \leq \chi_1^2(1-\eta), \\ 1 & \text{otherwise,} \end{cases} \tag{3.9}$$

and the null hypothesis $\alpha^* = 0$ is rejected if and only if $\psi_W(\eta) = 1$.

## 3.3 Decorrelated Partial Likelihood Ratio Test

In low dimsional settings, the partial likelihood ratio test statistic is PLRT $= 2n\{\mathcal{L}(0, \widehat{\boldsymbol{\theta}}_P(0)) - \mathcal{L}(\widehat{\alpha}_P, \widehat{\boldsymbol{\theta}}_P)\}$, where $\widehat{\boldsymbol{\theta}}_P(0) = \operatorname{argmin}_{\boldsymbol{\theta}} \mathcal{L}(0, \boldsymbol{\theta})$ and $(\widehat{\alpha}_P, \widehat{\boldsymbol{\theta}}_P) = \operatorname{argmin}_{\alpha, \boldsymbol{\theta}} \mathcal{L}(\alpha, \boldsymbol{\theta})$ are the maximum partial likelihood estimators under the null and alternative, respectively. Hence, PLRT evaluates the validity of the null hypothesis by comparing the partial likelihood under $H_0$ with that under $H_1$. Similar to the partial score test, the partial likelihood ratio test also fails in the high dimensional setting due to the presence of a large number of nuisance parameters. In this section, we propose a new version of the partial likelihood ratio test which is valid in high dimensions.

To handle the impact of high dimensional nuisance parameters, we define the (negative) decorrelated partial likelihood for $\alpha$ as $\mathcal{L}_{\text{decor}}(\alpha) = \mathcal{L}(\alpha, \widehat{\boldsymbol{\theta}} - \alpha\widehat{\mathbf{w}})$. The reason for this name is that the derivative of $\mathcal{L}_{\text{decor}}(\alpha)$ with respect to $\alpha$ evaluated at $\alpha = 0$ is identical to the decorrelated score function $\widehat{U}(0, \widehat{\boldsymbol{\theta}})$ in (3.3). The decorrelated partial likelihood $\mathcal{L}_{\text{decor}}(\alpha)$ plays the same role as the profile partial likelihood $\mathcal{L}(\alpha, \widehat{\boldsymbol{\theta}}(\alpha))$ in the low dimensional setting. Hence, the decorrelated partial likelihood ratio test statistic is defined as

$$\widehat{L}_n = 2n\{\mathcal{L}_{\text{decor}}(0) - \mathcal{L}_{\text{decor}}(\widetilde{\alpha})\}, \quad \text{where} \quad \mathcal{L}_{\text{decor}}(\alpha) = \mathcal{L}(\alpha, \widehat{\boldsymbol{\theta}} - \alpha\widehat{\mathbf{w}}), \tag{3.10}$$

and $\widetilde{\alpha}$ is given in (3.7). As discussed in the previous subsection, $\widetilde{\alpha}$ is a one-step approximation of the global minimizer of $\mathcal{L}_{\text{decor}}(\alpha)$. Hence, the log-likelihood ratio $\widehat{L}_n$ evaluates the validity of the null hypothesis by comparing the decorrelated partial likelihood under $H_0$ with that under $H_1$. This is a natural extension of the classical partial likelihood ratio test to the high dimensional setting.

In the next section, we show that $\widehat{L}_n$ converges weakly to a chi-squared distribution with one degree of freedom. Therefore, a decorrelated partial likelihood ratio test with significance level $\eta$ is

$$\psi_L(\eta) = \begin{cases} 0 & \text{if } \widehat{L}_n \leq \chi_1^2(1-\eta) \\ 1 & \text{otherwise} \end{cases}, \tag{3.11}$$

and $\psi_L(\eta) = 1$ indicates a rejection of the null hypothesis.

## 4 Asymptotic Properties

In this section, we derive the limiting distributions of the decorrelated test statistics under the null hypothesis. More detailed proofs are provided in Appendix A. In our analysis, we make the following regularity assumptions.

**Assumption 4.1.** The true hazard is uniformly bounded, i.e., $\sup_{t\in[0,\tau]} \max_{i\in[n]} |\boldsymbol{X}_i^T(t)\boldsymbol{\beta}^*| = \mathcal{O}(1)$.

**Assumption 4.2.** It holds that $\|\mathbf{w}^*\|_0 = s' \asymp s$, and $\sup_{t\in[0,\tau]} \max_{i\in[n]} |\boldsymbol{X}_{i,2:d}^T(t)\mathbf{w}^*| = \mathcal{O}(1)$.

**Assumption 4.3.** The Fisher information matrix is bounded, $\|\mathbf{H}^*\|_\infty = \mathcal{O}(1)$, and its minimum eigenvalue is also bounded from below, $\Lambda_{\min}(\mathbf{H}^*) \geq C_h > 0$, which implies that $H_{\alpha|\boldsymbol{\theta}} = \mathbf{H}_{\alpha\alpha}^* - \mathbf{H}_{\alpha\boldsymbol{\theta}}^* \mathbf{H}_{\boldsymbol{\theta}\boldsymbol{\theta}}^{*-1} \mathbf{H}_{\boldsymbol{\theta}\alpha}^* \geq C_h$.

To connect these assumptions with existing literature, Assumptions 4.1 and 4.2 extend Assumption (iv) of Theorem 3.3 in van de Geer et al. (2014a) to the proportional hazards model. In particular, the sparsity assumption of $\mathbf{w}^*$ ensures that the Dantzig selector $\widehat{\mathbf{w}}$ converges to $\mathbf{w}^*$ at a fast rate. Assumption 4.3 is related to the Fisher information matrix, which is essential even in low dimensional settings.

Our main result characterizes the asymptotic normality of the decorrelated score function $\widehat{U}(0, \widehat{\boldsymbol{\theta}})$ in (3.3) under the null.

**Theorem 4.4.** Under Assumptions 2.1, 2.2, 4.1, 4.2 and 4.3, let $\lambda \asymp \sqrt{n^{-1}\log d}$, $\lambda' \asymp s\sqrt{n^{-1}\log d}$ and $n^{-1/2}s^3 \log d = o(1)$. Under the null hypothesis that $\alpha^* = 0$, the decorrelated score function $\widehat{U}(0, \widehat{\boldsymbol{\theta}})$ defined in (3.3) satisfies

$$\sqrt{n}\widehat{U}(0, \widehat{\boldsymbol{\theta}}) \xrightarrow{d} Z, \text{ where } Z \sim N(0, H_{\alpha|\boldsymbol{\theta}}), \tag{4.1}$$

and $H_{\alpha|\boldsymbol{\theta}} = \mathbf{H}_{\alpha\alpha}^* - \mathbf{H}_{\alpha\boldsymbol{\theta}}^* \mathbf{H}_{\boldsymbol{\theta}\boldsymbol{\theta}}^{*-1} \mathbf{H}_{\boldsymbol{\theta}\alpha}^*$.

As we have discussed before, the limiting variance of the decorrelated score function can be estimated by $\widehat{H}_{\alpha|\boldsymbol{\theta}} = \nabla^2_{\alpha\alpha}\mathcal{L}(\widehat{\alpha}, \widehat{\boldsymbol{\theta}}) - \widehat{\mathbf{w}}^T \nabla^2_{\boldsymbol{\theta}\alpha}\mathcal{L}(\widehat{\alpha}, \widehat{\boldsymbol{\theta}})$. The next lemma shows the consistency of $\widehat{H}_{\alpha|\boldsymbol{\theta}}$.

**Lemma 4.5.** Suppose Assumptions 2.1, 2.2, 4.1, 4.2 and 4.3 hold. If $\lambda \asymp \sqrt{n^{-1}\log d}$ and $\lambda' \asymp s\sqrt{n^{-1}\log d}$, we have

$$|H_{\alpha|\boldsymbol{\theta}} - \widehat{H}_{\alpha|\boldsymbol{\theta}}| = \mathcal{O}_{\mathbb{P}}\left(s^2 \sqrt{\frac{\log d}{n}}\right),$$

where $\widehat{H}_{\alpha|\boldsymbol{\theta}}$ is defined in (3.4).

By Theorem 4.4 and Lemma 4.5, the next corollary shows that under the null hypothesis, type I error of the decorrelated score test $\psi_S(\eta)$ in (3.6) converges asymptotically to the significance level $\eta$. Let the associated $p$-value of the decorrelated score test be $P_S = 2\{1 - \Phi(\widehat{S}_n)\}$, where $\Phi(\cdot)$ is the cumulative distribution function of the standard normal random variable and $\widehat{S}_n$ is the score test statistic defined in (3.5). The distribution of $P_S$ converges to a uniform distribution asymptotically.

**Corollary 4.6.** Suppose Assumptions 2.1, 2.2, 4.1, 4.2 and 4.3 hold, $\lambda \asymp \sqrt{n^{-1}\log d}$, $\lambda' \asymp s\sqrt{n^{-1}\log d}$, and $n^{-1/2}s^3 \log d = o(1)$. The decorrelated score test and the its corresponding $p$-value satisfy

$$\lim_{n\to\infty} \mathbb{P}(\psi_S(\eta) = 1 | \alpha^* = 0) = \eta, \text{ and } P_S \xrightarrow{d} \text{Unif}[0,1], \text{ when } \alpha^* = 0,$$

where Unif$[0,1]$ denotes a random variable uniformly distributed in $[0,1]$.

We then analyze the decorrelated Wald test under the null. We derive the limiting distribution of the one-step estimator $\widetilde{\alpha}$ defined in (3.7) in the next theorem.

**Theorem 4.7.** Suppose Assumptions 2.1, 2.2, 4.1, 4.2 and 4.3 hold, and $\lambda \asymp \sqrt{n^{-1}\log d}$, $\lambda' \asymp s\sqrt{n^{-1}\log d}$, $n^{-1/2}s^3 \log d = o(1)$. When the null hypothesis $\alpha^* = 0$ holds, the decorrelated estimator $\widetilde{\alpha}$ satisfies

$$\sqrt{n}\widetilde{\alpha} \xrightarrow{d} Z, \text{ where } Z \sim N(0, H_{\alpha|\boldsymbol{\theta}}^{-1}). \tag{4.2}$$

Utilizing the asymptotic normality of $\widetilde{\alpha}$, we can establish the limiting type I error of $\psi_W(\eta)$ in (3.9), in the next corollary. Note that, it is straightforward to generalize the result to be $\sqrt{n}(\widetilde{\alpha} - \alpha^*) \xrightarrow{d} Z$, where $Z \sim N(0, H_{\alpha|\boldsymbol{\theta}}^{-1})$ for any $\alpha^*$. This gives us a confidence interval of $\alpha^*$.

**Corollary 4.8.** Under Assumptions 2.1, 2.2, 4.1, 4.2 and 4.3, suppose $\lambda \asymp \sqrt{n^{-1}\log d}$, $\lambda' \asymp s\sqrt{n^{-1}\log d}$ and $n^{-1/2}s^3 \log d = o(1)$. The type I error of the decorrelated Wald test $\psi_W(\eta)$ and its corresponding $p$-value $P_W = 2\{1 - \Phi(\widehat{W}_n)\}$ satisfy

$$\lim_{n \to} \mathbb{P}(\psi_W(\eta) = 1 | \alpha^* = 0) = \eta, \text{ and } P_W \xrightarrow{d} \text{Unif}[0,1] \text{ when } \alpha^* = 0.$$

In addition, an asymptotic $(1-\eta) \times 100\%$ confidence interval of $\alpha^*$ is

$$\left(\widetilde{\alpha} - \frac{\Phi^{-1}(1-\eta/2)}{\sqrt{n\widehat{H}_{\alpha|\boldsymbol{\theta}}}}, \widetilde{\alpha} + \frac{\Phi^{-1}(1-\eta/2)}{\sqrt{n\widehat{H}_{\alpha|\boldsymbol{\theta}}}}\right).$$

Finally, we characterize the limiting distribution of the decorrelated partial likelihood ratio test statistic $\widehat{L}_n$ introduced in (3.10).

**Theorem 4.9.** Suppose Assumptions 2.1, 2.2, 4.1, 4.2 and 4.3 hold, $\lambda \asymp \sqrt{n^{-1}\log d}$, $\lambda' \asymp s\sqrt{n^{-1}\log d}$ and $n^{-1/2}s^3 \log d = o(1)$. If the null hypothesis $\alpha^* = 0$ holds, the decorrelated likelihood ratio test statistic $\widehat{L}_n$ in (3.10) satisfies

$$\widehat{L}_n \xrightarrow{d} Z_\chi, \text{ where } Z_\chi \sim \chi_1^2. \tag{4.3}$$

This theorem justifies the decorrelated partial likelihood ratio test $\psi_L(\eta)$ in (3.11). Also, let the $p$-value associated with the decorrelated partial likelihood ratio test be $P_L = 1 - F(\widehat{L}_n)$, where $F(\cdot)$ is the cumulative distribution function of $\chi_1^2$. Similar to Corollaries 4.6 and 4.8, we characterize the type I error of the test $\psi_L(\eta)$ in (3.11) and its corresponding $p$-value below.

**Corollary 4.10.** Suppose Assumptions 2.1, 2.2, 4.1, 4.2 and 4.3 hold, $\lambda \asymp \sqrt{n^{-1}\log d}$, $\lambda' \asymp s\sqrt{n^{-1}\log d}$ and $n^{-1/2}s^3 \log d = o(1)$. The type I error of the decorrelated partial likelihood ratio test $\psi_L(\eta)$ with significance level $\eta$ and its associated $p$-value $P_L$ satisfy

$$\lim_{n \to \infty} \mathbb{P}(\psi_L(\eta) = 1 | \alpha^* = 0) = \eta, \text{ and } P_L \xrightarrow{d} \text{Unif}[0,1] \text{ when } \alpha^* = 0.$$

By Corollaries 4.6, 4.8 and 4.10, we see that the decorrelated score, Wald and partial likelihood ratio tests are asymptotically equivalent as summarized in the next corollary.

**Corollary 4.11.** Suppose Assumptions 2.1, 2.2, 4.1, 4.2 and 4.3 hold, $\lambda \asymp \sqrt{n^{-1} \log d}$, $\lambda' \asymp s\sqrt{n^{-1} \log d}$ and $n^{-1/2} s^3 \log d = o(1)$. If the null hypothesis $\alpha^* = 0$ holds, the test statistics $\widehat{S}_n$ in (3.5), $\widehat{W}_n$ in (3.8), and $\widehat{L}_n$ in (3.10) are asymptotically equivalent, i.e.,

$$\widehat{S}_n = \widehat{W}_n + o_\mathbb{P}(1) = \widehat{L}_n + o_\mathbb{P}(1).$$

To summarize this subsection, Corollaries 4.6, 4.8 and 4.10 characterize the asymptotic distributions of the proposed decorrelated test statistics under the scaling when $n^{-1/2} s^3 \log d = o(1)$ under the null hypothesis. It is known that $H_{\alpha|\theta}$ is the semiparametric information lower bound for inferring $\alpha$. Theorem 4.7 shows that $\widetilde{\alpha}$ achieves the semiparametric information bound, which indicates the semiparametric efficiency of $\widetilde{\alpha}$. Using the asymptotic equivalence in Corollary 4.11, all of our test statistics are semiparametrically efficient (van der Vaart, 2000).

**Remark 4.12.** All the theoretical results in this section are still valid if we replace the Lasso penalty with nonconvex SCAD or MCP penalties as long as the consistency result (2.2) holds.

**Remark 4.13.** When the model is misspecified, we denote the oracle parameter as

$$\boldsymbol{\beta}^o = \underset{\boldsymbol{\beta}}{\operatorname{argmin}}\, \mathbb{E}^* \{\mathcal{L}(\boldsymbol{\beta})\},$$

where $\mathbb{E}^*$ is the expectation under the true model. Our proposed methods are still applicable to test if $\beta_1^o = 0$ and construct confidence intervals for $\beta_1^o$.

**Remark 4.14.** Existing works mainly consider high dimensional inferences for linear and generalized models; see Lockhart et al. (2014); Chernozhukov et al. (2013); van de Geer et al. (2014b); Javanmard and Montanari (2013) and Zhang and Zhang (2014). More specifically, Lockhart et al. (2014) consider conditional inference, while we consider unconditional inference. The others propose estimators that are asymptotically normal. Compared with existing approaches, we provide a unified framework which are more general in two aspects: (i) Our framework can deal with nonconvex penalties, while it is unclear if existing works are still valid under nonconvex penalities. (ii) Our framework based on the decorrelated score function provides a natural approach to deal with the misspecified model. In contrast, most existing methods assume the model must be correct.

## 5 Inference on the Baseline Hazard Function

The baseline hazard function

$$\Lambda_0(t) = \int_0^t \lambda_0(u) du$$

is treated as a nuisance function in the log-partial likelihood method. In practice, inferences on the baseline hazard function is also of interest. To the best of our knowledge, estimating the baseline hazard function or the survival function and construct confidence intervals in high dimensions remains unexplored. In this section, we extend the decorrelation approach to construct confidence intervals for the baseline hazard function and the survival function. All the proof details are provided in Appendix B.

We consider the following Breslow-type estimator for the baseline hazard function. Given an $\ell_1$-penalized estimator $\widehat{\boldsymbol{\beta}}$ derived from (2.1), the direct plug-in estimator for the baseline hazard function at time $t$ is

$$\widehat{\Lambda}_0(t, \widehat{\boldsymbol{\beta}}) = \int_0^t \frac{\sum_{i=1}^n dN_i(u)}{\sum_{i=1}^n Y_i(u) \exp\{\boldsymbol{X}_i^T(u)\widehat{\boldsymbol{\beta}}\}}. \tag{5.1}$$

Since the plug-in estimator $\widehat{\boldsymbol{\beta}}$ does not posses a tractable distribution, inference based on the estimator $\widehat{\Lambda}_0(t, \widehat{\boldsymbol{\beta}})$ is difficult. To handle this problem, we adopt the decorrelation approach as in the previous sections and estimate $\Lambda_0(t)$ by the sample version of $\widehat{\Lambda}_0(t, \widehat{\boldsymbol{\beta}}) - \{\nabla \Lambda_0(t, \boldsymbol{\beta}^*)\}^T \mathbf{H}^{*-1} \nabla \mathcal{L}(\widehat{\boldsymbol{\beta}})$, where

$$\Lambda_0(t, \boldsymbol{\beta}) = \mathbb{E} \int_0^t \frac{dN_i(u)}{S^{(0)}(u, \boldsymbol{\beta})},$$

and the gradient $\nabla \Lambda_0(t, \boldsymbol{\beta}^*)$ is taken with respect to the corresponding $\boldsymbol{\beta}$ component, and $\mathbf{H}^*$ is the Fisher information matrix defined in (2.7). Similar to Section 3.1, we directly estimate $\mathbf{H}^{*-1} \nabla \widehat{\Lambda}_0(t, \widehat{\boldsymbol{\beta}})$ by the following Dantzig selector

$$\widehat{\mathbf{u}}(t) = \operatorname{argmin} \|\mathbf{u}(t)\|_1, \text{ subject to } \|\nabla \widehat{\Lambda}_0(t, \widehat{\boldsymbol{\beta}}) - \nabla^2 \mathcal{L}(\widehat{\boldsymbol{\beta}}) \mathbf{u}(t)\|_\infty \leq \delta, \tag{5.2}$$

where $\delta$ is a tuning parameter. It can be shown that the estimator $\widehat{\mathbf{u}}(t)$ converges to $\mathbf{u}^*(t) = \mathbf{H}^{*-1} \nabla \Lambda_0(t, \boldsymbol{\beta}^*)$ under the following regularity assumption.

**Assumption 5.1.** It holds that $\|\mathbf{u}^*(t)\|_0 = s' \asymp s$ for all $0 \leq t \leq \tau$.

Note that Assumption 5.1 plays the same role as Assumption 4.2 in the previous section. Corollary B.2 in Appendix B characterizes the rate of convergence of $\widehat{\mathbf{u}}(t)$. Hence, the decorrelated baseline hazard function estimator at time $t$ is

$$\widetilde{\Lambda}_0(t, \widehat{\boldsymbol{\beta}}) = \widehat{\Lambda}_0(t, \widehat{\boldsymbol{\beta}}) - \widehat{\mathbf{u}}(t)^T \nabla \mathcal{L}(\widehat{\boldsymbol{\beta}}), \text{ where } \widehat{\mathbf{u}}(t) \text{ is defined in (5.2)}. \tag{5.3}$$

Based on the estimator (5.3), the survival function $S_0(t) = \exp\{-\Lambda_0(t)\}$ is estimated by $\widetilde{S}(t, \widehat{\boldsymbol{\beta}}) = \exp\{-\widetilde{\Lambda}_0(t, \widehat{\boldsymbol{\beta}})\}$. The main theorem of this section characterizes the asymptotic normality of $\widetilde{\Lambda}_0(t, \widehat{\boldsymbol{\beta}})$ and $\widetilde{S}(t, \widehat{\boldsymbol{\beta}})$ as follows.

**Theorem 5.2.** Suppose Assumptions 2.1, 2.2, 4.1, 4.3 and 5.1 hold, $\lambda \asymp \sqrt{n^{-1} \log d}$, $\delta \asymp s'\sqrt{n^{-1} \log d}$ and $n^{-1/2} s^3 \log d = o(1)$. We have, for any $t \in [0, \tau]$, the decorrelated baseline hazard function estimator $\widetilde{\Lambda}_0(t, \widehat{\boldsymbol{\beta}})$ in (5.3) satisfies

$$\sqrt{n}\{\Lambda_0(t) - \widetilde{\Lambda}_0(t, \widehat{\boldsymbol{\beta}})\} \xrightarrow{d} Z, \text{ where } Z \sim N(0, \sigma_1^2(t) + \sigma_2^2(t)),$$

and

$$\sigma_1^2(t) = \int_0^t \frac{\lambda_0(u) du}{\mathbb{E}\left[\exp\{\boldsymbol{X}^T(u) \boldsymbol{\beta}^*\} Y(u)\right]} \text{ and } \sigma_2^2(t) = \nabla \Lambda_0(t, \boldsymbol{\beta}^*)^T \mathbf{H}^{*-1} \nabla \Lambda_0(t, \boldsymbol{\beta}^*). \tag{5.4}$$

The estimated survival function $\widetilde{S}(t, \widehat{\boldsymbol{\beta}})$ satisfies

$$\sqrt{n}\{\widetilde{S}(t, \widehat{\boldsymbol{\beta}}) - S_0(t)\} \xrightarrow{d} Z', \text{ where } Z' \sim N\left(0, \frac{\sigma_1^2(t) + \sigma_2^2(t)}{\exp(2\Lambda_0(t))}\right).$$

Given Theorem 5.2, we further need to estimate the limiting variances $\sigma_1^2(t)$ and $\sigma_2^2(t)$. To this end, we use

$$\widehat{\sigma}_1^2(t) = \int_0^t \frac{d\widehat{\Lambda}_0(u, \widehat{\boldsymbol{\beta}})}{n^{-1}\sum_{i'=1}^n \exp\{\boldsymbol{X}_{i'}^T(u)\widehat{\boldsymbol{\beta}}\}Y_{i'}(u)} \quad \text{and} \quad \widehat{\sigma}_2^2(t) = \{\nabla\widehat{\Lambda}_0(t, \widehat{\boldsymbol{\beta}})\}^T \widehat{\mathbf{u}}(t),$$

where $\widehat{\Lambda}_0(t, \widehat{\boldsymbol{\beta}})$ is defined in (5.1).

We conclude this section by the following corollary which provides confidence intervals for $\Lambda_0(t)$ and $S_0(t)$.

**Corollary 5.3.** Suppose Assumptions 2.1, 2.2, 4.2, 4.3 and 5.1 hold, $\lambda \asymp \sqrt{n^{-1}\log d}$, $\delta \asymp s\sqrt{n^{-1}\log d}$ and $n^{-1/2}s^3 \log d = o(1)$. For any $t > 0$ and $0 < \eta < 1$,

$$\lim_{n\to\infty} \mathbb{P}\left(|\Lambda_0(t) - \widetilde{\Lambda}_0(t, \widehat{\boldsymbol{\beta}})| \leq \frac{\Phi^{-1}(1-\eta/2)\{\widehat{\sigma}_1^2(t) + \widehat{\sigma}_2^2(t)\}^{1/2}}{\sqrt{n}}\right) = 1 - \eta,$$

and

$$\lim_{n\to\infty} \mathbb{P}\left(|S_0(t) - \widetilde{S}_0(t, \widehat{\boldsymbol{\beta}})| \leq \frac{\Phi^{-1}(1-\eta/2)\{\widehat{\sigma}_1^2(t) + \widehat{\sigma}_2^2(t)\}^{1/2} \exp\{-\widetilde{\Lambda}_0(t, \widehat{\boldsymbol{\beta}})\}}{\sqrt{n}}\right) = 1 - \eta.$$

# 6 Numerical Results

This section reports numerical results of our proposed methods using both simulated and real data. We test the methods proposed in Section 3 and Section 5 by considering empirical behaviors for inferences on the individual regression coefficients $\beta_j$'s and the baseline hazard function $\Lambda_0(t)$.

## 6.1 Inference on the Parametric Component

We first investigate empirical performances of the decorrelated score, Wald and partial likelihood ratio tests on the parametric component $\boldsymbol{\beta}$ as proposed in Section 3. To estimate $\boldsymbol{\beta}^*$ and $\mathbf{w}^*$, we choose the tuning parameters $\lambda$ by 10-fold cross-validation and set $\lambda' = \frac{1}{2}\sqrt{n^{-1}\log d}$. We find that our simulation results are insensitive to the choice of $\lambda'$. We conduct decorrelated score, Wald and partial likelihood ratio tests for $\beta_1$ which is set to be 0 under null hypothesis $H_0$: $\beta_1 = 0$ versus alternative $H_a$: $\beta_1 \neq 0$, where we set the significance level to be $\eta = 0.05$. In each setting, we simulate $n = 150$ independent samples from a multivariate Gaussian distribution $N_d(\mathbf{0}, \boldsymbol{\Sigma})$ for $d = 100, 200$, or $500$, where $\boldsymbol{\Sigma}$ is a Toeplitz matrix with $\Sigma_{jk} = \rho^{|j-k|}$ and $\rho = 0.25, 0.4, 0.6$ or $0.75$. The cardinality of the active set $s$ is either 2 or 3, and the regression coefficients in the active set are either all 1's (Dirac) or drawn randomly from the uniform distribution Unif$[0, 2]$. We set the baseline hazard rate function to be identity. Thus, the $i$-th survival time follows an exponential distribution with mean $\exp(\boldsymbol{X}_i^T \boldsymbol{\beta}^*)$. The $i$-th censoring time is independently generated from an exponential distribution with mean $U \times \exp(\boldsymbol{X}_i^T \boldsymbol{\beta}^*)$, where $U \sim$ Unif$[1, 3]$. As discussed in Fan and Li (2002), this censoring scheme results in about 30% censored samples.

The above simulation is repeated 1,000 times. The empirical type I errors of the decorrelated score, Wald and partial likelihood ratio tests are summarized in Tables 1 and 2. We see that the empirical type I errors of all three tests are close to the desired 5% significance level, which supports our theoretical results. This observation holds for the whole range of $\rho$, $s$ and $d$ specified in the data generating procedures. In addition, as expected, the empirical type I errors further deviate from the significance level as $d$ increases for all three tests, illustrating the effects of dimensionality $d$ on finite sample performance.

Table 1: Average Type I error of the decorrelated tests with $\eta = 5\%$ where $(n, s) = (150, 2)$.

| Method | $d$ | $\rho = 0.25$ | | $\rho = 0.4$ | | $\rho = 0.6$ | | $\rho = 0.75$ | |
|---|---|---|---|---|---|---|---|---|---|
| | | Dirac | Unif[0, 2] | Dirac | Unif[0, 2] | Dirac | Unif[0, 2] | Dirac | Unif[0, 2] |
| Score | 100 | 5.1% | 5.2% | 5.1% | 4.9% | 5.2% | 5.1% | 4.9% | 5.0% |
| | 200 | 5.2% | 4.8% | 5.3% | 4.8% | 5.3% | 5.6% | 4.7% | 4.6% |
| | 500 | 6.1% | 6.4% | 5.5% | 4.6% | 4.2% | 4.4% | 3.9% | 3.7% |
| Wald | 100 | 5.2% | 5.3% | 5.1% | 5.0% | 5.2% | 4.9% | 5.0% | 5.1% |
| | 200 | 5.4% | 4.7% | 5.3% | 4.8% | 4.6% | 4.7% | 4.3% | 4.6% |
| | 500 | 6.3% | 6.1% | 5.9% | 5.5% | 5.8% | 4.2% | 4.5% | 3.9% |
| PLRT | 100 | 4.9% | 4.8% | 5.1% | 5.2% | 5.0% | 5.2% | 4.8% | 4.7% |
| | 200 | 5.7% | 5.5% | 5.3% | 5.5% | 4.8% | 5.6% | 4.6% | 4.5% |
| | 500 | 6.2% | 6.2% | 5.9% | 5.3% | 4.5% | 4.2% | 3.8% | 3.6% |

Table 2: Average type I error of the decorrelated tests with $\eta = 5\%$ where $(n, s) = (150, 3)$.

| | $d$ | $\rho = 0.25$ | | $\rho = 0.4$ | | $\rho = 0.6$ | | $\rho = 0.75$ | |
|---|---|---|---|---|---|---|---|---|---|
| | | Dirac | Unif[0, 2] | Dirac | Unif[0, 2] | Dirac | Unif[0, 2] | Dirac | Unif[0, 2] |
| Score | 100 | 5.2% | 5.2% | 4.8% | 5.3% | 5.3% | 4.9% | 5.3% | 4.8% |
| | 200 | 5.2% | 4.6% | 4.7% | 5.3% | 5.4% | 5.8% | 4.5% | 4.8% |
| | 500 | 6.3% | 6.5% | 5.8% | 4.4% | 5.2% | 4.6% | 3.6% | 3.4% |
| Wald | 100 | 5.1% | 4.9% | 5.3% | 4.7% | 5.2% | 4.9% | 5.0% | 5.1% |
| | 200 | 4.8% | 4.6% | 4.9% | 5.1% | 5.2% | 5.7% | 4.2% | 4.4% |
| | 500 | 6.5% | 6.8% | 6.2% | 5.9% | 5.1% | 4.5% | 3.9% | 4.2% |
| PLRT | 100 | 5.3% | 5.2% | 5.0% | 5.3% | 5.4% | 5.2% | 4.9% | 4.8% |
| | 200 | 5.5% | 5.3% | 5.4% | 4.6% | 5.2% | 5.7% | 5.4% | 4.3% |
| | 500 | 6.5% | 6.3% | 5.7% | 5.5% | 4.8% | 4.1% | 3.7% | 3.2% |

We also investigate the empirical power of the proposed tests. Instead of setting $\beta_1 = 0$, we generate the data with $\beta_1 = 0.05, 0.1, 0.15, ..., 0.55$, following the same simulation scheme introduced above. We plot the rejection rates of the three decorrelated tests for testing $H_0 : \beta_1 = 0$ with significance level 0.05 and $\rho = 0.25$ in Figure 2. We see that when $d = 100$, the three tests share similar power. However, for larger $d$ (e.g., $d = 500$), the decorrelated partial likelihood ratio test is the most powerful test. In addition, the Wald test is less effective for problems with higher dimensionality. Based on our simulation results, we recommend the decorrelated partial likelihood ratio test for inference in high dimensional problems.

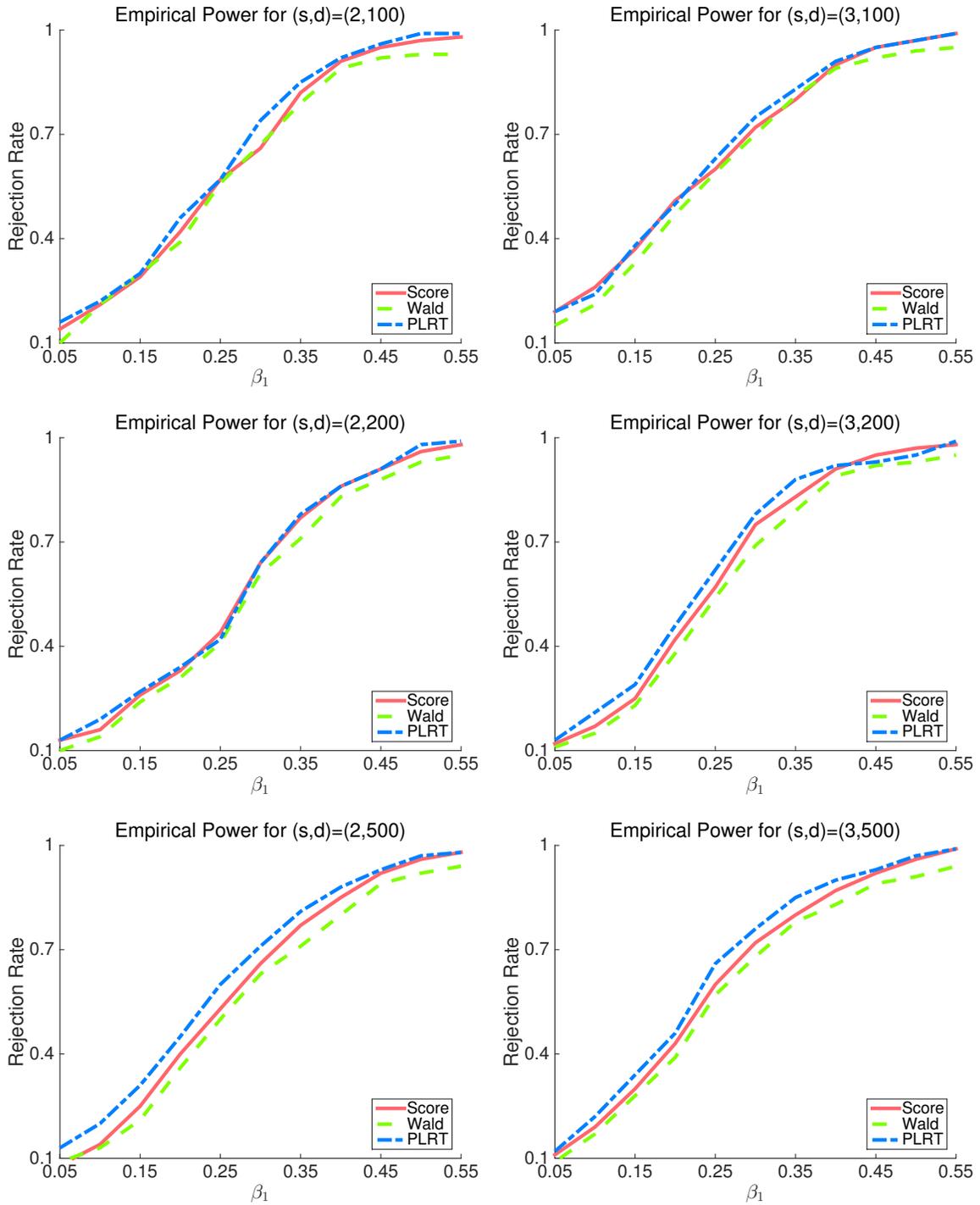

Figure 2: Empirical rejection rates of the decorrelated score, Wald and partial likelihood ratio tests on simulated data with different active set sizes and dimensionality.

## 6.2 Inference on the Baseline Hazard Function on Simulated Data

In this section, we demonstrate the empirical performance of the decorrelated inference procedure on the baseline hazard function $\Lambda_0(t)$ proposed as in Section 5. We consider three scenarios with $\Lambda_0(t) = t$, $t^2/2$ and $t^3/3$. Note that when $\Lambda_0(t) = p^{-1}t^p$, the survival time follows a Weibull distribution with shape parameter $p$ and scale parameter $\{p\exp(-\boldsymbol{X}_i^T\boldsymbol{\beta}^*)\}^{1/p}$, i.e., $W(p, \{p\exp(-\boldsymbol{X}_i^T\boldsymbol{\beta}^*)\}^{1/p})$. We use the same data generating procedures for the covariate $\boldsymbol{X}_i$'s, parameter $\boldsymbol{\beta}^*$ and censoring time $R$ as in the previous subsection.

In each simulation, we construct 95% confidence intervals for $\Lambda_0(t)$ at $t = 0.2$ using the procedures proposed in Section 5. The simulation is repeated 1,000 times. The results for the empirical coverage probabilities of $\Lambda_0(t)$ are summarized in Tables 3 and 4. It is seen that the coverage probabilities are all between 93% and 97%, which matches our theoretical results.

Table 3: Empirical coverage probability of 95% confidence intervals for $\Lambda_0(t)$ at $t = 0.2$ with $(n, s) = (150, 2)$

| $\Lambda_0(t)$ | $d$ | $\rho = 0.25$ | | $\rho = 0.4$ | | $\rho = 0.6$ | | $\rho = 0.75$ | |
|---|---|---|---|---|---|---|---|---|---|
| | | Dirac | Unif[0,2] | Dirac | Unif[0,2] | Dirac | Unif[0,2] | Dirac | Unif[0,2] |
| $t$ | 100 | 95.3% | 95.1% | 94.7% | 95.1% | 95.2% | 94.6% | 95.4% | 94.9% |
| | 200 | 95.5% | 95.8% | 95.7% | 95.3% | 94.6% | 94.5% | 94.4% | 94.2% |
| | 500 | 95.9% | 96.2% | 95.5% | 94.8% | 94.3% | 94.1% | 93.7% | 93.5% |
| $t^2$ | 100 | 95.1% | 95.3% | 95.2% | 95.0% | 95.4% | 94.7% | 95.2% | 95.3% |
| | 200 | 95.5% | 94.8% | 95.4% | 94.7% | 94.6% | 94.0% | 94.4% | 94.5% |
| | 500 | 96.6% | 96.7% | 96.1% | 95.4% | 94.9% | 94.3% | 93.8% | 93.6% |
| $t^3$ | 100 | 95.2% | 95.0% | 95.1% | 95.3% | 94.8% | 95.1% | 95.2% | 94.7% |
| | 200 | 95.4% | 94.7% | 94.6% | 95.5% | 95.2% | 95.8% | 94.6% | 94.3% |
| | 500 | 96.6% | 95.9% | 96.3% | 95.9% | 94.5% | 94.7% | 93.6% | 93.4% |

To further examine the performance of our method, we conduct additional simulation studies by plotting the 95% confidence intervals of $\Lambda_0(t)$ at $t = 0.05, 0.1, 0.15, ..., 0.5$, with $\Lambda_0(t) = t$ and $t^2/2$. The results are presented in Figures 3 and 4.

## 6.3 Analyzing a Gene Expression Dataset

We apply the proposed testing procedures to analyze a genomic data set, which is collected from a diffuse large B-cell lymphoma study analyzed by Alizadeh et al. (2000). One of the goals in this study is to investigate how the gene expression levels in B-cell malignancies are associated with the survival time. The expression values for over 13,412 genes in B-cell malignancies are measured by microarray experiments. The data setcontains 40 patients with diffuse large B-cell lymphoma who are recruited and followed until death or the end of the study. A small proportion ($\approx 5\%$) of the gene expression values are not well measured and are treated as missing values by Alizadeh et al. (2000). For simplicity, we impute the missing values of each gene by the median of the observed values of the same gene. The average survival time is 43.9 months and the censored rate is 55%.

Table 4: Empirical coverage probability of 95% confidence intervals for $\Lambda_0(t)$ at $t = 0.2$ with $(n, s) = (150, 3)$

| $\Lambda_0(t)$ | $d$ | $\rho = 0.25$ | | $\rho = 0.4$ | | $\rho = 0.6$ | | $\rho = 0.75$ | |
|---|---|---|---|---|---|---|---|---|---|
| | | Dirac | Unif[0,2] | Dirac | Unif[0,2] | Dirac | Unif[0,2] | Dirac | Unif[0,2] |
| $t$ | 100 | 95.1% | 94.8% | 94.8% | 95.2% | 95.3% | 95.1% | 94.8% | 95.4% |
| | 200 | 95.6% | 95.3% | 95.4% | 95.2% | 94.7% | 94.8% | 94.2% | 94.3% |
| | 500 | 96.2% | 95.9% | 95.8% | 96.1% | 95.2% | 94.3% | 93.3% | 93.6% |
| $t^2$ | 100 | 95.3% | 94.7% | 95.3% | 94.9% | 94.5% | 95.3% | 95.4% | 95.2% |
| | 200 | 94.7% | 94.5% | 95.4% | 95.2% | 94.1% | 94.9% | 94.3% | 93.8% |
| | 500 | 96.5% | 96.2% | 95.8% | 96.0% | 95.5% | 95.1% | 93.2% | 93.7% |
| $t^3$ | 100 | 95.0% | 95.2% | 94.6% | 94.8% | 95.1% | 95.4% | 94.9% | 95.5% |
| | 200 | 95.3% | 95.5% | 95.2% | 94.5% | 94.3% | 94.6% | 93.8% | 93.5% |
| | 500 | 95.9% | 96.3% | 95.7% | 96.0% | 95.4% | 94.7% | 93.6% | 93.1% |

Since the sample size $n = 40$ is small, we conduct pre-screening by fitting univariate proportional hazards models and only keep $d = 200$ genes with the smallest $p$-values.

We apply the proposed score, Wald and partial likelihood ratio tests to the pre-screened data. The same strategy for choosing the tuning parameters as that in the simulation studies is adopted. We repeatedly apply the hypothesis tests for all parameters. To control the family-wise error rate due to the multiple testing, the $p$-values are adjusted by the Bonferroni's method. To be more conservative, we only report the genes with adjusted $p$-values less than 0.05 by all of the three methods in Table 5. Many of the genes which are significant in the hypothesis tests are biologically related to lymphoma. For instance, the relation between lymphoma and genes FLT3 (Meierhoff et al., 1995), CDC10 (Di Gaetano et al., 2003), CHN2 (Nishiu et al., 2002) and Emv11 (Hiai et al., 2003) have been experimentally confirmed. This provides evidence that our methods can be used to discover scientific findings in applications involving high dimensional datasets.

Table 5: Genes with the adjuste $p$-values less than 0.05 using score, Wald and partial likelihood ratio tests for the large B-cell lymphoma gene expression dataset.

| Gene | Score | Wald | PLRT |
|---|---|---|---|
| FLT3 | $1.01 \times 10^{-2}$ | $2.86 \times 10^{-2}$ | $1.72 \times 10^{-2}$ |
| GPD2 | $3.91 \times 10^{-2}$ | $4.67 \times 10^{-3}$ | $7.44 \times 10^{-3}$ |
| PTMAP1 | $7.86 \times 10^{-3}$ | $4.84 \times 10^{-3}$ | $3.75 \times 10^{-3}$ |
| CDC10 | $3.52 \times 10^{-3}$ | $2.63 \times 10^{-3}$ | $1.10 \times 10^{-3}$ |
| Emv11 | $4.96 \times 10^{-3}$ | $2.77 \times 10^{-4}$ | $3.49 \times 10^{-4}$ |
| CHN2 | $1.79 \times 10^{-2}$ | $2.73 \times 10^{-2}$ | $3.58 \times 10^{-3}$ |
| Ptger2 | $1.78 \times 10^{-2}$ | $1.32 \times 10^{-2}$ | $2.47 \times 10^{-3}$ |
| Swq1 | $4.04 \times 10^{-3}$ | $4.21 \times 10^{-2}$ | $3.67 \times 10^{-2}$ |
| Cntn2 | $4.05 \times 10^{-3}$ | $4.84 \times 10^{-2}$ | $4.03 \times 10^{-2}$ |

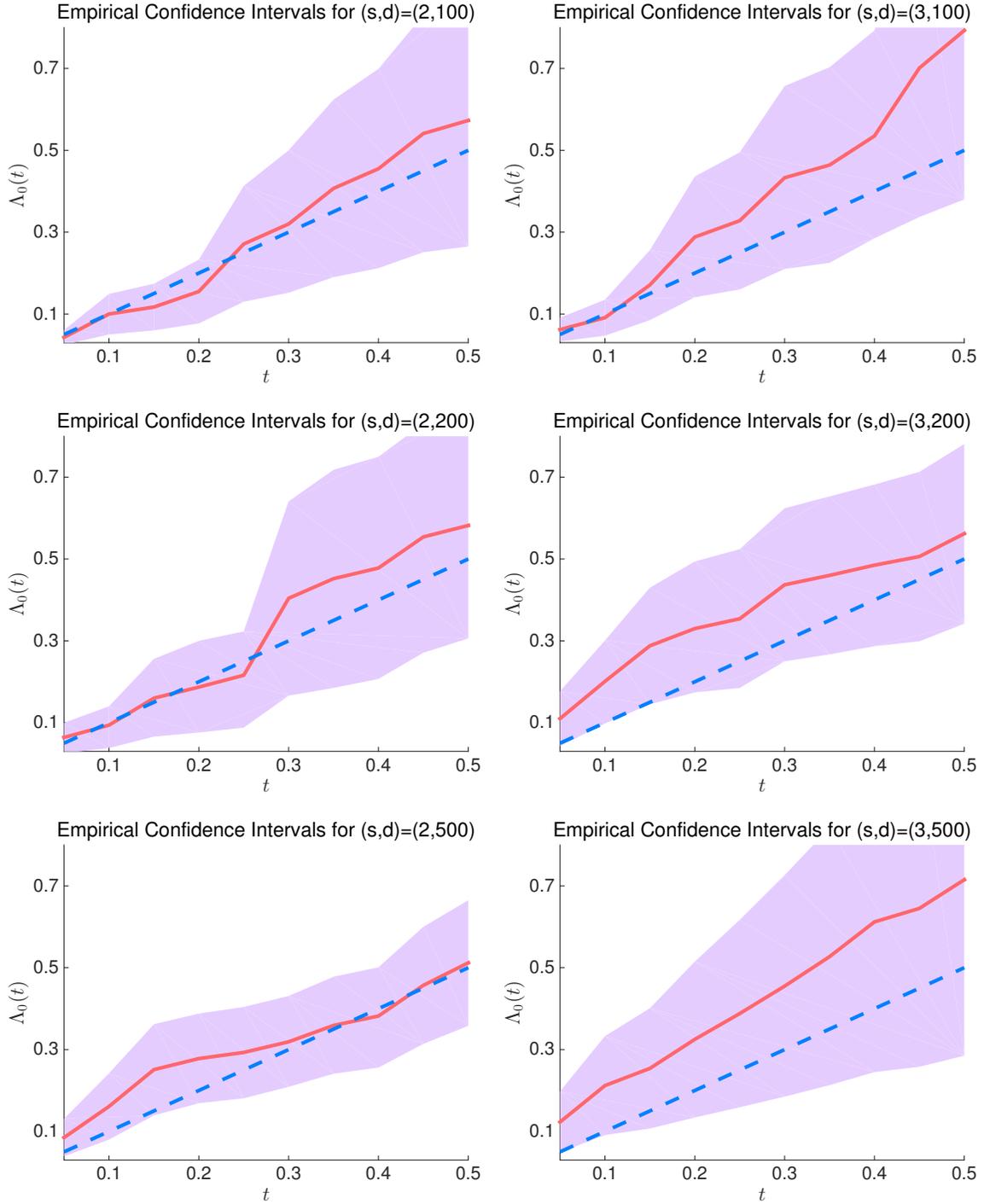

Figure 3: 95% confidence intervals for the baseline hazard function at $t = 0.05, 0.1, ..., 0.5$. The red solid line denotes the estimated baseline hazard function $\widetilde{\Lambda}(t)$, and blue dashed line denotes $\Lambda_0(t) = t$.

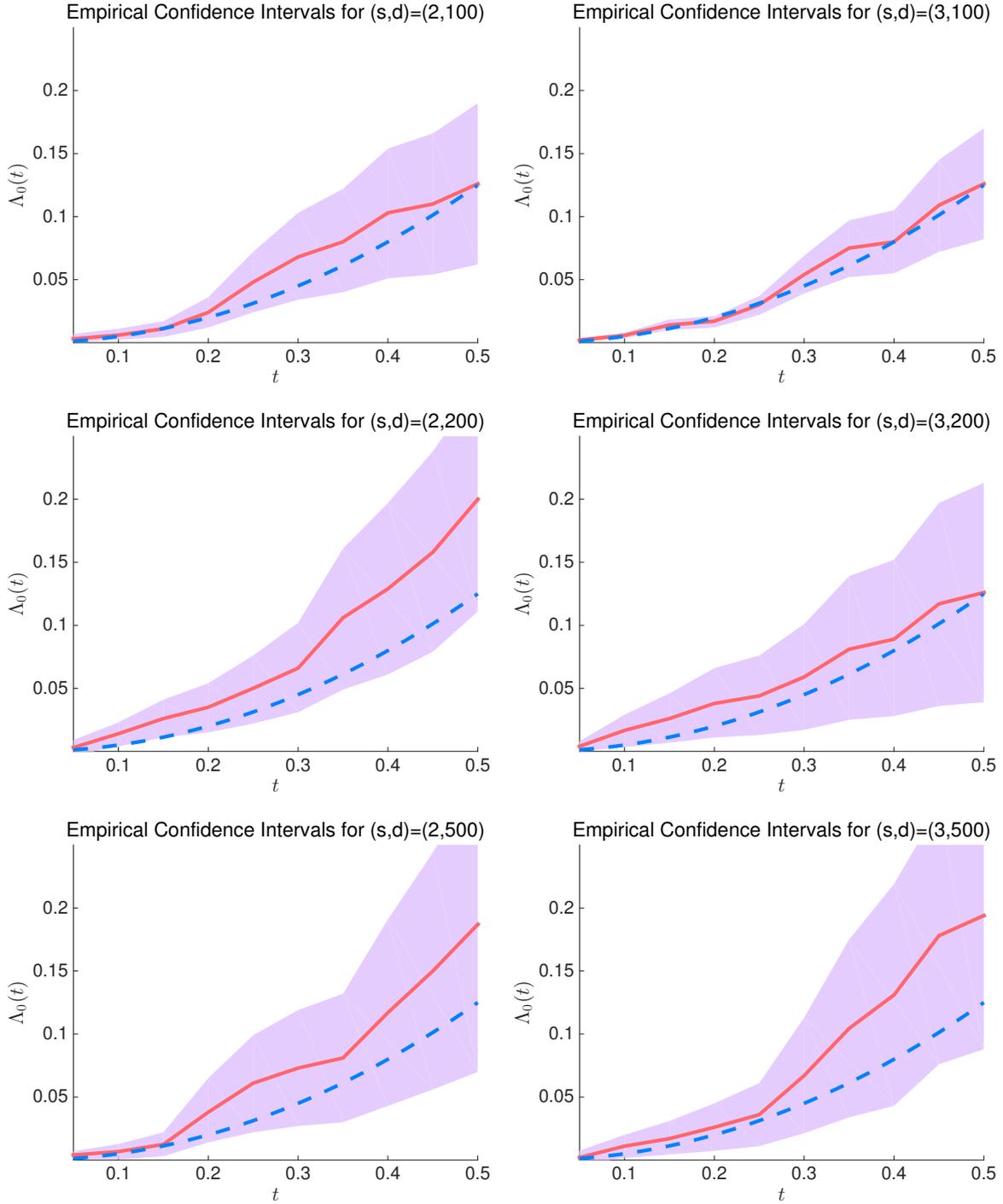

Figure 4: 95% confidence intervals for the baseline hazard function at $t = 0.05, 0.1, ..., 0.5$. The red solid line denotes the estimated baseline hazard function $\widetilde{\Lambda}(t)$, and the blue dashed line denotes $\Lambda_0(t) = t^2/2$.

# 7 Discussion

We proposed a novel decorrelation-based approach to conduct inference for both the parametric and nonparametric components of high dimensional Cox's proportional hazards models. Unlike existing works, our methods do not require conditions on model selection consistency or minimal signal strength. Theoretical properties of the proposed methods are established. Extensive numerical investigations are conducted on the simulated and real datasets to examine the finite sample performances of our methods. To the best of our knowledge, this paper for the first time provides a unified framework on uncertainty assessment of high dimensional Cox's proportional hazards models. Our methods can be extended to conduct inference for other high-dimensional survival models such as censored linear model (Müller and van de Geer, 2014) and additive hazards model (Lin and Lv, 2013).

In this paper, we focus on the Cox's proportional hazards model for the univariate survival data. In practice, many biomedical studies involve multiple survival outcomes. For instance, in the Framingham Heart Study by Dawber (1980), both time to coronary heart disease and time to cerebrovascular accident are observed. How the inference can be drawn by jointly analyzing the multivariate survival data in the high dimensional setting remains largely unexplored. To address this problem, we extend the proposed hypothesis testing procedures to deal with the multivariate survival data. More details are presented in Appendix D.

The proposed methods involve two tuning parameters $\lambda$ and $\lambda'$. The presence of multiple tuning parameters in the inferential procedures is encountered in many recent works even under high dimensional linear models (Chernozhukov et al., 2013; van de Geer et al., 2014b; Javanmard and Montanari, 2013; Zhang and Zhang, 2014). Theoretically, we establish the asymptotic normality of the test statistics when $\lambda \asymp \sqrt{n^{-1} \log d}$ and $\lambda' \asymp s\sqrt{n^{-1} \log d}$. Empirically, our numerical results suggest that cross-validation seems to be a practical procedure for the choice of $\lambda$. As an important future investigation, it is of interest to provide rigorous theoretical justification of practical procedures such as cross-validation for the choice of tuning parameters.

**Acknowledgement** We thank Professor Bradic for providing very helpful comments. This research is partially supported by the grants NSF CAREER DMS 1454377, NSF IIS1408910, NSF IIS1332109, NIH R01MH102339, NIH R01GM083084, and NIH R01HG06841.

# A  Proofs in Section 4

In this section, we provide the detailed proofs in Section 4. We first provide a key lemma which characterizes the asymptotic normality of $\nabla \mathcal{L}(\boldsymbol{\beta}^*)$. This lemma is essential in our later proofs to derive the asymptotic distributions of the test statistics.

**Lemma A.1.** Under Assumptions 2.1, 4.2 and 4.3, for any vector $\mathbf{v} \in \mathbb{R}^d$, if $\|\mathbf{v}\|_0 \leq s'$ and $n^{-1/2}\sqrt{s'^3 \log d} = o(1)$, it holds that

$$\frac{\sqrt{n}\mathbf{v}^T \nabla \mathcal{L}(\boldsymbol{\beta}^*)}{\sqrt{\mathbf{v}^T \mathbf{H}^* \mathbf{v}}} \xrightarrow{d} N(0,1), \quad \text{where } \mathbf{H}^* \text{ is defined in (2.7)}.$$

*Proof.* Let $M_i(t) = N_i(t) - \int_0^t Y_i(u)\lambda_0(u)du$. By the definition of $\nabla \mathcal{L}(\boldsymbol{\beta}^*)$ in (2.4), we have

$$\nabla \mathcal{L}(\boldsymbol{\beta}^*) = -\frac{1}{n}\sum_{i=1}^n \int_0^\tau \{\boldsymbol{X}_i(u) - \overline{\boldsymbol{Z}}_n(u, \boldsymbol{\beta}^*)\}dM_i(u)$$

$$= -\frac{1}{n}\sum_{i=1}^n \int_0^\tau \{\boldsymbol{X}_i(u) - \mathbf{e}(u, \boldsymbol{\beta}^*)\}dM_i(u) - \frac{1}{n}\sum_{i=1}^n \int_0^\tau \{\mathbf{e}(u, \boldsymbol{\beta}^*) - \overline{\boldsymbol{X}}_n(u, \boldsymbol{\beta}^*)\}dM_i(u),$$

Thus, by the identity $\mathbf{H}^* = \sqrt{n}\,\mathrm{Var}\{\nabla \mathcal{L}(\boldsymbol{\beta}^*)\}$, we have

$$\frac{\sqrt{n}\mathbf{v}^T \nabla \mathcal{L}(\boldsymbol{\beta}^*)}{\sqrt{\mathbf{v}^T \mathbf{H}^* \mathbf{v}}} = -\frac{1}{\sqrt{n}}\underbrace{\frac{\mathbf{v}^T}{\sqrt{\mathbf{v}^T \mathbf{H}^* \mathbf{v}}} \sum_{i=1}^n \int_0^\tau \{\boldsymbol{X}_i(u) - \mathbf{e}(u, \boldsymbol{\beta}^*)\}dM_i(u)}_{S}$$

$$-\underbrace{\frac{1}{\sqrt{n}}\frac{\mathbf{v}^T}{\sqrt{\mathbf{v}^T \mathbf{H}^* \mathbf{v}}} \sum_{i=1}^n \int_0^\tau \{\mathbf{e}(u, \boldsymbol{\beta}^*) - \overline{\boldsymbol{X}}_n(u, \boldsymbol{\beta}^*)\}dM_i(u)}_{E}.$$

For the first term $S$, denote by

$$\xi_i = \frac{\mathbf{v}^T}{\sqrt{\mathbf{v}^T \mathbf{H}^* \mathbf{v}}} \int_0^\tau \{\boldsymbol{X}_i(u) - \mathbf{e}(u, \boldsymbol{\beta}^*)\}dM_i(u).$$

We have $\mathbb{E}(\xi_i) = 0$, and $\mathrm{Var}(n^{-1/2}S) = 1$. Thus $S$ is a sum of $n$ independent random variables with mean 0. To get the asymptotic distribution of $n^{-1/2}S$, we verify the Lyapunov condition. Indeed, we have

$$\frac{1}{n^{3/2}}\sum_{i=1}^n \mathbb{E}\left|\frac{\mathbf{v}^T}{\sqrt{\mathbf{v}^T \mathbf{H}^* \mathbf{v}}} \int_0^\tau \{\boldsymbol{X}_i(u) - \mathbf{e}(u, \boldsymbol{\beta}^*)\}dM_i(u)\right|^3$$

$$\leq \frac{C}{C_h^{3/2} n^{3/2}} \sum_{i=1}^n s'^{3/2} \sup_{u \in [0,\tau]} \|\boldsymbol{X}_i(u) - \mathbf{e}(u, \boldsymbol{\beta}^*)\|_\infty^3$$

$$= \mathcal{O}(s'^{3/2} n^{-1/2}),$$

where the inequality follows by Assumption 4.3 for some constant $C$, and the equality holds by Lemma C.1 and Assumption 2.1. Thus, the Lyapunov condition holds by our scaling assumption that $s'^{3/2}n^{-1/2} = o(1)$. Apply Lindeberg Feller Central Limit Theorem, we have $n^{-1/2}S \xrightarrow{d} N(0,1)$.

Next, we prove that the second term $E = o_\mathbb{P}(1)$. Since

$$E = \frac{1}{\sqrt{n}}\frac{\mathbf{v}^T}{\sqrt{\mathbf{v}^T \mathbf{H}^* \mathbf{v}}} \sum_{i=1}^n \int_0^\tau \left[\{\mathbf{e}(u, \boldsymbol{\beta}^*) - \overline{\boldsymbol{X}}_n(u, \boldsymbol{\beta}^*)\}1 dM_i(u)\right]$$

$$\leq \frac{1}{\sqrt{n}}\frac{s'^{1/2}}{\lambda_{\min}} \sup_{u \in [0,\tau]} \|\mathbf{e}(u, \boldsymbol{\beta}^*) - \overline{\boldsymbol{X}}_n(u, \boldsymbol{\beta}^*)\|_\infty \int_0^\tau \left|\sum_{i=1}^n 1 dM_i(u)\right|.$$

By Lemma C.1, it holds that $\sup_{u \in [0,\tau]} \|\mathbf{e}(u, \boldsymbol{\beta}^*) - \overline{\boldsymbol{X}}_n(u, \boldsymbol{\beta}^*)\|_\infty = \mathcal{O}_\mathbb{P}(\sqrt{n^{-1}\log d})$. It holds that, for some constant $C > 0$,

$$E \leq \frac{C}{\sqrt{n}}\frac{1}{\lambda_{\min}}\sqrt{\frac{s'\log d}{n}} \int_0^\tau \left|\sum_{i=1}^n 1 dM_i(u)\right|.$$

It remains to bound the term $\int_0^\tau \left|\sum_{i=1}^n 1 dM_i(u)\right|$. By Theorem 2.11.9 and Example of 2.11.16 of van der Vaart and Wellner (1996), $\overline{G}(t) := n^{-1/2} \sum_{i=1}^n M_i(t)$ converges weakly to a tight Gaussian process $G(t)$. Furthermore, by Strong Embedding Theorem of Shorack and Wellner (2009), there exists another probability space such that $(S^{*(0)}(\boldsymbol{\beta}, t), S^{*(1)}(\boldsymbol{\beta}, t), \overline{G}^*(t))$ converges almost surely to $(s^{*(0)}(\boldsymbol{\beta}, t), \mathbf{s}^{*(1)}(\boldsymbol{\beta}, t), G^*(t))$, where $*$ indicates the existences in a new probability space. This implies that $\int_0^\tau |dG(t)| = \int_0^\tau |dG^*(t)| + o_\mathbb{P}(1)$. We have, by our assumption $n^{-1}\sqrt{s' \log d} = o_\mathbb{P}(1)$, the term $E$ satisfies that

$$E = \mathcal{O}_\mathbb{P}\left(\sqrt{\frac{s' \log d}{n}} \cdot \frac{1}{\sqrt{n}}\right) = o_\mathbb{P}(1).$$

Combining this with the result that $n^{-1/2} S \xrightarrow{d} N(0,1)$ concludes the proof. □

Next, we characterize the rate of convergence of the Dantzig selector $\widehat{\mathbf{w}}$ in (3.2) in the following lemma.

**Lemma A.2.** Under Assumptions 2.1, 2.2, 4.1, 4.2 and 4.3, If $\lambda' \asymp s'\sqrt{n^{-1} \log d}$, we have

$$\|\widehat{\mathbf{w}} - \mathbf{w}^*\|_1 = \mathcal{O}_\mathbb{P}(s's\sqrt{n^{-1} \log d}), \tag{A.1}$$

where $\widehat{\mathbf{w}}$ and $\mathbf{w}^*$ are defined in (3.2) and (3.1), respectively.

*Proof.* As shown in Lemma C.6, under Assumptions 2.1, 2.2, 4.1, 4.2 and 4.3, the condition (C.7) in Lemma C.8 is satisfied for $\lambda' \asymp s'\sqrt{n^{-1} \log d}$. Consequently, we have

$$\|\widehat{\mathbf{w}} - \mathbf{w}^*\|_1 = \mathcal{O}_\mathbb{P}(s's\sqrt{n^{-1} \log d}),$$

which concludes the proof. □

*Proof of Theorem 4.4.* To derive the asymptotic distribution of $\sqrt{n}\widehat{U}(0, \widehat{\boldsymbol{\theta}})$, we start with decomposing $\widehat{U}(0, \widehat{\boldsymbol{\theta}})$ into several terms.

$$\begin{aligned}
\widehat{U}(0, \widehat{\boldsymbol{\theta}}) &= \nabla_\alpha \mathcal{L}(0, \widehat{\boldsymbol{\theta}}) - \widehat{\mathbf{w}}^T \nabla_{\boldsymbol{\theta}} \mathcal{L}(0, \widehat{\boldsymbol{\theta}}) \\
&= \nabla_\alpha \mathcal{L}(0, \boldsymbol{\theta}^*) + \nabla^2_{\alpha\boldsymbol{\theta}} \mathcal{L}(0, \bar{\boldsymbol{\theta}})(\widehat{\boldsymbol{\theta}} - \boldsymbol{\theta}^*) - \left\{\widehat{\mathbf{w}}^T \nabla_{\boldsymbol{\theta}} \mathcal{L}(0, \boldsymbol{\theta}^*) + \widehat{\mathbf{w}}^T \nabla^2_{\boldsymbol{\theta}\boldsymbol{\theta}} \mathcal{L}(0, \widetilde{\boldsymbol{\theta}})(\widehat{\boldsymbol{\theta}} - \boldsymbol{\theta}^*)\right\} \\
&= \underbrace{\nabla_\alpha \mathcal{L}(0, \boldsymbol{\theta}^*) - \mathbf{w}^{*T} \nabla_{\boldsymbol{\theta}} \mathcal{L}(0, \boldsymbol{\theta}^*)}_{S} + \underbrace{(\mathbf{w}^* - \widehat{\mathbf{w}})^T \nabla_{\boldsymbol{\theta}} \mathcal{L}(0, \boldsymbol{\theta}^*)}_{E_1} + \underbrace{\left\{\nabla^2_{\alpha\boldsymbol{\theta}} \mathcal{L}(0, \bar{\boldsymbol{\theta}}) - \widehat{\mathbf{w}}^T \nabla^2_{\boldsymbol{\theta}\boldsymbol{\theta}} \mathcal{L}(0, \widetilde{\boldsymbol{\theta}})\right\}(\widehat{\boldsymbol{\theta}} - \boldsymbol{\theta}^*)}_{E_2},
\end{aligned} \tag{A.2}$$

where the second equality holds by the mean value theorem for some $\bar{\boldsymbol{\theta}} = \boldsymbol{\theta}^* + u(\widehat{\boldsymbol{\theta}} - \boldsymbol{\theta}^*)$, $\widetilde{\boldsymbol{\theta}} = \boldsymbol{\theta}^* + u'(\widehat{\boldsymbol{\theta}} - \boldsymbol{\theta}^*)$ and $u, u' \in [0,1]$.

We consider the terms $S$, $E_1$ and $E_2$ separately. For the first term $S$, by Lemma A.1, taking $\mathbf{v} = (1, -\mathbf{w}^{*T})^T$, we have

$$\sqrt{n} S \xrightarrow{d} Z, \text{ where } Z \sim N(0, H_{\alpha|\boldsymbol{\theta}}). \tag{A.3}$$

For the term $E_1$, we have,

$$E_1 \leq \|\widehat{\mathbf{w}} - \mathbf{w}^*\|_1 \|\nabla_{\boldsymbol{\theta}} \mathcal{L}(0, \boldsymbol{\theta}^*)\|_\infty = \mathcal{O}_\mathbb{P}(s'\lambda'\sqrt{n^{-1}\log d}), \tag{A.4}$$

where $\|\widehat{\mathbf{w}} - \mathbf{w}^*\|_1 = \mathcal{O}_\mathbb{P}(s'\lambda')$ by Lemma C.8, and $\|\nabla_{\boldsymbol{\theta}} \mathcal{L}(0, \boldsymbol{\theta}^*)\|_\infty = \mathcal{O}_\mathbb{P}(\sqrt{n^{-1}\log d})$ by Lemma C.3.

For the term $E_2$, we have,

$$E_2 = \underbrace{\{\nabla^2_{\alpha\boldsymbol{\theta}}\mathcal{L}(0,\bar{\boldsymbol{\theta}}) - \mathbf{H}^*_{\alpha\boldsymbol{\theta}}\mathbf{H}^{*-1}_{\boldsymbol{\theta\theta}}\nabla^2_{\boldsymbol{\theta\theta}}\mathcal{L}(0,\widetilde{\boldsymbol{\theta}})\}(\widehat{\boldsymbol{\theta}} - \boldsymbol{\theta}^*)}_{E_{21}} + \underbrace{(\mathbf{w}^* - \widehat{\mathbf{w}})^T \nabla^2_{\boldsymbol{\theta\theta}}\mathcal{L}(0,\widetilde{\boldsymbol{\theta}})(\widehat{\boldsymbol{\theta}} - \boldsymbol{\theta}^*)}_{E_{22}}. \quad (A.5)$$

Considering the terms $E_{21}$ and $E_{22}$ separately, first, we have,

$$\begin{aligned} E_{21} &= \{\nabla^2_{\alpha\boldsymbol{\theta}}\mathcal{L}(0,\bar{\boldsymbol{\theta}}) - \mathbf{H}^*_{\alpha\boldsymbol{\theta}} + \mathbf{H}^*_{\alpha\boldsymbol{\theta}} - \mathbf{H}^*_{\alpha\boldsymbol{\theta}}\mathbf{H}^{*-1}_{\boldsymbol{\theta\theta}}\nabla^2_{\boldsymbol{\theta\theta}}\mathcal{L}(0,\widetilde{\boldsymbol{\theta}})\}(\widehat{\boldsymbol{\theta}} - \boldsymbol{\theta}^*) \\ &\leq \|\nabla^2_{\alpha\boldsymbol{\theta}}\mathcal{L}(0,\bar{\boldsymbol{\theta}}) - \mathbf{H}^*_{\alpha\boldsymbol{\theta}}\|_\infty \|\widehat{\boldsymbol{\theta}} - \boldsymbol{\theta}^*\|_1 + |\mathbf{H}^*_{\alpha\boldsymbol{\theta}}(\mathbf{I}_{d-1} - \mathbf{H}^{*-1}_{\boldsymbol{\theta\theta}}\nabla^2_{\boldsymbol{\theta\theta}}\mathcal{L}(0,\widetilde{\boldsymbol{\theta}}))(\widehat{\boldsymbol{\theta}} - \boldsymbol{\theta}^*)|, \end{aligned} \quad (A.6)$$

where the inequality holds by Hölder's inequality. For the first term in the above inequality, we have

$$\|\nabla^2_{\alpha\boldsymbol{\theta}}\mathcal{L}(0,\bar{\boldsymbol{\theta}}) - \mathbf{H}^*_{\alpha\boldsymbol{\theta}}\|_\infty \|\widehat{\boldsymbol{\theta}} - \boldsymbol{\theta}^*\|_1 = \mathcal{O}_\mathbb{P}(s^2\lambda^2), \quad (A.7)$$

since $\|\widehat{\boldsymbol{\theta}} - \boldsymbol{\theta}^*\|_1 = \mathcal{O}_\mathbb{P}(s\lambda)$ by (2.2) and $\|\nabla_{\alpha\boldsymbol{\theta}}\mathcal{L}(0,\bar{\boldsymbol{\theta}}) - \mathbf{H}^*_{\alpha\boldsymbol{\theta}}\|_\infty = \mathcal{O}_\mathbb{P}(s\lambda)$ by Lemma C.5.

For the second term in (A.6), by Hölder's inequality, we have

$$\begin{aligned} |\mathbf{H}^*_{\alpha\boldsymbol{\theta}}(\mathbf{I}_{d-1} - \mathbf{H}^{*-1}_{\boldsymbol{\theta\theta}}\nabla^2_{\boldsymbol{\theta\theta}}\mathcal{L}(0,\widetilde{\boldsymbol{\theta}}))(\widehat{\boldsymbol{\theta}} - \boldsymbol{\theta}^*)| &= |\mathbf{H}^*_{\alpha\boldsymbol{\theta}}\mathbf{H}^{*-1}_{\boldsymbol{\theta\theta}}(\mathbf{H}^*_{\boldsymbol{\theta\theta}} - \nabla^2_{\boldsymbol{\theta\theta}}\mathcal{L}(0,\widetilde{\boldsymbol{\theta}}))(\widehat{\boldsymbol{\theta}} - \boldsymbol{\theta}^*)| \\ &\leq \|\mathbf{w}^*\|_1 \|\mathbf{H}^*_{\boldsymbol{\theta\theta}} - \nabla^2_{\boldsymbol{\theta\theta}}\mathcal{L}(0,\widetilde{\boldsymbol{\theta}})\|_\infty \|\widehat{\boldsymbol{\theta}} - \boldsymbol{\theta}^*\|_1 = \mathcal{O}_\mathbb{P}(s's^2\lambda^2), \end{aligned} \quad (A.8)$$

where the last equality holds since $\|\mathbf{w}^*\|_1 = \mathcal{O}(s')$ by Assumption 4.2, $\|\mathbf{H}^*_{\boldsymbol{\theta\theta}} - \nabla^2_{\boldsymbol{\theta\theta}}\mathcal{L}(0,\widetilde{\boldsymbol{\theta}})\|_\infty = \mathcal{O}_\mathbb{P}(s\lambda)$ by Lemma C.5, and $\|\widehat{\boldsymbol{\theta}} - \boldsymbol{\theta}^*\|_1 = \mathcal{O}_\mathbb{P}(s\lambda)$ by (2.2). Plugging (A.7) and (A.8) into (A.6), we have

$$|E_{21}| = \mathcal{O}_\mathbb{P}(s's^2\lambda^2). \quad (A.9)$$

For the second term $E_{22}$ in (A.5), we have,

$$|E_{22}| \leq \|\widehat{\mathbf{w}} - \mathbf{w}^*\|_1 \|\nabla^2_{\boldsymbol{\theta\theta}}\mathcal{L}(0,\widetilde{\boldsymbol{\theta}})\|_\infty \|\widehat{\boldsymbol{\theta}} - \boldsymbol{\theta}^*\|_1 = \mathcal{O}_\mathbb{P}(s's\lambda'\lambda), \quad (A.10)$$

where we use the results that $\|\widehat{\mathbf{w}} - \mathbf{w}^*\|_1 = \mathcal{O}_\mathbb{P}(s'\lambda')$ by Lemma C.8, $\|\widehat{\boldsymbol{\theta}} - \boldsymbol{\theta}^*\|_1 \leq \mathcal{O}_\mathbb{P}(s\lambda)$ by (2.2), and $\|\nabla^2_{\boldsymbol{\theta\theta}}\mathcal{L}(0,\widetilde{\boldsymbol{\theta}})\|_\infty = \mathcal{O}_\mathbb{P}(1)$ by Lemma C.5.

Plugging (A.6) and (A.10) into (A.5), we have $E_2 = \mathcal{O}_\mathbb{P}(n^{-1}s's^2\log d)$. Combining it with (A.4), we have

$$|E_1| + |E_2| = \mathcal{O}_\mathbb{P}\Big(\frac{s's^2\log d}{n}\Big) = o_\mathbb{P}\Big(\frac{1}{\sqrt{n}}\Big), \quad (A.11)$$

where the last equality holds by the assumption that $n^{-1/2}s^3\log d = o(1)$ and $s \asymp s'$. Combining (A.11), (A.3) and (A.2), our claim (4.1) holds as desired. $\square$

*Proof of Lemma 4.5.* By the definition of $H_{\alpha|\boldsymbol{\theta}}$ and $\widehat{H}_{\alpha|\boldsymbol{\theta}}$, we have

$$|H_{\alpha|\boldsymbol{\theta}} - \widehat{H}_{\alpha|\boldsymbol{\theta}}| \leq \underbrace{|\mathbf{H}^*_{\alpha\alpha} - \nabla^2_{\alpha\alpha}\mathcal{L}(\widehat{\alpha},\widehat{\boldsymbol{\theta}})|}_{E_1} + \underbrace{|\mathbf{H}^*_{\alpha\boldsymbol{\theta}}\mathbf{H}^{*-1}_{\boldsymbol{\theta\theta}}\mathbf{H}^*_{\boldsymbol{\theta}\alpha} - \widehat{\mathbf{w}}^T \nabla^2_{\boldsymbol{\theta}\alpha}\mathcal{L}(\widehat{\alpha},\widehat{\boldsymbol{\theta}})|}_{E_2}. \quad (A.12)$$

We consider the two terms separately. For the first term $E_1$, we have by Lemma C.5, $E_1 = \mathcal{O}_\mathbb{P}(s\lambda)$. For the second term $E_2$, we have,

$$\begin{aligned} E_2 &= |\mathbf{H}^*_{\alpha\boldsymbol{\theta}}\mathbf{H}^{*-1}_{\boldsymbol{\theta\theta}}\mathbf{H}^*_{\boldsymbol{\theta}\alpha} - \widehat{\mathbf{w}}^T \nabla^2_{\boldsymbol{\theta}\alpha}\mathcal{L}(\widehat{\alpha},\widehat{\boldsymbol{\theta}})| = |\mathbf{H}^*_{\alpha\boldsymbol{\theta}}\mathbf{H}^{*-1}_{\boldsymbol{\theta\theta}}\mathbf{H}^*_{\boldsymbol{\theta}\alpha} - \widehat{\mathbf{w}}^T \mathbf{H}^*_{\boldsymbol{\theta}\alpha} + \widehat{\mathbf{w}}^T \mathbf{H}^*_{\boldsymbol{\theta}\alpha} - \widehat{\mathbf{w}}^T \nabla^2_{\boldsymbol{\theta}\alpha}\mathcal{L}(\widehat{\alpha},\widehat{\boldsymbol{\theta}})| \\ &\leq \underbrace{|\mathbf{H}^*_{\alpha\boldsymbol{\theta}}\mathbf{H}^{*-1}_{\boldsymbol{\theta\theta}}\mathbf{H}^*_{\boldsymbol{\theta}\alpha} - \widehat{\mathbf{w}}^T \mathbf{H}^*_{\boldsymbol{\theta}\alpha}|}_{E_{21}} + \underbrace{|\widehat{\mathbf{w}}^T \mathbf{H}^*_{\boldsymbol{\theta}\alpha} - \widehat{\mathbf{w}}^T \nabla^2_{\boldsymbol{\theta}\alpha}\mathcal{L}(\widehat{\alpha},\widehat{\boldsymbol{\theta}})|}_{E_{22}}. \end{aligned}$$

For the term $E_{21}$, we have, by Hölder's inequality,
$$E_{21} \leq \|\mathbf{H}^*_{\alpha\boldsymbol{\theta}}\mathbf{H}^{*-1}_{\boldsymbol{\theta}\boldsymbol{\theta}} - \widehat{\mathbf{w}}^T\|_1 \|\mathbf{H}^*_{\boldsymbol{\theta}\alpha}\|_\infty = \mathcal{O}_\mathbb{P}(s'\lambda'), \tag{A.13}$$
where the last inequality holds by the fact that $\|\mathbf{H}^*_{\alpha\boldsymbol{\theta}}\mathbf{H}^{*-1}_{\boldsymbol{\theta}\boldsymbol{\theta}} - \widehat{\mathbf{w}}^T\|_1 = \mathcal{O}_\mathbb{P}(s'\lambda')$, and $\|\mathbf{H}^*_{\boldsymbol{\theta}\alpha}\|_\infty = \mathcal{O}(1)$ by Assumption 4.3.

For the second term $E_{22}$, we have, by Hölder's inequality,
$$E_{22} \leq \|\widehat{\mathbf{w}}\|_1 \|\mathbf{H}^*_{\boldsymbol{\theta}\alpha} - \nabla^2_{\boldsymbol{\theta}\alpha}\mathcal{L}(\widehat{\alpha},\widehat{\boldsymbol{\theta}})\|_\infty = \mathcal{O}_\mathbb{P}(s's\lambda), \tag{A.14}$$
where the last equality holds by the assumption that $\|\mathbf{w}^*\|_1 = \mathcal{O}(s')$, the result $\|\widehat{\mathbf{w}} - \mathbf{w}^*\| = \mathcal{O}_\mathbb{P}(s'\lambda')$ by (A.1) and by Lemma C.5 that $\|\mathbf{H}^*_{\boldsymbol{\theta}\alpha} - \nabla^2_{\boldsymbol{\theta}\alpha}\mathcal{L}(\widehat{\alpha},\widehat{\boldsymbol{\theta}})\|_\infty = \mathcal{O}_\mathbb{P}(s\lambda)$.

Combining (A.13) and (A.14), we have, $E_2 \leq E_{21} + E_{22} = \mathcal{O}_\mathbb{P}(s'\lambda')$. Together with the result that $E_1 = \mathcal{O}_\mathbb{P}(s^2\lambda)$, the claim holds as desired. $\square$

*Proof of Theorem 4.7.* Based on our construction of $\widetilde{\alpha}$ in (3.7), we have
$$\widetilde{\alpha} = \widehat{\alpha} - \left\{\frac{\partial \widehat{U}(\widehat{\alpha},\widehat{\boldsymbol{\theta}})}{\partial \alpha}\right\}^{-1} \widehat{U}(\widehat{\alpha},\widehat{\boldsymbol{\theta}}) = \widehat{\alpha} - H^{-1}_{\alpha|\boldsymbol{\theta}}\widehat{U}(\widehat{\alpha},\widehat{\boldsymbol{\theta}}) + \underbrace{\widehat{U}(\widehat{\alpha},\widehat{\boldsymbol{\theta}})\left[H^{-1}_{\alpha|\boldsymbol{\theta}} - \left\{\frac{\partial \widehat{U}(\widehat{\alpha},\widehat{\boldsymbol{\theta}})}{\partial \alpha}\right\}^{-1}\right]}_{R_1}$$
$$= \widehat{\alpha} - H^{-1}_{\alpha|\boldsymbol{\theta}}\left\{\widehat{U}(0,\widehat{\boldsymbol{\theta}}) + \frac{(\widehat{\alpha} - 0)\partial \widehat{U}(\bar{\alpha},\widehat{\boldsymbol{\theta}})}{\partial \alpha}\right\} + R_1 \tag{A.15}$$
$$= \widehat{\alpha} - H^{-1}_{\alpha|\boldsymbol{\theta}}\widehat{U}(0,\widehat{\boldsymbol{\theta}}) - \widehat{\alpha}H^{-1}_{\alpha|\boldsymbol{\theta}}H_{\alpha|\boldsymbol{\theta}} + \underbrace{\widehat{\alpha}H^{-1}_{\alpha|\boldsymbol{\theta}}\left\{H_{\alpha|\boldsymbol{\theta}} - \frac{\partial \widehat{U}(\bar{\alpha},\widehat{\boldsymbol{\theta}})}{\partial \alpha}\right\}}_{R_2} + R_1 = -H^{-1}_{\alpha|\boldsymbol{\theta}}\widehat{U}(0,\widehat{\boldsymbol{\theta}}) + R_1 + R_2,$$

where (A.15) holds by the mean value theorem for some $\bar{\alpha} = u\widehat{\alpha}$ and $u \in [0,1]$. For the term $R_1$, note that
$$|\widehat{U}(\widehat{\alpha},\widehat{\boldsymbol{\theta}}) - \widehat{U}(0,\widehat{\boldsymbol{\theta}})| = |\widehat{\alpha}| \cdot \left|\frac{\partial \widehat{U}(\bar{\alpha}',\widehat{\boldsymbol{\theta}})}{\partial \alpha}\right|$$
where the equality holds by mean-value theorem with $\bar{\alpha}' = u\widehat{\alpha}$ for some $u \in [0,1]$. Under the null hypothesis $\alpha^* = 0$, by Theorem 3.2 of Huang et al. (2013), $|\widehat{\alpha} - \alpha^*| \leq \|\widehat{\boldsymbol{\beta}} - \boldsymbol{\beta}^*\|_1 = \mathcal{O}_\mathbb{P}(s\lambda)$. By regularity condition $H_{\alpha|\boldsymbol{\theta}} = \mathcal{O}(1)$ and Lemma 4.5, it also holds that $|\partial\widehat{U}(\bar{\alpha}',\widehat{\boldsymbol{\theta}})/\partial\alpha| = \mathcal{O}_\mathbb{P}(1)$. Thus, we have
$$|\widehat{U}(\widehat{\alpha},\widehat{\boldsymbol{\theta}}) - \widehat{U}(0,\widehat{\boldsymbol{\theta}})| = \mathcal{O}_\mathbb{P}(s\lambda), \text{ and } |\widehat{U}(0,\widehat{\boldsymbol{\theta}})| = \mathcal{O}_\mathbb{P}(n^{-1/2}), \tag{A.16}$$
where the second equality holds by Theorem 4.4. Thus, by triangle inequality, we have

$$|R_1| \leq |\widehat{U}(\widehat{\alpha},\widehat{\boldsymbol{\theta}}) - \widehat{U}(0,\widehat{\boldsymbol{\theta}})| \cdot \left|H^{-1}_{\alpha|\boldsymbol{\theta}} - \left\{\frac{\partial \widehat{U}(\widehat{\alpha},\widehat{\boldsymbol{\theta}})}{\partial \alpha}\right\}^{-1}\right| + |\widehat{U}(0,\widehat{\boldsymbol{\theta}})| \cdot \left|H^{-1}_{\alpha|\boldsymbol{\theta}} - \left\{\frac{\partial \widehat{U}(\widehat{\alpha},\widehat{\boldsymbol{\theta}})}{\partial \alpha}\right\}^{-1}\right| = \mathcal{O}_\mathbb{P}\left(s^3\frac{\log d}{n}\right),$$

where the last equality holds by (A.16) and Lemma 4.5.

For the term $R_2$, we have,
$$|R_2| \leq |\widehat{\alpha}H^{-1}_{\alpha|\boldsymbol{\theta}}| \cdot \left|H_{\alpha|\boldsymbol{\theta}} - \frac{\partial \widehat{U}(\bar{\alpha},\widehat{\boldsymbol{\theta}})}{\partial \alpha}\right| = \mathcal{O}_\mathbb{P}\left(s^3\frac{\log d}{n}\right),$$

where the last inequality holds by the fact that $|\widehat{\alpha}| = \mathcal{O}_\mathbb{P}(s\lambda)$, $|H_{\alpha|\boldsymbol{\theta}}| = \mathcal{O}(1)$ and Lemma 4.5.

Consequently, it holds that,

$$\sqrt{n}\widetilde{\alpha} \xrightarrow{d} Z, \text{ where } Z \sim N(0, H_{\alpha|\boldsymbol{\theta}}^{-1}),$$

and the last equality follows by Theorem 4.4 and our the assumption that $n^{-1/2}s^3 \log d = o(1)$. The claim follows as desired. $\square$

*Proof of Theorem 4.9.* We have

$$\mathcal{L}(\widetilde{\alpha}, \widehat{\boldsymbol{\theta}} - \widetilde{\alpha}\widehat{\mathbf{w}}) - \mathcal{L}(0, \widehat{\boldsymbol{\theta}})$$
$$= \widetilde{\alpha}\nabla_\alpha \mathcal{L}(0, \widehat{\boldsymbol{\theta}}) - \widetilde{\alpha}\widehat{\mathbf{w}}^T \nabla_{\boldsymbol{\theta}}\mathcal{L}(0, \widehat{\boldsymbol{\theta}}) + \frac{\widetilde{\alpha}^2}{2}\nabla_{\alpha\alpha}^2\mathcal{L}(\bar{\alpha}, \widehat{\boldsymbol{\theta}}) + \frac{\widetilde{\alpha}^2}{2}\widehat{\mathbf{w}}^T\nabla_{\boldsymbol{\theta}\boldsymbol{\theta}}^2\mathcal{L}(0, \bar{\boldsymbol{\theta}})\widehat{\mathbf{w}} - \widetilde{\alpha}^2\widehat{\mathbf{w}}^T\nabla_{\boldsymbol{\theta}}\mathcal{L}(\bar{\alpha}', \widehat{\boldsymbol{\theta}})$$
$$= \underbrace{\widetilde{\alpha}\widehat{U}(0, \widehat{\boldsymbol{\theta}})}_{T_1} + \underbrace{\frac{\widetilde{\alpha}^2}{2}\left\{\nabla_{\alpha\alpha}^2\mathcal{L}(\bar{\alpha}, \widehat{\boldsymbol{\theta}}) + \widehat{\mathbf{w}}^T\nabla_{\boldsymbol{\theta}\boldsymbol{\theta}}^2\mathcal{L}(0, \bar{\boldsymbol{\theta}})\widehat{\mathbf{w}} - 2\widetilde{\mathbf{w}}^T\nabla_{\boldsymbol{\theta}\alpha}^2\mathcal{L}(\bar{\alpha}', \bar{\boldsymbol{\theta}}')\right\}}_{T_2}, \quad (A.17)$$

where the first equality follows by the mean-value theorem with $\bar{\alpha} = u_1\widehat{\alpha}$, $\bar{\alpha}' = u_2\widehat{\alpha}$, $\bar{\boldsymbol{\theta}} = \boldsymbol{\theta}^* + u_3(\widehat{\boldsymbol{\theta}} - \boldsymbol{\theta}^*)$, and $\bar{\boldsymbol{\theta}}' = \boldsymbol{\theta}^* + u_4(\widehat{\boldsymbol{\theta}} - \boldsymbol{\theta}^*)$ for some $0 \leq u_1, u_2, u_3, u_4 \leq 1$.

We first look at the term $T_1$. Under the null hypothesis $\alpha^* = 0$, $\sqrt{n}\widehat{U}(0, \widehat{\boldsymbol{\theta}}) \xrightarrow{d} Z + o_\mathbb{P}(1)$ and $\sqrt{n}\widetilde{\alpha} = -H_{\alpha|\boldsymbol{\theta}}^{-1}Z + o_\mathbb{P}(1)$ by Theorems 4.4 and 4.7, respectively, where $Z \sim N(0, H_{\alpha|\boldsymbol{\theta}})$. We have,

$$T_1 = \{n^{-1/2}Z + o_\mathbb{P}(n^{-1/2})\}\{-n^{-1/2}H_{\alpha|\boldsymbol{\theta}}^{-1}Z + o_\mathbb{P}(n^{-1/2})\} = -n^{-1}Z^2 H_{\alpha|\boldsymbol{\theta}}^{-1} + o_\mathbb{P}(n^{-1}). \quad (A.18)$$

Next, we look at the term $T_2$,

$$T_2 = \underbrace{\frac{\widetilde{\alpha}^2}{2}\left(\mathbf{H}_{\alpha\alpha}^* + \mathbf{H}_{\alpha\boldsymbol{\theta}}\mathbf{H}_{\boldsymbol{\theta}\boldsymbol{\theta}}^{*-1}\mathbf{H}_{\boldsymbol{\theta}\alpha}^* - 2\mathbf{H}_{\alpha\boldsymbol{\theta}}^*\mathbf{H}_{\boldsymbol{\theta}\boldsymbol{\theta}}^{*-1}\mathbf{H}_{\boldsymbol{\theta}\alpha}^*\right)}_{T_{21}} \quad (A.19)$$
$$+ \underbrace{\frac{\widetilde{\alpha}^2}{2}\left[\{\nabla_{\alpha\alpha}^2\mathcal{L}(\bar{\alpha}, \widehat{\boldsymbol{\theta}}) - \mathbf{H}_{\alpha\alpha}^*\} + \{\widehat{\mathbf{w}}^T\nabla_{\boldsymbol{\theta}\boldsymbol{\theta}}^2\mathcal{L}(0, \bar{\boldsymbol{\theta}})\widehat{\mathbf{w}} - \mathbf{w}^*\mathbf{H}_{\boldsymbol{\theta}\boldsymbol{\theta}}^*\mathbf{w}^*\} - 2\{\widetilde{\mathbf{w}}^T\nabla_{\boldsymbol{\theta}\alpha}^2\mathcal{L}(\bar{\alpha}', \bar{\boldsymbol{\theta}}') - \mathbf{H}_{\alpha\boldsymbol{\theta}}^*\mathbf{w}^*\}\right]}_{T_{22}}$$

It holds by Theorem 4.7 that $\sqrt{n}\widetilde{\alpha} \xrightarrow{d} H_{\alpha\boldsymbol{\theta}}^{-1}Z$. Together with the regularity condition $H_{\alpha|\boldsymbol{\theta}} = \mathcal{O}(1)$, we have,

$$2nT_{21} = n\widetilde{\alpha}^2 H_{\alpha|\boldsymbol{\theta}} \xrightarrow{d} H_{\alpha|\boldsymbol{\theta}}^{-1}Z^2. \quad (A.20)$$

Considering the term $T_{22}$, we have

$$T_{22} = \frac{\widetilde{\alpha}^2}{2}\Big[\underbrace{\{\nabla_{\alpha\alpha}^2\mathcal{L}(\bar{\alpha}, \widehat{\boldsymbol{\theta}}) - \mathbf{H}_{\alpha\alpha}^*\}}_{R_1} + \underbrace{\{\widehat{\mathbf{w}}^T\nabla_{\boldsymbol{\theta}\boldsymbol{\theta}}^2\mathcal{L}(0, \bar{\boldsymbol{\theta}})\widehat{\mathbf{w}} - \mathbf{w}^*\mathbf{H}_{\boldsymbol{\theta}\boldsymbol{\theta}}^*\mathbf{w}^*\}}_{R_2}$$
$$- 2\underbrace{\{\widetilde{\mathbf{w}}^T\nabla_{\alpha\boldsymbol{\theta}}^2\mathcal{L}(\bar{\alpha}', \bar{\boldsymbol{\theta}}') - \mathbf{w}^{*T}\mathbf{H}_{\alpha\boldsymbol{\theta}}^*\}}_{R_3}\Big]. \quad (A.21)$$

For the first term $|R_1|$, we have, by Lemma C.5, $|R_1| = |\nabla^2_{\alpha\alpha}\mathcal{L}(\bar{\alpha}, \widehat{\boldsymbol{\theta}}) - \mathbf{H}^*_{\alpha\alpha}| = \mathcal{O}_{\mathbb{P}}(s\lambda)$. For the second term,

$$\begin{aligned}|R_2| =& |\widehat{\mathbf{w}}^T \nabla^2_{\boldsymbol{\theta\theta}}\mathcal{L}(0, \bar{\boldsymbol{\theta}})\widehat{\mathbf{w}} - \mathbf{w}^* \mathbf{H}^*_{\boldsymbol{\theta\theta}} \mathbf{w}^*| \\ \leq& |(\widehat{\mathbf{w}} - \mathbf{w}^*)^T \nabla^2_{\boldsymbol{\theta\theta}}\mathcal{L}(0, \bar{\boldsymbol{\theta}})(\widehat{\mathbf{w}} - \mathbf{w}^*)| + 2|\mathbf{w}^* \nabla^2_{\boldsymbol{\theta\theta}}\mathcal{L}(0, \bar{\boldsymbol{\theta}})(\widehat{\mathbf{w}} - \mathbf{w}^*)| \\ & + |\mathbf{w}^{*T}(\nabla^2_{\boldsymbol{\theta\theta}}\mathcal{L}(0, \bar{\boldsymbol{\theta}}) - \mathbf{H}^*_{\boldsymbol{\theta\theta}})\mathbf{w}^*| \\ \leq& \|\nabla^2_{\boldsymbol{\theta\theta}}\mathcal{L}(0, \bar{\boldsymbol{\theta}})\|_\infty \|\widehat{\mathbf{w}} - \mathbf{w}^*\|_1^2 + 2\|\mathbf{w}^*\|_1 \|\nabla^2_{\boldsymbol{\theta\theta}}\mathcal{L}(0, \bar{\boldsymbol{\theta}})\|_\infty \|\widehat{\mathbf{w}} - \mathbf{w}^*\|_1 \\ & + \|\mathbf{w}^*\|_1^2 \|\nabla^2_{\boldsymbol{\theta\theta}}\mathcal{L}(0, \bar{\boldsymbol{\theta}}) - \mathbf{H}^*_{\boldsymbol{\theta\theta}}\|_\infty \\ =& \mathcal{O}_{\mathbb{P}}(s'^2\lambda'^2) + \mathcal{O}_{\mathbb{P}}(s'^2\lambda') + \mathcal{O}_{\mathbb{P}}(s'^2 s\lambda),\end{aligned} \quad (A.22)$$

where the last equality follows by (2.2), Lemma C.4, Lemma C.8 and the sparsity Assumption 4.1 of $\mathbf{w}^*$.

For the third term $|R_3|$, we have

$$\begin{aligned}|R_3| \leq & 2\Big[|\{\nabla^2_{\alpha\boldsymbol{\theta}}\mathcal{L}(\bar{\alpha}', \bar{\boldsymbol{\theta}}') - \mathbf{H}^*_{\alpha\boldsymbol{\theta}}\}\widehat{\mathbf{w}}| + |\mathbf{H}^*_{\alpha\boldsymbol{\theta}}(\widehat{\mathbf{w}} - \mathbf{w}^*)|\Big] \\ \leq & 2\Big[|\{\nabla^2_{\alpha\boldsymbol{\theta}}\mathcal{L}(\bar{\alpha}', \bar{\boldsymbol{\theta}}') - \mathbf{H}^*_{\alpha\boldsymbol{\theta}}\}(\widehat{\mathbf{w}} - \mathbf{w}^*)| + |\{\nabla^2_{\alpha\boldsymbol{\theta}}\mathcal{L}(\bar{\alpha}', \bar{\boldsymbol{\theta}}') - \mathbf{H}^*_{\alpha\boldsymbol{\theta}}\}\mathbf{w}^*| + |\mathbf{H}^*_{\alpha\boldsymbol{\theta}}(\widehat{\mathbf{w}} - \mathbf{w}^*)|\Big] \\ \leq & 2\|\nabla^2_{\alpha\boldsymbol{\theta}}\mathcal{L}(\bar{\alpha}', \bar{\boldsymbol{\theta}}') - \mathbf{H}^*_{\alpha\boldsymbol{\theta}}\|_\infty \|\widehat{\mathbf{w}} - \mathbf{w}^*\|_1 + 2\|\nabla^2_{\alpha\boldsymbol{\theta}}\mathcal{L}(\bar{\alpha}', \bar{\boldsymbol{\theta}}') - \mathbf{H}^*_{\alpha\boldsymbol{\theta}}\|_\infty \|\mathbf{w}^*\|_1 \\ & + 2\|\mathbf{H}^*_{\alpha\boldsymbol{\theta}}\|_\infty \|\widehat{\mathbf{w}} - \mathbf{w}^*\|_1.\end{aligned}$$

Note that $\|\nabla^2_{\alpha\boldsymbol{\theta}}\mathcal{L}(\bar{\alpha}', \bar{\boldsymbol{\theta}}') - \mathbf{H}^*_{\alpha\boldsymbol{\theta}}\|_\infty \|\widehat{\mathbf{w}} - \mathbf{w}^*\|_1 = \mathcal{O}_{\mathbb{P}}(s's\lambda'\lambda)$ by Lemma C.8 and Lemma C.4, $\|\nabla_{\alpha\boldsymbol{\theta}}\mathcal{L}(\bar{\alpha}', \bar{\boldsymbol{\theta}}') - \mathbf{H}^*_{\alpha\boldsymbol{\theta}}\|_\infty \|\mathbf{w}^*\|_1 = \mathcal{O}_{\mathbb{P}}(s's\lambda)$ by Lemma C.4 and Assumption 4.2, and $\|\mathbf{H}^*_{\alpha\boldsymbol{\theta}}\|_\infty \|\widehat{\mathbf{w}} - \mathbf{w}^*\|_1 = \mathcal{O}_{\mathbb{P}}(s'\lambda')$ by Assumption 4.3 and Lemma C.8. We have $|R_3| = \mathcal{O}_{\mathbb{P}}(s's\lambda)$.

Combining the results above, we have,

$$T_{22} = \frac{\widetilde{\alpha}^2}{2} \cdot \mathcal{O}_{\mathbb{P}}(s'^2 s\lambda) = \mathcal{O}_{\mathbb{P}}\Big(\frac{s'^2 s\sqrt{\log d}}{n^{3/2}}\Big) = o_{\mathbb{P}}(n^{-1}), \quad (A.23)$$

where the second equality follows by Theorem 4.7 that $\widetilde{\alpha} = \mathcal{O}_{\mathbb{P}}(n^{-1/2})$ under the null hypothesis, and the last equality follows by the assumption that $n^{-1/2} s' s^2 \log d = o(1)$.

Combining (A.20) and (A.23) with (A.19), we have

$$2nT_2 \xrightarrow{d} H^{-1}_{\alpha|\boldsymbol{\theta}} Z^2, \text{ where } Z \sim N(0, H_{\alpha|\boldsymbol{\theta}}). \quad (A.24)$$

Plugging (A.18) and (A.24) into (A.17), by Theorem 4.4,

$$-2n\{\mathcal{L}(\widetilde{\alpha}, \widehat{\boldsymbol{\theta}} - \widetilde{\alpha}\widehat{\mathbf{w}}) - \mathcal{L}(0, \widehat{\boldsymbol{\theta}})\} \xrightarrow{d} Z_\chi^2, \text{where } Z_\chi \sim \chi_1^2,$$

which concludes the proof. □

## B Proofs in Section 5

In this section, we provide detailed proofs in Section 5.

**Lemma B.1.** Under Assumptions 2.1, 2.2, 4.2, 4.3 and 5.1, $\|\nabla\widehat{\Lambda}_0(t, \widehat{\boldsymbol{\beta}}) - \nabla\Lambda_0(t, \boldsymbol{\beta}^*)\|_\infty = \mathcal{O}_{\mathbb{P}}(s\sqrt{n^{-1}\log d})$.

*Proof.* By the definition of $\widehat{\Lambda}_0(t, \widehat{\boldsymbol{\beta}})$ in (5.1), we have,

$$\|\nabla\widehat{\Lambda}_0(t, \widehat{\boldsymbol{\beta}}) - \nabla\Lambda_0(t, \boldsymbol{\beta}^*)\|_\infty = \Big\|\frac{1}{n}\int_0^t \frac{S^{(1)}(u, \widehat{\boldsymbol{\beta}})d\overline{N}(u)}{\{S^{(0)}(u, \widehat{\boldsymbol{\beta}})\}^2} + \mathbb{E}\int_0^t \frac{\mathbf{s}^{(1)}(u, \boldsymbol{\beta}^*)dN(u)}{\{\mathbf{s}^{(0)}(u, \boldsymbol{\beta}^*)\}^2}\Big\|_\infty = \mathcal{O}_\mathbb{P}\Big(s\sqrt{\frac{\log d}{n}}\Big),$$

where the last inequality follows by the same argument in Lemma C.5. $\square$

A corollary of Lemma B.1 and Lemma C.8 follows immediately which characterizes the rate of convergence of $\widehat{\mathbf{u}}(t)$.

**Corollary B.2.** *Under Assumptions 2.1, 2.2, 4.2, 4.3 and 5.1, if $\delta \asymp s'\sqrt{n^{-1}\log d}$ we have,*

$$\|\widehat{\mathbf{u}}(t) - \mathbf{u}^*(t)\|_1 = \mathcal{O}_\mathbb{P}\Big(ss'\sqrt{\frac{\log d}{n}}\Big).$$

*Proof of Theorem 5.2.* We first decompose $\sqrt{n}\{\Lambda_0(t) - \widetilde{\Lambda}_0(t, \widehat{\boldsymbol{\beta}})\}$ into two terms that

$$\sqrt{n}\{\Lambda_0(t) - \widetilde{\Lambda}_0(t, \widehat{\boldsymbol{\beta}})\} = \sqrt{n}\underbrace{\{\Lambda_0(t) - \widehat{\Lambda}_0(t, \boldsymbol{\beta}^*)\}}_{I_1(t)} + \sqrt{n}\underbrace{\{\widehat{\Lambda}_0(t, \boldsymbol{\beta}^*) - \widetilde{\Lambda}_0(t, \widehat{\boldsymbol{\beta}})\}}_{I_2(t)}.$$

Let $M_i(t) = N_i(t) - \int_0^t Y_i(u)\lambda_0(u)du$. For the first term $\sqrt{n}I_1(t)$, we have

$$\sqrt{n}I_1(t) = \int_0^t \frac{\sqrt{n}\sum_{i=1}^n dM_i(u)}{\sum_{i=1}^n Y_i(u)\exp\{\mathbf{X}_i^T(u)\boldsymbol{\beta}^*\}}.$$

Since $M_i(t)$ is a martingale, $\sqrt{n}I_1(t)$ becomes a sum of martingale residuals. By Andersen and Gill (1982), we have, as $n \to \infty$, $\sqrt{n}I_1(t) \xrightarrow{d} N(0, \sigma_1^2(t))$, where

$$\sigma_1^2(t) = \int_0^t \frac{\lambda_0(u)du}{\mathbb{E}\big[\exp\{\mathbf{X}^T(u)\boldsymbol{\beta}^*\}Y(u)\big]}.$$

For the second term $I_2(t)$, we have, by mean value theorem, for some $\widetilde{\boldsymbol{\beta}} = \boldsymbol{\beta}^* + t(\widehat{\boldsymbol{\beta}} - \boldsymbol{\beta}^*)$, $\widetilde{\boldsymbol{\beta}}' = \boldsymbol{\beta}^* + t'(\widehat{\boldsymbol{\beta}} - \boldsymbol{\beta}^*)$ and $0 \leq t, t' \leq 1$,

$$I_2(t) = \widehat{\Lambda}_0(t, \boldsymbol{\beta}^*) - \widehat{\Lambda}_0(t, \widehat{\boldsymbol{\beta}}) + \{\widehat{\mathbf{u}}(t)\}^T\nabla\mathcal{L}(\widehat{\boldsymbol{\beta}})$$

$$= (\boldsymbol{\beta}^* - \widehat{\boldsymbol{\beta}})^T\nabla\widehat{\Lambda}_0(t, \widetilde{\boldsymbol{\beta}}) + \{\widehat{\mathbf{u}}(t)\}^T\{\nabla\mathcal{L}(\boldsymbol{\beta}^*) + \nabla^2\mathcal{L}(\widetilde{\boldsymbol{\beta}}')(\widehat{\boldsymbol{\beta}} - \boldsymbol{\beta}^*)\}$$

$$= \{\mathbf{u}^*(t)\}^T\nabla\mathcal{L}(\boldsymbol{\beta}^*) + \underbrace{(\boldsymbol{\beta}^* - \widehat{\boldsymbol{\beta}})^T\nabla\widehat{\Lambda}_0(t, \widetilde{\boldsymbol{\beta}}) + \{\mathbf{u}^*(t)\}^T\nabla^2\mathcal{L}(\widetilde{\boldsymbol{\beta}}')(\widehat{\boldsymbol{\beta}} - \boldsymbol{\beta}^*)}_{R_1}$$

$$+ \underbrace{\{\widehat{\mathbf{u}}(t) - \mathbf{u}^*(t)\}^T\{\nabla\mathcal{L}(\boldsymbol{\beta}^*) + \nabla^2\mathcal{L}(\widetilde{\boldsymbol{\beta}}')(\widehat{\boldsymbol{\beta}} - \boldsymbol{\beta}^*)\}}_{R_2}.$$

Next, we consider the two terms $R_1$ and $R_2$. For the term $R_1$, we have

$$R_1 = (\boldsymbol{\beta}^* - \widehat{\boldsymbol{\beta}})^T\nabla\widehat{\Lambda}_0(t, \widetilde{\boldsymbol{\beta}}) + \{\mathbf{u}^*(t)\}^T\nabla^2\mathcal{L}(\widetilde{\boldsymbol{\beta}}')(\widehat{\boldsymbol{\beta}} - \boldsymbol{\beta}^*)$$

$$= (\boldsymbol{\beta}^* - \widehat{\boldsymbol{\beta}})^T\big[\mathbf{H}^*\mathbf{H}^{*-1}\nabla\widehat{\Lambda}_0(t, \widetilde{\boldsymbol{\beta}}) - \nabla^2\mathcal{L}(\widetilde{\boldsymbol{\beta}}')\mathbf{H}^{*-1}\nabla\widehat{\Lambda}_0(t, \boldsymbol{\beta}^*)\big]$$

$$= \underbrace{(\boldsymbol{\beta}^* - \widehat{\boldsymbol{\beta}})^T\{\nabla\widehat{\Lambda}_0(t, \widetilde{\boldsymbol{\beta}}) - \nabla\Lambda_0(t, \boldsymbol{\beta}^*)\}}_{R_{11}} + \underbrace{(\boldsymbol{\beta}^* - \widehat{\boldsymbol{\beta}})^T\big[\mathbf{H}^* - \nabla^2\mathcal{L}(\widetilde{\boldsymbol{\beta}}')\big]\mathbf{H}^{*-1}\nabla\Lambda_0(t, \boldsymbol{\beta}^*)}_{R_{12}}.$$

It holds that $|R_{11}| \leq \|\boldsymbol{\beta}^* - \widehat{\boldsymbol{\beta}}\|_1 \|\nabla \Lambda_0(t, \widetilde{\boldsymbol{\beta}}) - \nabla \widehat{\Lambda}_0(t, \boldsymbol{\beta}^*)\|_\infty = \mathcal{O}_{\mathbb{P}}(s^2 n^{-1} \log d)$ by (2.2) and Lemma B.1, and $|R_{12}| \leq \|\boldsymbol{\beta}^* - \widehat{\boldsymbol{\beta}}\|_1 \|\mathbf{H}^* - \nabla^2 \mathcal{L}(\widetilde{\boldsymbol{\beta}}')\|_\infty \|\mathbf{u}^*(t)\|_1 = \mathcal{O}_{\mathbb{P}}(s's^2 n^{-1} \log d)$. Summing them up, by triangle inequality, we have $|R_1| = \mathcal{O}_{\mathbb{P}}(s's^2 n^{-1} \log d)$.

For the term $R_2$, we have

$$|R_2| \leq \|\widehat{\mathbf{u}}(t) - \mathbf{u}^*(t)\|_1 \|\nabla \mathcal{L}(\boldsymbol{\beta}^*)\|_\infty + \|\widehat{\mathbf{u}}(t) - \mathbf{u}^*(t)\|_1 \|\nabla^2 \mathcal{L}(\widetilde{\boldsymbol{\beta}}')\|_\infty \|\widehat{\boldsymbol{\beta}} - \boldsymbol{\beta}^*\|_1$$
$$= \mathcal{O}_{\mathbb{P}}(s' s n^{-1} \log d) + \mathcal{O}_{\mathbb{P}}(s' s^2 n^{-1} \log d),$$

where the last inequality holds by Lemma C.3 and C.5.

Meanwhile, by Lemma A.1, taking $\mathbf{v} = \mathbf{u}^*(t)$, we have the term $\sqrt{n} \mathbf{u}^{*T}(t) \nabla \mathcal{L}(\boldsymbol{\beta}^*) \xrightarrow{d} N(0, \sigma_2^2(t))$, where $\sigma_2^2(t) = \nabla \Lambda_0(t, \boldsymbol{\beta}^*)^T \mathbf{H}^{*-1} \nabla \Lambda_0(t, \boldsymbol{\beta}^*)$. Thus, we have,

$$\sqrt{n} I_2(t) \xrightarrow{d} Z, \text{ where } Z \sim N(0, \sigma_2^2(t)),$$

and $\sigma_2^2(t) = \nabla \Lambda_0(t, \boldsymbol{\beta}^*)^T \mathbf{H}^{*-1} \nabla \Lambda_0(t, \boldsymbol{\beta}^*)$.

Following the standard martingale theory, the covariance between $I_1(t)$ and $I_2(t)$ is 0. Our claim holds as desired. □

## C  Technical Lemmas

In this section, we prove some concentration results of the sample gradient $\nabla \mathcal{L}(\boldsymbol{\beta}^*)$ and sample Hessian matrix $\nabla^2 \mathcal{L}(\boldsymbol{\beta}^*)$. The mathematical tools we use are mainly from empirical process theory.

We start from introducing the following notations. Let $\|\cdot\|_{\mathbb{P},r}$ denote the $L_r(\mathbb{P})$-norm. For any given $\epsilon > 0$ and the function class $\mathcal{F}$, let $N_{[\,]}(\epsilon, \mathcal{F}, L_r(\mathbb{P}))$ and $N(\epsilon, \mathcal{F}, L_2(\mathbb{Q}))$ denote the bracketing number and the covering number, respectively. The quantifies $\log N_{[\,]}(\epsilon, \mathcal{F}, L_r(\mathbb{P}))$ and $\log N(\epsilon, \mathcal{F}, L_2(\mathbb{Q}))$ are called entropy with bracketing and entropy, respectively. In addition, let $F$ be an envelope of $\mathcal{F}$ where $|f| \leq F$ for all $f \in \mathcal{F}$. The bracketing integral and uniform entropy integral are defined as

$$J_{[\,]}(\delta, \mathcal{F}, L_r(\mathbb{P})) = \int_0^\delta \sqrt{\log N_{[\,]}(\epsilon, \mathcal{F}, L_r(\mathbb{P}))} d\epsilon,$$

and

$$J(\delta, \mathcal{F}, L_2) = \int_0^\delta \sqrt{\log \sup_{\mathbb{Q}} N(\epsilon \|F\|_{\mathbb{Q},2}, \mathcal{F}, L_2(\mathbb{Q}))} d\epsilon,$$

respectively, where the supremum is taken over all probability measures $\mathbb{Q}$ with $\|F\|_{\mathbb{Q},2} > 0$. Denote the empirical process by $\mathbb{G}_n(f) = n^{1/2}(\mathbb{P}_n - \mathbb{P})(f)$, where $\mathbb{P}_n(f) = n^{-1} \sum_{i=1}^n f(X_i)$ and $\mathbb{P}(f) = \mathbb{E}(f(X_i))$. The following three Lemmas characterize the bounds for the expected maximal empirical processes and the concentration of the maximal empirical processes.

**Lemma C.1.** Under Assumptions 2.1, 2.2, 4.1, 4.2 and 4.3, there exist some constant $C > 0$, such that, for $r = 0, 1, 2$, with probability at least $1 - \mathcal{O}(d^{-3})$,

$$\sup_{t \in [0,\tau]} \|\mathbf{s}^{(r)}(t, \boldsymbol{\beta}^*) - S^{(r)}(t, \boldsymbol{\beta}^*)\|_\infty \leq C \sqrt{\frac{\log d}{n}},$$

where $\mathbf{s}^{(r)}(t, \boldsymbol{\beta}^*)$ and $S^{(r)}(t, \boldsymbol{\beta}^*)$ are defined in (2.6) and (2.3).

*Proof.* We will only prove the case for $r = 1$, and the cases for $r = 0$ and $2$ follow by the similar argument. For $j = 1, ..., d$, let

$$E_j = \sup_{t \in [0,\tau]} |S_j^{(1)}(t, \boldsymbol{\beta}^*) - s_j^{(1)}(t, \boldsymbol{\beta}^*)|,$$

where $S_j^{(1)}(t, \boldsymbol{\beta}^*)$ and $s_j^{(1)}(t, \boldsymbol{\beta}^*)$ denote the $j$-th component of $S^{(1)}(t, \boldsymbol{\beta}^*)$ and $s^{(1)}(t, \boldsymbol{\beta}^*)$, respectively. We will prove a concentration result of $E_j$.

First, we show the class of functions $\{X_j(t)Y(t)\exp(\boldsymbol{X}^T(t)\boldsymbol{\beta}^*) : t \in [0,\tau]\}$ has bounded uniform entropy integral. By Lemma 9.10 of Kosorok (2007), the class $\mathcal{F} = \{X_j(t) : t \in [0,\tau]\}$ is a VC-hull class associated with a VC class of index 2. By Corollary 2.6.12 of van der Vaart and Wellner (1996), the entropy of the class $\mathcal{F}$ satisfies $\log N(\epsilon \|F\|_{Q,2}, \mathcal{F}, L_2(\mathbb{Q})) \leq C'(1/\epsilon)$ for some constant $C' > 0$, and hence $\mathcal{F}$ has the uniform entropy integral $J(1, \mathcal{F}, L_2) \leq \int_0^1 \sqrt{K(1/\epsilon)} d\epsilon < \infty$. By the same argument, we have that $\{\exp\{\boldsymbol{X}(t)^T \boldsymbol{\beta}^*\} : t \in [0,\tau]\}$ also has a uniform entropy integral. Meanwhile, by example 19.16 of van der Vaart and Wellner (1996), $\{Y(t) : t \in [0,\tau]\}$ is a VC class and hence has bounded uniform entropy integral. Thus, by Theorem 9.15 of Kosorok (2007), we have $\{X_j(t)Y(t)\exp\{\boldsymbol{X}(t)^T \boldsymbol{\beta}^*\} : t \in [0,\tau]\}$ has bounded uniform entropy integral.

Next, taking the envelop $F$ as $\sup_{t \in [0,\tau]} |X_j(t)Y(t)\exp\{\boldsymbol{X}^T(t)\boldsymbol{\beta}^*\}|$, by Lemma 19.38 of van der Vaart (2000),

$$\mathbb{E}(E_j) \leq C_1 n^{-1/2} J(1, \mathcal{F}, L_2) \|F\|_{\mathbb{P},2} \leq C n^{-1/2},$$

for some positive constants $C_1$ and $C$. By McDiarmid's inequality, we have, for any $\Delta > 0$,

$$\mathbb{P}(E_j \geq C n^{-1/2}(1+\Delta)) \leq \mathbb{P}(E_j \geq \mathbb{E}(E_j) + n^{-1/2} C \Delta) \leq \exp(-C_2 \Delta^2 L^{-2}),$$

for some positive constant $C_2$ and $L$, and the desired result follows by taking $\Delta = \sqrt{n^{-1} \log d}$ a union bound over $j = 1, ..., d$. $\square$

**Lemma C.2.** *Suppose the Assumptions 2.1, 2.2, 4.1, 4.2 and 4.3 hold, and $\lambda \asymp \sqrt{n^{-1} \log d}$. We have, for $r = 0, 1, 2$ and $t \in [0, \tau]$,*

$$\|S^{(r)}(t, \widehat{\boldsymbol{\beta}}) - S^{(r)}(t, \boldsymbol{\beta}^*)\|_\infty = \mathcal{O}_\mathbb{P}\left(s \sqrt{\frac{\log d}{n}}\right).$$

*Proof.* Similar to the previous Lemma, we only prove the case for $r = 1$, and the other two cases follow by the similar argument. For the case $r = 1$, we have

$$\|S^{(1)}(t, \widehat{\boldsymbol{\beta}}) - S^{(1)}(t, \boldsymbol{\beta}^*)\|_\infty = \left\| \frac{1}{n} \sum_{i=1}^n Y_i(t) \left[ \exp\{\boldsymbol{X}_i^T(t)\widehat{\boldsymbol{\beta}}\} - \exp\{\boldsymbol{X}_i^T(t)\boldsymbol{\beta}^*\} \right] \boldsymbol{X}_i(t) \right\|_\infty$$

$$\leq \max_i \left\{ Y_i(t) \|\boldsymbol{X}_i(t)\|_\infty \big| \exp\{\boldsymbol{X}_i^T(t)\widehat{\boldsymbol{\beta}}\} - \exp\{\boldsymbol{X}_i^T(t)\boldsymbol{\beta}^*\} \big| \right\}$$

$$\leq C_X \cdot \max_i \big| \exp\{\boldsymbol{X}_i^T(t)\boldsymbol{\beta}^*\} \left[ \exp\{\boldsymbol{X}_i^T(t)(\widehat{\boldsymbol{\beta}} - \boldsymbol{\beta}^*)\} - 1 \right] \big| \quad \text{(C.1)}$$

$$\leq C_X \cdot C_1 \cdot \max_i \|\boldsymbol{X}_i(t)\|_\infty \|\widehat{\boldsymbol{\beta}} - \boldsymbol{\beta}^*\|_1 \quad \text{(C.2)}$$

$$= \mathcal{O}_\mathbb{P}\left(s \sqrt{\frac{\log d}{n}}\right),$$

where (C.1) holds by the Assumption 2.1 for some constant $C_X > 0$; (C.2) holds by Assumption 4.1 that $\boldsymbol{X}_i^T(t)\boldsymbol{\beta}^* = \mathcal{O}(1)$ and $\exp(|x|) \leq 1 + 2|x|$ for any $|x|$ sufficiently small, and the last equality holds by (2.2). Our claim holds as desired. $\square$

**Lemma C.3.** Under Assumptions 2.1, 2.2, 4.1, 4.2 and 4.3, there exists a positive constant $C$, such that with probability at least $1 - \mathcal{O}(d^{-3})$,

$$\|\nabla \mathcal{L}(\boldsymbol{\beta}^*)\|_\infty \leq C\sqrt{\frac{\log d}{n}}.$$

*Proof.* By definition, we have, for all $j = 1, ..., d$,

$$\nabla_j \mathcal{L}(\boldsymbol{\beta}^*) = -\frac{1}{n}\sum_{i=1}^n \int_0^\tau \{X_{ij}(u, \boldsymbol{\beta}^*) - \overline{X}_j(u, \boldsymbol{\beta}^*)\} dM_i(u)$$

$$= \frac{1}{n}\sum_{i=1}^n \int_0^\tau \overline{X}_j(u, \boldsymbol{\beta}^*) dM_i(u) - \frac{1}{n}\sum_{i=1}^n \int_0^\tau X_{ij}(u, \boldsymbol{\beta}^*) dM_i(u). \quad (C.3)$$

For the first term, we have for all $t \in [0, \tau]$,

$$\overline{X}_j(t, \boldsymbol{\beta}^*) - e_j(t, \boldsymbol{\beta}^*) = \frac{S_j^{(1)}(t, \boldsymbol{\beta}^*) - s_j^{(1)}(t, \boldsymbol{\beta}^*)}{S^{(0)}(t, \boldsymbol{\beta}^*)} - \frac{s_j^{(1)}(t, \boldsymbol{\beta}^*)\{S^{(0)}(t, \boldsymbol{\beta}^*) - s^{(0)}(t, \boldsymbol{\beta}^*)\}}{S^{(0)}(t, \boldsymbol{\beta}^*) s^{(0)}(t, \boldsymbol{\beta}^*)}. \quad (C.4)$$

By Assumption 2.1 and the fact that $\mathbb{P}(y(\tau) > 0) > 0$, we have that $\sup_{t \in [0,\tau]} |\overline{X}_j(t, \boldsymbol{\beta}^*) - e_j(t)| \leq C_1$ for some constant $C_1 > 0$. In addition,

$$\frac{1}{n}\sum_{i=1}^n \int_0^\tau \overline{X}_j(u, \boldsymbol{\beta}^*) dM_i(u) \leq \sup_{f \in \mathcal{F}_j} \frac{1}{n}\sum_{i=1}^n \int_0^\tau f(u) dM_i(u),$$

where $\mathcal{F}_j$ denotes the class of functions $f : [0, \tau] \to \mathbb{R}$ which have uniformly bounded variation and satisfy $\sup_{t \in [0,\tau]} |f(t) - e_j(t)| \leq \delta_1$ for some $\delta_1$. By constructing $\ell_\infty$ balls centered at piecewise constant functions on a regular grid, one can show that the covering number of the class $\mathcal{F}_j$ satisfies $N(\epsilon, \mathcal{F}_j, \ell_\infty) \leq (C_2 \epsilon^{-1})^{C_3 \epsilon^{-1}}$ for some positive constants $C_2, C_3$. Let $\mathcal{G}_j = \{\int_0^\infty f(t) dM(t) : f \in \mathcal{F}_j\}$. Note that for any two $f_1, f_2 \in \mathcal{F}_j$,

$$\Big|\int_0^\tau f_1(t) - f_2(t) dM(t)\Big| \leq \sup_{u \in [0,\tau]} |f_1(u) - f_2(u)| \int_0^\tau |dM(t)|.$$

By Theorem 2.7.11 of van der Vaart and Wellner (1996), the bracketing number of the class $\mathcal{G}_j$ satisfies $N_{[\,]}(2\epsilon \|F\|_{\mathbb{P},2}, \mathcal{G}_j, \ell_2(\mathbb{P})) \leq N(\epsilon, \mathcal{F}_j, \|\cdot\|_\infty) \leq (C_2 \epsilon^{-1})^{C_3 \epsilon^{-1}}$, where $F = \int_0^\tau |dM(t)|$. Hence, $\mathcal{G}_j$ has bounded bracketing integral. An application of Corollary 19.35 of van der Vaart (2000) yields that

$$\mathbb{E}\Big(\sup_{f \in \mathcal{F}_j} \frac{1}{n}\sum_{i=1}^n \int_0^\tau f(u) dM_i(u)\Big) \leq n^{-1/2} C_4$$

for some constant $C_4 > 0$. Then, by McDiarmid's inequality,

$$\mathbb{P}\Big(\frac{1}{n}\sum_{i=1}^n \int_0^\tau \overline{X}_j(u, \boldsymbol{\beta}^*) dM_i(u) > t\Big) \leq \mathbb{P}\Big(\sup_{f \in \mathcal{F}_j} \frac{1}{n}\sum_{i=1}^n \int_0^\tau f(u) dM_i(u) > t\Big) \leq \exp\Big(-\frac{nt^2}{C_5}\Big),$$

for some constant $C_5$. Following by the union bound, we have with probability at least $1 - \mathcal{O}(d^{-3})$,

$$\left\|\frac{1}{n}\sum_{i=1}^{n}\int_{0}^{\tau}\overline{X}_{j}(u,\boldsymbol{\beta}^{*})dM_{i}(u)\right\|_{\infty} \leq C\sqrt{\frac{\log d}{n}}.$$

Note that the second term of (C.3) is a sum of i.i.d. mean-zero bounded random variables. Following by the Hoeffding inequality and the union bound, we have with probability at least $1 - \mathcal{O}(d^{-3})$,

$$\left\|\frac{1}{n}\sum_{i=1}^{n}\int_{0}^{\infty}X_{ij}(u,\boldsymbol{\beta}^{*})dM_{i}(u)\right\|_{\infty} \leq C\sqrt{\frac{\log d}{n}},$$

for some constant $C$. The claim follows as desired. $\square$

**Lemma C.4.** Under Assumptions 2.1, 2.2, 4.1, 4.2 and 4.3, for any $1 \leq j, k \leq d$, there exists a positive constant $C$, such that with probability at least $1 - \mathcal{O}(d^{-1})$,

$$\max_{j,k=1,\ldots,d}|\nabla_{jk}^{2}\mathcal{L}(\boldsymbol{\beta}^{*}) - \mathbf{H}_{jk}^{*}| \leq C\sqrt{\frac{\log d}{n}}. \tag{C.5}$$

*Proof.* By the definitions of $\nabla^{2}\mathcal{L}(\boldsymbol{\beta}^{*})$ and $\mathbf{H}^{*}$ in (2.5) and (2.7), we have

$$\nabla^{2}\mathcal{L}(\boldsymbol{\beta}^{*}) - \mathbf{H}^{*} = \underbrace{\frac{1}{n}\int_{0}^{\tau}\left\{\frac{S^{(2)}(t,\boldsymbol{\beta}^{*})}{S^{(0)}(t,\boldsymbol{\beta}^{*})} - \frac{\mathbf{s}^{(2)}(t,\boldsymbol{\beta}^{*})}{\mathbf{s}^{(0)}(t,\boldsymbol{\beta}^{*})}\right\}d\overline{N}(t)}_{T_1}$$

$$+ \underbrace{\frac{1}{n}\int_{0}^{\tau}\frac{\mathbf{s}^{(2)}(t,\boldsymbol{\beta}^{*})}{\mathbf{s}^{(0)}(t,\boldsymbol{\beta}^{*})}d\overline{N}(t) - \mathbb{E}\left[\int_{0}^{\tau}\frac{\mathbf{s}^{(2)}(t,\boldsymbol{\beta}^{*})}{\mathbf{s}^{(0)}(t,\boldsymbol{\beta}^{*})}dN(t)\right]}_{T_2}$$

$$+ \underbrace{\frac{1}{n}\int_{0}^{\tau}\left\{\mathbf{e}(t,\boldsymbol{\beta}^{*})^{\otimes 2} - \overline{\mathbf{Z}}(t,\boldsymbol{\beta}^{*})^{\otimes 2}\right\}d\overline{N}(t)}_{T_3}$$

$$+ \underbrace{\mathbb{E}\left[\int_{0}^{\tau}\mathbf{e}(t,\boldsymbol{\beta}^{*})^{\otimes 2}dN(t)\right] - \frac{1}{n}\int_{0}^{\tau}\mathbf{e}(t,\boldsymbol{\beta}^{*})^{\otimes 2}d\overline{N}(t)}_{T_4}.$$

For the term $T_1$, we have, with probability at least $1 - \mathcal{O}(d^{-1})$,

$$\|T_1\|_{\infty} \leq \sup_{t \in [0,\tau]}\left\|\frac{S^{(2)}(t,\boldsymbol{\beta}^{*})}{S^{(0)}(t,\boldsymbol{\beta}^{*})} - \frac{\mathbf{s}^{(2)}(t,\boldsymbol{\beta}^{*})}{\mathbf{s}^{(0)}(t,\boldsymbol{\beta}^{*})}\right\|_{\infty} \cdot \frac{1}{n}\int_{0}^{\tau}d\overline{N}(t) \leq C_1\sqrt{\frac{\log d}{n}},$$

where the last inequality follows by Lemma C.1. Next, by Assumption 2.1, we have

$$\left\|\frac{\mathbf{s}^{(2)}(t,\boldsymbol{\beta}^{*})}{\mathbf{s}^{(0)}(t,\boldsymbol{\beta}^{*})}\right\|_{\infty} < \infty.$$

Consequently, $T_2$ becomes an i.i.d. sum of mean 0 bounded random variables. Hoeffding's inequality gives that with probability at least $1 - \mathcal{O}(d^{-1})$, $\|T_2\|_{\infty} \leq C_2\sqrt{n^{-1}\log d}$. Meanwhile, the terms $T_3$ and $T_4$ can be bounded similarly. Our claim holds as desired. $\square$

**Lemma C.5.** Under Assumptions 2.1, 2.2, 4.1, 4.2 and 4.3, let $\widehat{\boldsymbol{\beta}}$ be the estimator for $\boldsymbol{\beta}^*$ estimated by (2.1) satisfying the result in (2.2) that $\|\widehat{\boldsymbol{\beta}} - \boldsymbol{\beta}^*\|_1 = \mathcal{O}_{\mathbb{P}}(s\lambda)$ with $\lambda \asymp \mathcal{O}(\sqrt{n^{-1}\log d})$. Then, we have, for any $\widetilde{\boldsymbol{\beta}} = \boldsymbol{\beta}^* + u(\widehat{\boldsymbol{\beta}} - \boldsymbol{\beta}^*)$ with $u \in [0,1]$,

$$\|\nabla^2 \mathcal{L}(\widetilde{\boldsymbol{\beta}})\|_\infty = \mathcal{O}_{\mathbb{P}}(1), \text{ and } \|\nabla^2 \mathcal{L}(\widetilde{\boldsymbol{\beta}}) - \mathbf{H}^*\|_\infty = \mathcal{O}_{\mathbb{P}}\Big(s\sqrt{\frac{\log d}{n}}\Big).$$

*Proof.* Let $\xi = \max_{u \geq 0} \max_{i,i'} |\boldsymbol{\Delta}^T \{\boldsymbol{X}_i(u) - \boldsymbol{X}_{i'}(u)\}|$, where $\boldsymbol{\Delta} = \widetilde{\boldsymbol{\beta}} - \boldsymbol{\beta}^*$. By Lemma 3.2 of Huang et al. (2013), it holds that,

$$\exp(-2\xi)\nabla^2\mathcal{L}(\boldsymbol{\beta}^*) \preceq \nabla^2\mathcal{L}(\widetilde{\boldsymbol{\beta}}) \preceq \exp(2\xi)\nabla^2\mathcal{L}(\boldsymbol{\beta}^*), \tag{C.6}$$

where $\mathbf{A} \preceq \mathbf{B}$ means that the matrix $\mathbf{B} - \mathbf{A}$ is a positive semidefinite matrix.

Note that the diagonal elements of a positive semidefinite matrix can only be nonnegative. In addition, for a positive semidefinite matrix $\mathbf{A} \in \mathbb{R}^{d \times d}$, it is easy to see that $\|\mathbf{A}\|_\infty = \max_j\{a_{jj}\}_{j=1}^d$. We have,

$$\exp(-2\xi)\|\nabla^2\mathcal{L}(\boldsymbol{\beta}^*)\|_\infty \leq \|\nabla^2\mathcal{L}(\widetilde{\boldsymbol{\beta}})\|_\infty \leq \exp(2\xi)\|\nabla^2\mathcal{L}(\boldsymbol{\beta}^*)\|_\infty.$$

By (2.2) that $\|\widehat{\boldsymbol{\beta}} - \boldsymbol{\beta}^*\|_1 = \mathcal{O}_{\mathbb{P}}(s\lambda)$, which implies that $\|\widetilde{\boldsymbol{\beta}} - \boldsymbol{\beta}^*\|_1 = \mathcal{O}(s\lambda)$ as $\widetilde{\boldsymbol{\beta}}$ is on the line segment connecting $\boldsymbol{\beta}^*$ and $\widehat{\boldsymbol{\beta}}$. Hence, $\xi = \mathcal{O}_{\mathbb{P}}(s\lambda)$. By triangle inequality,

$$\|\nabla^2\mathcal{L}(\widetilde{\boldsymbol{\beta}}) - \mathbf{H}^*\|_\infty \leq \underbrace{\|\nabla^2\mathcal{L}(\widetilde{\boldsymbol{\beta}}) - \nabla^2\mathcal{L}(\boldsymbol{\beta}^*)\|_\infty}_{E_1} + \underbrace{\|\nabla^2\mathcal{L}(\boldsymbol{\beta}^*) - \mathbf{H}^*\|_\infty}_{E_2}.$$

We consider the two terms separately, for the first term $E_1$, we have, by (C.6) and taking the Taylor's expansion of $\exp(2\xi)$,

$$\|\nabla^2\mathcal{L}(\widetilde{\boldsymbol{\beta}}) - \nabla^2\mathcal{L}(\boldsymbol{\beta}^*)\|_\infty \leq 2\|\xi \nabla^2\mathcal{L}(\boldsymbol{\beta}^*)\|_\infty + o_{\mathbb{P}}(\xi).$$

Since $\xi = \mathcal{O}_{\mathbb{P}}(s\lambda)$, and by Assumption 4.3, we have,

$$\|\nabla^2\mathcal{L}(\widetilde{\boldsymbol{\beta}}) - \nabla^2\mathcal{L}(\boldsymbol{\beta}^*)\|_\infty = \mathcal{O}_{\mathbb{P}}(s\lambda),$$

and $E_1 = \mathcal{O}_{\mathbb{P}}(s\sqrt{n^{-1}\log d})$ as $\lambda \asymp \sqrt{n^{-1}\log d}$. In addition, $E_2 = \mathcal{O}_{\mathbb{P}}(\sqrt{n^{-1}\log d})$ by Lemma C.4. It further implies that $\|\nabla^2\mathcal{L}(\widetilde{\boldsymbol{\beta}})\|_\infty = \mathcal{O}_{\mathbb{P}}(1)$. □

**Lemma C.6.** Under Assumptions 2.1, 2.2 4.1, 4.2 and 4.3, it holds that

$$\|\nabla^2_{\alpha\boldsymbol{\theta}}\mathcal{L}(\widehat{\boldsymbol{\beta}}) - \mathbf{w}^{*T}\nabla^2_{\boldsymbol{\theta}\boldsymbol{\theta}}\mathcal{L}(\widehat{\boldsymbol{\beta}})\|_\infty = \mathcal{O}_{\mathbb{P}}\Big(s\sqrt{\frac{\log d}{n}}\Big).$$

*Proof.* By triangle inequality, we have

$$\|\nabla^2_{\alpha\boldsymbol{\theta}}\mathcal{L}(\widehat{\boldsymbol{\beta}}) - \mathbf{w}^{*T}\nabla^2_{\boldsymbol{\theta}\boldsymbol{\theta}}\mathcal{L}(\widehat{\boldsymbol{\beta}})\|_\infty$$
$$\leq \underbrace{\|\mathbf{H}^*_{\alpha\boldsymbol{\theta}} - \mathbf{w}^{*T}\mathbf{H}^*_{\boldsymbol{\theta}\boldsymbol{\theta}}\|_\infty}_{E_1} + + \underbrace{\|\nabla^2_{\boldsymbol{\theta}\alpha}\mathcal{L}(\widehat{\boldsymbol{\beta}}) - \mathbf{H}^*_{\boldsymbol{\theta}\alpha}\|_\infty}_{E_2} + \underbrace{\|\mathbf{w}^{*T}\{\mathbf{H}^*_{\boldsymbol{\theta}\boldsymbol{\theta}} - \nabla^2_{\boldsymbol{\theta}\boldsymbol{\theta}}\mathcal{L}(\widehat{\boldsymbol{\beta}})\}\|_\infty}_{E_3}.$$

It is seen that $E_1 = 0$ by the definition of $\mathbf{w}^* = \mathbf{H}_{\boldsymbol{\theta\theta}}^{*-1}\mathbf{H}_{\boldsymbol{\theta}\alpha}^*$ in (3.1). In addition, $E_2 = \mathcal{O}_{\mathbb{P}}(s\sqrt{n^{-1}\log d})$ by Lemma C.5. For the term $E_3$, we have

$$E_3 \le \underbrace{\|\mathbf{w}^{*T}\{\nabla^2_{\boldsymbol{\theta\theta}}\mathcal{L}(\widehat{\boldsymbol{\beta}}) - \nabla^2_{\boldsymbol{\theta\theta}}\mathcal{L}(\boldsymbol{\beta}^*)\}\|_\infty}_{E_{31}} + \underbrace{\|\mathbf{w}^{*T}\{\nabla^2_{\boldsymbol{\theta\theta}}\mathcal{L}(\boldsymbol{\beta}^*) - \mathbf{H}_{\boldsymbol{\theta\theta}}^*\}\|_\infty}_{E_{32}}.$$

For the term $E_{31}$, by the definition of $\nabla^2\mathcal{L}(\cdot)$ in (2.5), we have

$$\mathbf{w}^{*T}\{\nabla^2_{\boldsymbol{\theta\theta}}\mathcal{L}(\widehat{\boldsymbol{\beta}}) - \nabla^2_{\boldsymbol{\theta\theta}}\mathcal{L}(\boldsymbol{\beta}^*)) = \mathbf{w}^{*T}\underbrace{\left\{\frac{1}{n}\sum_{i=1}^n \int_0^\tau \frac{S^{(2)}(t,\widehat{\boldsymbol{\beta}})}{S^{(0)}(t,\widehat{\boldsymbol{\beta}})} - \frac{S^{(2)}(t,\boldsymbol{\beta}^*)}{S^{(0)}(t,\boldsymbol{\beta}^*)} dN_i(t)\right\}_{\boldsymbol{\theta\theta}}}_{T_1}$$

$$+ \mathbf{w}^{*T}\underbrace{\left\{\frac{1}{n}\sum_{i=1}^n \int_0^\tau \overline{\mathbf{Z}}(t,\widehat{\boldsymbol{\beta}})^{\otimes 2} - \overline{\mathbf{Z}}(t,\boldsymbol{\beta}^*)^{\otimes 2}\right\}_{\boldsymbol{\theta\theta}}}_{T_2}.$$

For the term $T_1$, we have

$$T_1 = \frac{1}{n}\sum_{i=1}^n \int_0^\tau \frac{S^{(0)}(t,\boldsymbol{\beta}^*)\mathbf{w}^{*T}S^{(2)}_{\boldsymbol{\theta\theta}}(t,\widehat{\boldsymbol{\beta}}) - S^{(0)}(t,\widehat{\boldsymbol{\beta}})\mathbf{w}^{*T}S^{(2)}_{\boldsymbol{\theta\theta}}(t,\boldsymbol{\beta}^*)}{S^{(0)}(t,\widehat{\boldsymbol{\beta}})S^{(0)}(t,\boldsymbol{\beta}^*)}$$

For ease of notation, in the rest of the proof, let $\widehat{S}^{(r)}(t) := S^{(r)}(t,\widehat{\boldsymbol{\beta}})$ and $S^{*(r)}(t) := S^{(r)}(t,\boldsymbol{\beta}^*)$ for $r = 0,1,2$. We have, for the $k$-th component of $T_1$,

$$T_{1,k} = \frac{1}{n}\sum_{i=1}^n \int_0^\tau \frac{S^{*(0)}(t)\frac{1}{n}\sum_{i'=1}^n y_{i'}(t)\exp\{\mathbf{X}_{i'}^T(t)\widehat{\boldsymbol{\beta}}\}\mathbf{w}^{*T}\mathbf{X}_{i',\boldsymbol{\theta}}(t)X_{i',k}(t)}{\widehat{S}^{(0)}(t)S^{*(0)}(t)}dN_i(t)$$

$$- \frac{1}{n}\sum_{i=1}^n \int_0^\tau \frac{\widehat{S}^{(0)}(t)\sum_{i'=1}^n y_{i'}(t)\exp\{\mathbf{X}_{i'}^T(t)\boldsymbol{\beta}^*\}\mathbf{w}^{*T}\mathbf{X}_{i',\boldsymbol{\theta}}(t)X_{i',k}(t)}{\widehat{S}^{(0)}(t)S^{*(0)}(t)}dN_i(t).$$

Consequently, it holds that

$$|T_{1,k}|$$

$$\le \left|\frac{1}{n}\sum_{i=1}^n \int_0^\tau \frac{\{S^{*(0)}(t) - \widehat{S}^{(0)}(t)\}\frac{1}{n}\sum_{i'=1}^n Y_{i'}(t)\exp\{\mathbf{X}_{i'}^T(t)\widehat{\boldsymbol{\beta}}\}\mathbf{w}^{*T}\mathbf{X}_{i',\boldsymbol{\theta}}(t)X_{i',k}(t)}{\widehat{S}^{(0)}(t)S^{*(0)}(t)}dN_i(t)\right|$$

$$+ \left|\frac{1}{n}\sum_{i=1}^n \int_0^\tau \frac{\widehat{S}^{(0)}(t)\frac{1}{n}\sum_{i'=1}^n Y_{i'}(t)\left[\exp\{\mathbf{X}_{i'}^T(t)\widehat{\boldsymbol{\beta}}\} - \exp\{\mathbf{X}_{i'}^T(t)\boldsymbol{\beta}^*\}\right]\mathbf{w}^{*T}\mathbf{X}_{i',\boldsymbol{\theta}}(t)X_{i',k}(t)}{\widehat{S}^{(0)}(t)S^{*(0)}(t)}dN_i(t)\right|$$

$$\le \sup_{t\in[0,\tau]}\left|\frac{1}{n}\sum_{i=1}^n \frac{\{S^{*(0)}(t) - \widehat{S}^{(0)}(t)\}\left[\frac{1}{n}\sum_{i'=1}^n Y_{i'}(t)\exp\{\mathbf{X}_{i'}^T(t)\boldsymbol{\beta}^*\}\mathbf{w}^{*T}\mathbf{X}_{i',\boldsymbol{\theta}}(t)X_{i',k}(t)\right]}{\widehat{S}^{(0)}(t)S^{*(0)}(t)}\right|\cdot\tau$$

$$+ \left|\frac{1}{n}\sum_{i=1}^n \frac{\{S^{*(0)}(t) - \widehat{S}^{(0)}(t)\}\left[\frac{1}{n}\sum_{i'=1}^n Y_{i'}(t)\left[\exp\{\mathbf{X}_{i'}^T(t)\widehat{\boldsymbol{\beta}}\} - \exp\{\mathbf{X}_{i'}^T(t)\boldsymbol{\beta}^*\}\right]\mathbf{w}^{*T}\mathbf{X}_{i',\boldsymbol{\theta}}(t)X_{i',k}(t)\right]}{\widehat{S}^{(0)}(t)S^{*(0)}(t)}\right|\cdot\tau$$

$$+ \frac{1}{n}\sum_{i=1}^n \frac{\widehat{S}^{(0)}(t)\frac{1}{n}\sum_{i'=1}^n Y_{i'}(t)\left[\exp\{\mathbf{X}_{i'}^T(t)\widehat{\boldsymbol{\beta}}\} - \exp\{\mathbf{X}_{i'}^T(t)\boldsymbol{\beta}^*\}\right]\mathbf{w}^{*T}\mathbf{X}_{i',\boldsymbol{\theta}}(t)X_{i',k}(t)}{\widehat{S}^{(0)}(t)S^{*(0)}(t)}\cdot\tau$$

$$= \mathcal{O}_{\mathbb{P}}(s\sqrt{n^{-1}\log d}),$$

where the last equality holds by Assumptions 2.1 and 4.1 that $\boldsymbol{X}_i^T(t)\boldsymbol{\beta}^*$ is bounded, $S^{*(0)}(t)$ is bounded away from 0, and by Lemma C.2 that $|\widehat{S}^{(r)}(t) - S^{*(r)}(t)| = \mathcal{O}_{\mathbb{P}}(s\sqrt{n^{-1}\log d})$.

The term $T_2$ can be bounded by the similar argument, and our claim holds as desired. □

**Lemma C.7.** Under Assumptions 2.1 and 2.2, and if $n^{-1/2}s^3 \log d = o(1)$, the RE condition holds for the sample Hessian matrix $\nabla^2 \mathcal{L}(\widehat{\boldsymbol{\beta}})$. Specifically, for the vectors in the cone $\mathcal{C} = \{\mathbf{v}|\|\mathbf{v}_\mathcal{S}\|_1 \leq \xi \|\mathbf{v}_{\mathcal{S}^C}\|_1\}$, we have

$$\frac{\mathbf{v}^T \nabla^2 \mathcal{L}(\widehat{\boldsymbol{\beta}})\mathbf{v}}{\|\mathbf{v}\|_2} \geq \frac{1}{2}\kappa^2\big(\xi, |\mathcal{S}|; \nabla^2 \mathcal{L}(\boldsymbol{\beta}^*)\big), \text{ for all } \mathbf{v} \in \mathcal{C}.$$

*Proof.* By Lemma 3.2 of Huang et al. (2013), we have $\exp(-2\xi_\mathbf{b})\nabla^2 \mathcal{L}(\boldsymbol{\beta}) \preceq \nabla^2 \mathcal{L}(\boldsymbol{\beta} + \mathbf{b})$, where $\xi_\mathbf{b} = \max_{u \geq 0} \max_{i,i',k,k'} |\mathbf{b}^T\{\boldsymbol{X}_{ik}(u) - \boldsymbol{X}_{i'k'}(u)\}|$. Let $\mathbf{b} = \widehat{\boldsymbol{\beta}} - \boldsymbol{\beta}^*$. By Assumption 2.1 that $\|\{\boldsymbol{X}_{ik}(u) - \boldsymbol{X}_{i'k'}(u)\}\|_\infty \leq C_X$, we have $\xi_\mathbf{b} = \mathcal{O}_{\mathbb{P}}(s\sqrt{n^{-1}\log d})$ by (2.2), we have $\|\widehat{\boldsymbol{\beta}} - \boldsymbol{\beta}^*\|_1 = \mathcal{O}_{\mathbb{P}}(s\lambda)$. By the scaling assumption that $n^{-1/2}s^3 \log d = o(1)$, we have $\xi_\mathbf{b} \leq \frac{1}{2}\log 2$. Consequently, $\exp(-2\xi_\mathbf{b}) \geq 1/2$. We have $\nabla^2 \mathcal{L}(\widehat{\boldsymbol{\beta}}) \succeq \frac{1}{2} \cdot \nabla^2 \mathcal{L}(\boldsymbol{\beta}^*)$. Since the cone $\mathcal{C}$ is a subset of $\mathbb{R}^d$, our claim follows as desired. □

**Lemma C.8.** Under Assumptions 2.1, 2.2, 4.1, 4.2 and 4.3, if

$$\|\nabla^2_{\boldsymbol{\theta}\alpha}\mathcal{L}(\widehat{\boldsymbol{\beta}}) - \mathbf{w}^{*T}\nabla^2_{\boldsymbol{\theta}\boldsymbol{\theta}}\mathcal{L}(\widehat{\boldsymbol{\beta}})\|_\infty \leq \lambda', \quad (C.7)$$

we have, the Dantzig selector $\widehat{\mathbf{w}}$ defined in (3.2) satisfies

$$\|\widehat{\mathbf{w}} - \mathbf{w}^*\|_1 \leq \frac{16\lambda' s'}{\kappa^2(1, s'; \nabla^2 \mathcal{L}(\boldsymbol{\beta}^*))}.$$

*Proof.* We first derive the result that the vector $\widehat{\Delta} = \widehat{\mathbf{w}} - \mathbf{w}^*$ belongs to the cone $\mathcal{C} = \{\mathbf{v}|\|\mathbf{v}_{\mathcal{S}^C}\|_1 \leq \|\mathbf{v}_\mathcal{S}\|_1\}$. By our assumption (C.7), and since $\|\widehat{\mathbf{w}}\|_1 \leq \|\mathbf{w}^*\|_1$ by the optimality condition of Dantzig selector in (D.2), we have

$$\|\widehat{\mathbf{w}}_\mathcal{S}\|_1 + \|\widehat{\mathbf{w}}_{\mathcal{S}^C}\|_1 \leq \|\mathbf{w}^*_\mathcal{S}\|_1,$$

where we use the fact that $\|\mathbf{w}^*_{\mathcal{S}^C}\|_1 = 0$.

By triangle inequality, we have

$$\|\mathbf{w}^*_\mathcal{S}\|_1 \leq \|\widehat{\mathbf{w}}_\mathcal{S}\|_1 + \|\widehat{\Delta}_\mathcal{S}\|_1.$$

Summing up the above two inequalities, we have

$$\|\widehat{\Delta}_{\mathcal{S}^C}\|_1 \leq \|\widehat{\Delta}_\mathcal{S}\|_1. \quad (C.8)$$

Meanwhile, by the feasibility conditions of the Dantzig selector $\widehat{\mathbf{w}}$ and $\mathbf{w}^*$, we have

$$\|\nabla^2_{\boldsymbol{\theta}\boldsymbol{\theta}}\mathcal{L}(\widehat{\boldsymbol{\beta}})\widehat{\Delta}\|_\infty \leq \|\nabla^2_{\boldsymbol{\theta}\alpha}\mathcal{L}(\widehat{\boldsymbol{\beta}}) - \mathbf{w}^{*T}\nabla^2_{\boldsymbol{\theta}\boldsymbol{\theta}}\mathcal{L}(\widehat{\boldsymbol{\beta}})\|_\infty + \|\nabla^2_{\boldsymbol{\theta}\alpha}\mathcal{L}(\widehat{\boldsymbol{\beta}}) - \widehat{\mathbf{w}}\nabla^2_{\boldsymbol{\theta}\boldsymbol{\theta}}\mathcal{L}(\widehat{\boldsymbol{\beta}})\|_\infty \leq 2\lambda'. \quad (C.9)$$

By (C.8) and (C.9), we have

$$\widehat{\Delta}^T \nabla^2_{\boldsymbol{\theta}\boldsymbol{\theta}}\mathcal{L}(\widehat{\boldsymbol{\beta}})\widehat{\Delta} \leq \|\widehat{\Delta}\|_1 \|\nabla^2_{\boldsymbol{\theta}\boldsymbol{\theta}}\mathcal{L}(\widehat{\boldsymbol{\beta}})\widehat{\Delta}\|_\infty \leq 2\lambda'\|\widehat{\Delta}\|_1 \leq 4\lambda'\|\widehat{\Delta}_\mathcal{S}\|_1.$$

By Lemma C.8, it holds that

$$\widehat{\Delta}^T \nabla^2_{\boldsymbol{\theta\theta}} \mathcal{L}(\widehat{\boldsymbol{\beta}})\widehat{\Delta} \geq \frac{1}{2}\kappa^2(1, s'; \nabla^2 \mathcal{L}(\boldsymbol{\beta}^*))\|\widehat{\Delta}_{\mathcal{S}}\|_2^2,$$

which implies that

$$\widehat{\Delta}^T \nabla^2_{\boldsymbol{\theta\theta}} \mathcal{L}(\widehat{\boldsymbol{\beta}})\widehat{\Delta} \geq \frac{1}{2}\kappa^2(1, s'; \nabla^2 \mathcal{L}(\boldsymbol{\beta}^*))s'^{-1}\|\widehat{\Delta}_{\mathcal{S}}\|_2^1.$$

Consequently, we have

$$\|\widehat{\Delta}_{\mathcal{S}}\|_1 \leq \frac{8\lambda' s'}{\kappa^2(1, s'; \nabla^2 \mathcal{L}(\boldsymbol{\beta}^*))}.$$

By (C.8), it holds that

$$\|\widehat{\Delta}\|_1 \leq 2\|\widehat{\Delta}_{\mathcal{S}}\|_1 \leq \frac{16\lambda' s'}{\kappa^2(1, s'; \nabla^2 \mathcal{L}(\boldsymbol{\beta}^*))}$$

as desired. □

## D Extensions to Multivariate Failure Time Data

In real applications, it is also of interest to study multivariate failure time outcomes. For example, Cai et al. (2005) consider the time to coronary heart disease and time to cerebrovascular accident. In their study, the primary sampling unit is the family. Using multivariate model, it takes the advantage to incorporate the assumption that the failure times for subjects within a family are likely to be correlated. In this section, we extend our method to conduct inference in the high dimensional multivariate failure time setting.

To be more specific about the model, assume there are $n$ independent clusters (families). Each cluster $i$ contains $M_i$ subjects, and for each subject, there are $K$ types of failure may occur. Thus, it is reasonable to assume that the number $K$ is fixed that does not increase with dimensionality $d$ and sample size $n$. For example, Cai et al. (2005) study the time to coronary heart disease and the time to cerebrovascular accident where $K = 2$. Denote the covariates of the $k$th failure type of subject $m$ in cluster $i$ at time $t$ by $\boldsymbol{X}_{ikm}(t)$. The marginal hazards model is taken as

$$\Lambda_{ikm}\{t|\boldsymbol{X}_{ikm}(t)\} = \Lambda_{0k}(t)\exp\{\boldsymbol{X}_{ikm}^T(t)\boldsymbol{\beta}\},$$

where the baseline hazard functions $\Lambda_{0k}(t)$'s are treated as nuisance parameters, and the model is known as mixed baseline hazards model. Using this model, our inference procedures are conducted based on the pseudo-partial likelihood approach, since the working model does not assume any correlation for the different failure times within each cluster. The log pseudo-partial likelihood loss function is

$$\mathcal{L}(\boldsymbol{\beta}) = -\frac{1}{n}\Big[\sum_{k=1}^{K}\sum_{i=1}^{n}\sum_{m=1}^{M_i}\int_0^{\tau}\boldsymbol{X}_{ikm}^T(u)\boldsymbol{\beta} dN_{ikm}(u) -$$

$$\sum_{k=1}^{K}\int_0^{\tau}\log\Big[\sum_{i=1}^{n}\sum_{m=1}^{M_i}Y_{ikm}(u)\exp\{\boldsymbol{X}_{ikm}^T(u)\boldsymbol{\beta}\}\Big]d\overline{N}_k(u)\Big],$$

where $Y_{ikm}(t)$ and $N_{ikm}(t)$ denote the at risk indicator and the number of observed failure event at time $t$ of the $k$th type on subject $m$ in cluster $i$, and $\overline{N}_k = \sum_{i=1}^{n} \sum_{m=1}^{M_i} N_{ikm}$ for each $k$. The penalized maximum pseudo likelihood estimator is

$$\widehat{\boldsymbol{\beta}} = \underset{\boldsymbol{\beta} \in \mathbb{R}^d}{\operatorname{argmin}} \mathcal{L}(\boldsymbol{\beta}) + \mathcal{P}_\lambda(\boldsymbol{\beta}). \tag{D.1}$$

To connect the multivariate failure time model with Cox's proportional hazards model, first, we observe that we can drop the index $m$. This is by the fact that, for each $(i,m)$ where $i \in \{1,...n\}$ and $m \in \{1,...,M_i\}$, we can map $(i,m)$ to $i' = \sum_{j=1}^{i-1} M_j + m$, and we define $\sum_{j=1}^{0} M_j = 0$. It is not difficult to see the mapping is a bijection. After the mapping, the penalized estimator remains the same. Thus, without loss of generality, we assume $M_i = 1$ for all $i$, and we drop the index $m$. Next, we observe that the loss function $\mathcal{L}(\boldsymbol{\beta})$ is decomposable that

$$\mathcal{L}(\boldsymbol{\beta}) = \sum_{k=1}^{K} \mathcal{L}^{(k)}(\boldsymbol{\beta}),$$

where

$$\mathcal{L}^{(k)}(\boldsymbol{\beta}) = -\frac{1}{n}\Big[\sum_{i=1}^{n} \int_0^t \boldsymbol{X}_{ik}^T(u)\boldsymbol{\beta} dN_{ik}(u) - \int_0^t \log\Big[\sum_{i=1}^{n} Y_{ik}(u)\exp\big\{\boldsymbol{X}_{ik}^T(u)\boldsymbol{\beta}\big\}\Big]d\overline{N}_k(u)\Big].$$

Thus, the loss function of multivariate failure time model can be decomposed into a sum of $K$ loss functions of Cox's proportional hazards models. However, the extension of the inference of the Cox model to multivariate failure time model is not trivial since the loss function is derived from a pseudo-likelihood function.

First, we extend the estimation procedure to the multivariate failure time model in the high dimensional setting, where we take $\mathcal{P}_\lambda(\boldsymbol{\beta}) = \lambda\|\boldsymbol{\beta}\|_1$. It is not difficult to obtain that (2.2) holds for the multivariate failure time model. An alternative approach is that we estimate $\boldsymbol{\beta}^*$ using each type $k$ of failure time independently. Specifically, we construct the estimator $\widehat{\boldsymbol{\beta}}$ by

$$\widehat{\boldsymbol{\beta}} = K^{-1}\sum_{k=1}^{K} \widehat{\boldsymbol{\beta}}^{(k)}, \text{ where } \widehat{\boldsymbol{\beta}}^{(k)} = \underset{\boldsymbol{\beta}^{(k)}}{\operatorname{argmin}} \mathcal{L}^{(k)}(\boldsymbol{\beta}^{(k)}) + \lambda\|\boldsymbol{\beta}^{(k)}\|_1, \text{ for all } k.$$

Since for each $\widehat{\boldsymbol{\beta}}^{(k)}$, $\|\widehat{\boldsymbol{\beta}}^{(k)} - \boldsymbol{\beta}^*\|_1 = \mathcal{O}_\mathbb{P}(\lambda s)$ by (2.2), it is readily seen that $\|\widehat{\boldsymbol{\beta}} - \boldsymbol{\beta}^*\|_1 = \mathcal{O}_\mathbb{P}(\lambda s)$.

We extend the decorrelated score, Wald and partial likelihood ratio tests to the multivariate failure time model. We first introduce some notation. For $k = 1,...,K$,

$$S_k^{(r)}(t,\boldsymbol{\beta}) = \frac{1}{n}\sum_{i=1}^{n} \boldsymbol{X}_{ik}^{\otimes r}(t)Y_{ik}(t)\exp\{\boldsymbol{X}_{ik}^T(t)\boldsymbol{\beta}\}, \text{ for } r = 0,1,2, \text{ and } \overline{\boldsymbol{Z}}_{kn}(t,\boldsymbol{\beta}) = \frac{S_k^{(1)}(t,\boldsymbol{\beta})}{S_k^{(0)}(t,\boldsymbol{\beta})},$$

where their corresponding population versions are

$$\mathbf{s}_k^{(r)}(t,\boldsymbol{\beta}) = \mathbb{E}\big[Y_k(t)\boldsymbol{X}_{ik}(t)^{\otimes r}\exp\{\boldsymbol{X}_{ik}(t)\boldsymbol{\beta}\}\big], \text{ for } r = 0,1,2, \text{ and } \mathbf{e}_k(t,\boldsymbol{\beta}) = \mathbf{s}_k^{(1)}(t,\boldsymbol{\beta})/\mathbf{s}_k^{(0)}(t,\boldsymbol{\beta}).$$

Next, we derive the gradient and Hessian matrix at the point $\boldsymbol{\beta}$ of the loss function,

$$\nabla \mathcal{L}(\boldsymbol{\beta}) = -\frac{1}{n}\sum_{k=1}^{K}\sum_{i=1}^{n} \int_0^\tau \{\boldsymbol{X}_{ik}(u) - \overline{\boldsymbol{Z}}_{kn}(u,\boldsymbol{\beta})\}dN_{ik}(u),$$

and

$$\nabla^2 \mathcal{L}(\boldsymbol{\beta}) = \frac{1}{n} \sum_{k=1}^{K} \int_0^{\tau} \Big\{ \frac{S_k^{(2)}(u, \boldsymbol{\beta})}{S_k^{(0)}(u, \boldsymbol{\beta})} - \overline{\mathbf{Z}}_{kn}(u, \boldsymbol{\beta})^{\otimes 2} \Big\} d\overline{N}_k(u).$$

The population version of the gradient and Hessian matrix are

$$\mathbf{g}(\boldsymbol{\beta}) = \sum_{k=1}^{K} \mathbb{E}\Big[ \int_0^{\tau} \{ \mathbf{X}(u) - \mathbf{e}_k(u, \boldsymbol{\beta}) \} d\overline{N}_k(u) \Big],$$

and

$$\mathbf{H}(\boldsymbol{\beta}) = \sum_{k=1}^{K} \mathbb{E}\Big[ \int_0^{\tau} \Big\{ \frac{\mathbf{s}_k^{(2)}(u, \boldsymbol{\beta})}{\mathbf{s}_k^{(0)}(u, \boldsymbol{\beta})} - \mathbf{e}(u, \boldsymbol{\beta})^{\otimes 2} \Big\} d\overline{N}_k(u) \Big].$$

For notational simplicity, let $\mathbf{H}^* = \mathbf{H}(\boldsymbol{\beta}^*)$.

Note that, utilizing the decomposable structure, by the similar argument, the concentration results in Appendix C hold for the empirical gradient and Hessian matrix. We estimate the decorrelation vector $\mathbf{w}^* = \mathbf{H}_{\boldsymbol{\theta}\boldsymbol{\theta}}^{*-1} \mathbf{H}_{\boldsymbol{\theta}\alpha}^*$ by the following Dantzig selector

$$\widehat{\mathbf{w}} = \operatorname{argmin} \|\mathbf{w}\|_1, \quad \text{subject to } \|\nabla^2_{\boldsymbol{\theta}\alpha}\mathcal{L}(0, \widehat{\boldsymbol{\theta}}) - \mathbf{w}^T \nabla^2_{\boldsymbol{\theta}\boldsymbol{\theta}}\mathcal{L}(0, \widehat{\boldsymbol{\theta}})\|_\infty \leq \delta, \tag{D.2}$$

where $\delta$ is a tuning parameter. The rate of convergence of $\widehat{\mathbf{w}}$ follows by the similar argument as in Lemma C.8.

We first introduce the decorrelated score test in multivariate failure time model. Suppose the null hypothesis is $H_0$: $\alpha^* = 0$, and the alternative hypothesis is $H_a$: $\alpha^* \neq 0$. The decorrelated score function is constructed similar to (3.3) that

$$\widehat{U}^M(0, \widehat{\boldsymbol{\theta}}) = \nabla_\alpha \mathcal{L}(0, \widehat{\boldsymbol{\theta}}) - \widehat{\mathbf{w}}^T \nabla_{\boldsymbol{\theta}} \mathcal{L}(0, \widehat{\boldsymbol{\theta}}). \tag{D.3}$$

The main technical difference between the multivariate failure time model and the univariate Cox's model is that, the loss function of Cox's model is a log profile likelihood function, and Bartlett's identity $\operatorname{Var}\{\nabla \mathcal{L}(\boldsymbol{\beta}^*)\} = \nabla^2 \mathcal{L}(\boldsymbol{\beta}^*)$ holds. In multivariate case, this identity does not hold. We need the following lemma which is analogous to Lemma A.1. We omit the proof details to avoid repetition.

**Lemma D.1.** For any vector $\mathbf{v} \in \mathbb{R}^d$, if $\|\mathbf{v}\|_0 \leq s'$ and $n^{-1}\sqrt{s' \log d} = o(1)$, it holds that

$$\frac{\sqrt{n} \mathbf{v}^T \nabla \mathcal{L}(\boldsymbol{\beta}^*)}{\sqrt{\mathbf{v}^T \boldsymbol{\Omega} \mathbf{v}}} \xrightarrow{d} N(0, 1). \quad \text{where } \boldsymbol{\Omega} = \operatorname{Var}\{\sqrt{n} \nabla \mathcal{L}(\boldsymbol{\beta}^*)\} \in \mathbb{R}^{d \times d}.$$

By the similar argument as in Theorem 4.4, we derive the asymptotic normality of $\widehat{U}^M(0, \widehat{\boldsymbol{\theta}})$ in the next theorem.

**Theorem D.2.** Suppose that Assumptions 2.1, 2.2, 4.1, 4.2 and 4.3 hold. Let $\widehat{U}^M(0, \widehat{\boldsymbol{\theta}})$ be defined in (D.3). Under the null hypothesis that $\alpha^* = 0$ and if $\lambda \asymp \sqrt{n^{-1} \log d}$, $\delta \asymp s'\sqrt{n^{-1} \log d}$, $n^{-1/2} s^3 \log d = o(1)$, we have

$$\sqrt{n}\widehat{U}^M(0, \widehat{\boldsymbol{\theta}}) \xrightarrow{d} Z, \text{ where } Z \sim N(0, \sigma^2) \text{ and } \sigma^2 = \Omega_{\alpha\alpha} - 2\mathbf{w}^{*T}\Omega_{\boldsymbol{\theta}\alpha} + \mathbf{w}^{*T}\Omega_{\boldsymbol{\theta}\boldsymbol{\theta}}\mathbf{w}^*.$$

*Proof.* By the definition of $\widehat{U}^M(0, \widehat{\boldsymbol{\theta}})$ and mean value theorem, we have, for some $z, z' \in [0,1]$, $\bar{\boldsymbol{\theta}} = \boldsymbol{\theta}^* + z(\widehat{\boldsymbol{\theta}} - \boldsymbol{\theta}^*)$ and $\widetilde{\boldsymbol{\theta}} = \boldsymbol{\theta}^* + z'(\widehat{\boldsymbol{\theta}} - \boldsymbol{\theta}^*)$,

$$\begin{aligned}
\widehat{U}^M(0, \widehat{\boldsymbol{\theta}}) &= \nabla_\alpha \mathcal{L}(0, \widehat{\boldsymbol{\theta}}) - \widehat{\mathbf{w}}^T \nabla_{\boldsymbol{\theta}} \mathcal{L}(0, \widehat{\boldsymbol{\theta}}) \\
&= \nabla_\alpha \mathcal{L}(0, \boldsymbol{\theta}^*) + \nabla_{\alpha\boldsymbol{\theta}} \mathcal{L}(0, \boldsymbol{\theta}^*) - \{\widehat{\mathbf{w}}^T \nabla_{\boldsymbol{\theta}} \mathcal{L}(0, \boldsymbol{\theta}^*) + \widehat{\mathbf{w}} \nabla_{\boldsymbol{\theta}\boldsymbol{\theta}} \mathcal{L}(0, \widetilde{\boldsymbol{\theta}})(\widehat{\boldsymbol{\theta}} - \boldsymbol{\theta}^*)\} \\
&= \underbrace{\nabla_\alpha \mathcal{L}(0, \boldsymbol{\theta}^*) - \mathbf{w}^{*T} \nabla_{\boldsymbol{\theta}} \mathcal{L}(0, \boldsymbol{\theta}^*)}_{S} + \underbrace{(\mathbf{w}^* - \widehat{\mathbf{w}})^T \nabla_{\boldsymbol{\theta}} \mathcal{L}(0, \boldsymbol{\theta}^*)}_{E_1} \\
&\quad + \underbrace{\{\nabla_{\alpha\boldsymbol{\theta}} \mathcal{L}(0, \bar{\boldsymbol{\theta}}) - \widehat{\mathbf{w}}^T \nabla_{\boldsymbol{\theta}\boldsymbol{\theta}} \mathcal{L}(0, \widetilde{\boldsymbol{\theta}})\}(\widehat{\boldsymbol{\theta}} - \boldsymbol{\theta}^*)}_{E_2}.
\end{aligned}$$

Using Lemma D.1, taking $\mathbf{b} = (1, -\mathbf{w}^{*T})^T$ and by the assumption that $\|\mathbf{w}^*\|_0 \leq s'$, it holds that

$$\sqrt{n} S \xrightarrow{d} Z, \text{ where } Z \sim N(0, \sigma^2) \text{ and } \sigma^2 = \Omega_{\alpha\alpha} - 2\mathbf{w}^{*T}\Omega_{\boldsymbol{\theta}\alpha} + \mathbf{w}^{*T}\Omega_{\boldsymbol{\theta}\boldsymbol{\theta}}\mathbf{w}^*.$$

Following a similar proof as that in Theorem 4.4 and utilizing the separable structure in multivariate failure time model, we have $\sqrt{n} E_1 = o_\mathbb{P}(1)$ and $\sqrt{n} E_2 = o_\mathbb{P}(1)$. This concludes our proof. $\square$

**Remark D.3.** Under the assumptions of D.2, using plug-in estimator $\widehat{\sigma}^2 = \widehat{\Omega}_{\alpha\alpha} - 2\widehat{\mathbf{w}}\widehat{\Omega}_{\boldsymbol{\theta}\alpha} + \widehat{\mathbf{w}}^T\widehat{\Omega}_{\boldsymbol{\theta}\boldsymbol{\theta}}\widehat{\mathbf{w}}$ converges to $\sigma^2$ at the rate of $\mathcal{O}_\mathbb{P}(s's\sqrt{n^{-1}\log d}) = o_\mathbb{P}(1)$.

Next, we extend the decorrelated Wald test to the multivariate failure time model, which constructs confidence intervals for $\alpha^*$. We first estimate $\boldsymbol{\beta}^*$ by $\ell_1$-penalized estimator $\widehat{\boldsymbol{\beta}} = (\widehat{\alpha}, \widehat{\boldsymbol{\theta}})$. Let

$$\widetilde{\alpha}^M = \widehat{\alpha} - \left\{\frac{\partial \widehat{U}^M(\widehat{\alpha}, \widehat{\boldsymbol{\theta}})}{\partial \alpha}\right\}^{-1} \widehat{U}^M(\widehat{\alpha}, \widehat{\boldsymbol{\theta}}).$$

We derive the asymptotic normality of $\widetilde{\alpha}^M$ in the next theorem.

**Theorem D.4.** Suppose Assumptions 2.1, 2.2, 4.1, 4.2 and 4.3 hold. For $\lambda \asymp \sqrt{n^{-1}\log d}$, $\delta \asymp s'\sqrt{n^{-1}\log d}$ and $n^{-1/2} s^3 \log d = o(1)$, under the null hypothesis that $\alpha^* = 0$, we have

$$\sqrt{n}\widetilde{\alpha} \xrightarrow{d} Z, \quad \text{where } Z \sim N(0, \sigma^2/\gamma^4),$$

and $\sigma^2 = \Omega_{\alpha\alpha} - 2\mathbf{w}^{*T}\Omega_{\boldsymbol{\theta}\alpha} + \mathbf{w}^{*T}\Omega_{\boldsymbol{\theta}\boldsymbol{\theta}}\mathbf{w}^*$, $\gamma^2 = \mathbf{H}^*_{\alpha\alpha} - \mathbf{w}^{*T}\mathbf{H}^*_{\boldsymbol{\theta}\alpha}$.

*Proof.* By the definition of $\widetilde{\alpha}$, we have,

$$\begin{aligned}
\widetilde{\alpha} &= \widehat{\alpha} - \left[\gamma^{-2} - \gamma^{-2} + \left\{\frac{\partial \widehat{U}^M(\widehat{\alpha}, \widehat{\boldsymbol{\theta}})}{\partial \alpha}\right\}^{-1}\right] \widehat{U}(\widehat{\alpha}, \widehat{\boldsymbol{\theta}}) \\
&= \widehat{\alpha} - \gamma^{-2}\left\{\widehat{U}^M(0, \widehat{\boldsymbol{\theta}}) + \frac{(\widehat{\alpha}-0)\partial \widehat{U}^M(\bar{\alpha}, \widehat{\boldsymbol{\theta}})}{\partial \alpha}\right\} + \left[\gamma^{-2} - \left\{\frac{\partial \widehat{U}^M(\widehat{\alpha}, \widehat{\boldsymbol{\theta}})}{\partial \alpha}\right\}^{-1}\right]\widehat{U}(\widehat{\alpha}, \widehat{\boldsymbol{\theta}}), \text{ where} \\
&= \widehat{\alpha} - \gamma^{-2}\widehat{U}^M(0, \widehat{\boldsymbol{\theta}}) - \widehat{\alpha}\gamma^2\gamma^{-2} + \widehat{\alpha}\gamma^{-2}\left\{\gamma^2 - \frac{\partial \widehat{U}^M(\bar{\alpha}, \widehat{\boldsymbol{\theta}})}{\partial \alpha}\right\} + \widehat{U}^M(\widehat{\alpha}, \widehat{\boldsymbol{\theta}})\left[\gamma^{-2} - \left\{\frac{\partial \widehat{U}^M(\widehat{\alpha}, \widehat{\boldsymbol{\theta}})}{\partial \alpha}\right\}^{-1}\right], \\
&= \underbrace{-\gamma^{-2}\widehat{U}^M(0, \widehat{\boldsymbol{\theta}})}_{S} + \underbrace{\widehat{\alpha}\gamma^{-2}\left\{\gamma^2 - \frac{\partial \widehat{U}^M(\bar{\alpha}, \widehat{\boldsymbol{\theta}})}{\partial \alpha}\right\}}_{R_1} + \underbrace{\widehat{U}^M(\widehat{\alpha}, \widehat{\boldsymbol{\theta}})\left[\gamma^{-2} - \left\{\frac{\partial \widehat{U}^M(\widehat{\alpha}, \widehat{\boldsymbol{\theta}})}{\partial \alpha}\right\}^{-1}\right]}_{R_2},
\end{aligned}$$

where the second equality holds by mean value theorem for some $\bar{\alpha} = v\widehat{\alpha}$ and $v \in [0,1]$. For the first term above, we have $\sqrt{n}S \xrightarrow{d} Z$ where $Z \sim N(0, \sigma^2/\gamma^4)$ by Theorem D.2. In addition, $\sqrt{n}R_1 = o_{\mathbb{P}}(1)$ and $\sqrt{n}R_2 = o_{\mathbb{P}}(1)$ by the similar argument in Theorem 4.7. This concludes the proof. □

Finally, we extend the decorrelated partial likelihood ratio test to the multivariate failure time model. The test statistic is
$$2n\{\mathcal{L}(0, \widehat{\boldsymbol{\theta}}) - \mathcal{L}(\widetilde{\alpha}, \widehat{\boldsymbol{\theta}} - \widetilde{\alpha}\widehat{\mathbf{w}})\}.$$
Under the null hypothesis, the test statistic follows a weighted chi-squared distribution as shown in the following theorem.

**Theorem D.5.** Suppose Assumptions 2.1, 2.2, 4.1, 4.2 and 4.3 hold. If $\lambda \asymp \sqrt{n^{-1}\log d}$, $\delta \asymp s'\sqrt{n^{-1}\log d}$ and $n^{-1/2}s^3\log d$, under the null hypothesis $\alpha^* = 0$, we have
$$2n\{\mathcal{L}(0, \widehat{\boldsymbol{\theta}}) - \mathcal{L}(\widetilde{\alpha}, \widehat{\boldsymbol{\theta}} - \widetilde{\alpha}\widehat{\mathbf{w}})\} \xrightarrow{d} \sigma^2\gamma^{-2}Z_\chi, \text{ where } Z_\chi \sim \chi_1^2,$$
and $\sigma^2 = \Omega_{\alpha\alpha} - 2\mathbf{w}^{*T}\Omega_{\boldsymbol{\theta}\alpha} + \mathbf{w}^{*T}\Omega_{\boldsymbol{\theta\theta}}\mathbf{w}^*$, $\gamma^2 = \mathbf{H}^*_{\alpha\alpha} - \mathbf{w}^{*T}\mathbf{H}^*_{\boldsymbol{\theta}\alpha}$.

*Proof.* We have, by mean value theorem, for some $\bar{\alpha} = v_1\widehat{\alpha}$, $\bar{\alpha}' = v_2\widehat{\alpha}$, $\bar{\boldsymbol{\theta}} = \boldsymbol{\theta}^* + t_3(\widehat{\boldsymbol{\theta}} - \boldsymbol{\theta}^*)$ and $\bar{\boldsymbol{\theta}}' = \boldsymbol{\theta}^* + v_4(\widehat{\boldsymbol{\theta}} - \boldsymbol{\theta}^*)$ and $0 \le v_1, v_2, v_3, v_4 \le 1$,

$$\mathcal{L}(\widetilde{\alpha}, \widehat{\boldsymbol{\theta}} - \widetilde{\alpha}\widehat{\mathbf{w}}) - \mathcal{L}(0, \widehat{\boldsymbol{\theta}})$$
$$= \widetilde{\alpha}\nabla_\alpha\mathcal{L}(0, \widehat{\boldsymbol{\theta}}) - \widetilde{\alpha}\widehat{\mathbf{w}}^T\nabla_{\boldsymbol{\theta}}\mathcal{L}(0, \widehat{\boldsymbol{\theta}}) + \frac{\widetilde{\alpha}^2}{2}\nabla_{\alpha\alpha}(\mathcal{L}(\bar{\alpha}, \widehat{\boldsymbol{\theta}}) + \widehat{\mathbf{w}}^T\nabla_{\boldsymbol{\theta\theta}}\mathcal{L}(0, \bar{\boldsymbol{\theta}})\widehat{\mathbf{w}} - \widetilde{\alpha}^2\widehat{\mathbf{w}}^T\nabla_{\alpha\boldsymbol{\theta}}\mathcal{L}(\bar{\alpha}', \bar{\boldsymbol{\theta}}')$$
$$= \underbrace{\widetilde{\alpha}\widehat{U}(0, \widehat{\boldsymbol{\theta}})}_{L} + \underbrace{\frac{\widetilde{\alpha}^2}{2}\{\nabla_{\alpha\alpha}\mathcal{L}(\bar{\alpha}, \widehat{\boldsymbol{\theta}}) + \widehat{\mathbf{w}}^T\nabla_{\boldsymbol{\theta\theta}}\mathcal{L}(0, \bar{\boldsymbol{\theta}})\widehat{\mathbf{w}} - 2\widehat{\mathbf{w}}\nabla_{\alpha\boldsymbol{\theta}}\mathcal{L}(\bar{\alpha}', \bar{\boldsymbol{\theta}}')\}}_{E}.$$

We first look at the term $L$. By Theorem D.2, we have $\widehat{U}(0, \widehat{\boldsymbol{\theta}}) = \widehat{U}(0, \widehat{\boldsymbol{\theta}}^*) + o_{\mathbb{P}}(n^{-1/2})$, and by Theorem D.4 $\widetilde{\alpha} = -\gamma^{-2}\widehat{U}(0, \widehat{\boldsymbol{\theta}}) + o_{\mathbb{P}}(n^{-1/2})$, we have
$$L = -\gamma^{-2}\widehat{U}^M(0, \widehat{\boldsymbol{\theta}})^2 + o_{\mathbb{P}}(n^{-1})$$

Next, we look at the term $E$,
$$E = \underbrace{\frac{\widetilde{\alpha}^2}{2}\left(\mathbf{H}^*_{\alpha\alpha} + \mathbf{H}^*_{\alpha\boldsymbol{\theta}}\mathbf{H}^{*-1}_{\boldsymbol{\theta\theta}}\mathbf{H}^*_{\boldsymbol{\theta}\alpha} - 2\mathbf{H}^*_{\alpha\boldsymbol{\theta}}\mathbf{H}^{*-1}_{\boldsymbol{\theta\theta}}\mathbf{H}^*_{\boldsymbol{\theta}\alpha}\right)}_{E_1}$$
$$+ \underbrace{\frac{\widetilde{\alpha}^2}{2}\left[\{\nabla_{\alpha\alpha}\mathcal{L}(\bar{\alpha}, \widehat{\boldsymbol{\theta}}) - \mathbf{H}^*_{\alpha\alpha}\} + \{\widehat{\mathbf{w}}^T\nabla_{\boldsymbol{\theta\theta}}\mathcal{L}(0, \bar{\boldsymbol{\theta}})\widehat{\mathbf{w}} - \mathbf{w}^*\mathbf{H}^*_{\boldsymbol{\theta\theta}}\mathbf{w}^*\} - 2\{\widetilde{\mathbf{w}}\nabla_{\alpha\boldsymbol{\theta}}\mathcal{L}(\bar{\alpha}', \bar{\boldsymbol{\theta}}') - \mathbf{H}^*_{\alpha\boldsymbol{\theta}}\mathbf{w}^*\}\right]}_{E_2}.$$

By Theorem D.4, it holds that $2nE_1 \xrightarrow{d} \sigma^2\gamma^{-2}Z_\chi$. In addition, by the similar argument as in Theorem 4.9, we have $E_2 = o_{\mathbb{P}}(n^{-1})$. Thus, we have
$$2n\{\mathcal{L}(0, \widehat{\boldsymbol{\theta}}) - \mathcal{L}(\widetilde{\alpha}, \widehat{\boldsymbol{\theta}} - \widetilde{\alpha}\widehat{\mathbf{w}})\} \xrightarrow{d} \sigma^2\gamma^{-2}Z_\chi, \text{ where } Z_\chi \sim \chi_1^2,$$
which concludes our proof. □